\documentclass[sn-basic]{sn-jnl}

\jyear{2022}%

\theoremstyle{thmstyleone}%
\theoremstyle{thmstyletwo}%

\theoremstyle{thmstylethree}%
\UseRawInputEncoding
\usepackage{amssymb}
\usepackage{graphicx}
\usepackage{epsfig}
\usepackage{mathtools}
\usepackage{amsmath}
\usepackage{xparse}
\usepackage{textcmds}
\usepackage[export]{adjustbox}
\usepackage{tabularx}
\usepackage[justification=centering]{caption}
\usepackage{booktabs}
\usepackage{xcolor}
\usepackage{multirow}
\usepackage{tabularx}
\usepackage{fixltx2e}
\usepackage{natbib}
\newcolumntype{L}[1]{>{\raggedright\arraybackslash}p{#1}}
\newcolumntype{C}[1]{>{\centering\arraybackslash}p{#1}}
\newcolumntype{R}[1]{>{\raggedleft\arraybackslash}p{#1}}
\usepackage{algorithm,algpseudocode,float}
\usepackage{lipsum}

\setcitestyle{citepsep={;}}
\usepackage{enumitem}
\algrenewcommand\algorithmicrequire{\textbf{Input:}}
\algrenewcommand\algorithmicensure{\textbf{Output:}}
\algnewcommand\algorithmicforeach{\textbf{for each}}
\algdef{S}[FOR]{ForEach}[1]{\algorithmicforeach\ #1\ \algorithmicdo}
\usepackage{algpseudocode}
\usepackage{color}

\usepackage{amssymb}
\usepackage{txfonts}
\usepackage{mathdots}

\usepackage{subfigure}
\usepackage{longtable}
\usepackage{xcolor}
\usepackage{adjustbox}

\makeatletter
\def\newmaketag{%
  \def\maketag@@@##1{\hbox{\m@th\normalfont\normalsize##1}}%
  }
\makeatother



\def\makeheadbox{\relax}

\begin{document}
\def\makeheadbox{\relax}
\title[Forest-ORE: Mining Optimal Rule Ensemble to interpret Random Forest models]{\textbf{Forest-ORE}: Mining Optimal Rule Ensemble to interpret Random Forest models}


\author*[1]{\fnm{Haddouchi}\sur{ Maissae}}\email{maissaehaddouchi@research.emi.ac.ma}
\author[1]{\fnm{Berrado}\sur{ Abdelaziz}}\email{berrado@emi.ac.ma}
\affil[1]{\orgdiv{Mohammed V University in Rabat}, \orgname{Ecole Mohammadia d\textsc{\char13}Ing\'enieurs (EMI)}, \orgaddress{\city{Rabat}, \country{Morocco}}}


\abstract{Random Forest (RF) is well-known as an efficient ensemble learning method in terms of predictive performance. It is also considered a ``black box'' because of its hundreds of deep decision trees. This lack of interpretability can be a real drawback for acceptance of RF models in several real-world applications, especially those affecting one’s lives, such as in healthcare, security, and law. 
\\In this work, we present Forest-ORE, a method that makes RF interpretable via an optimized rule ensemble (ORE) for local and global interpretation. Unlike other rule-based approaches aiming at interpreting the RF model, this method simultaneously considers several parameters that influence the choice of an interpretable rule ensemble. Existing methods often prioritize predictive performance over interpretability coverage and do not provide information about existing overlaps or interactions between rules. Forest-ORE uses a mixed-integer optimization program to build an ORE that considers the trade-off between predictive performance, interpretability coverage, and model size (size of the rule ensemble, rule lengths, and rule overlaps). In addition to providing an ORE competitive in predictive performance with RF, this method enriches the ORE through other rules that afford complementary information. It also enables monitoring of the rule selection process and delivers various metrics that can be used to generate a graphical representation of the final model.
\\This framework is illustrated through an example, and its robustness is assessed through 36 benchmark datasets. A comparative analysis of well-known methods shows that Forest-ORE provides an excellent trade-off between predictive performance, interpretability coverage, and model size.
}




\maketitle

\section{Introduction}
\label{intro}
\begin{figure*}[htb]
\centering
\includegraphics[width=1\textwidth]{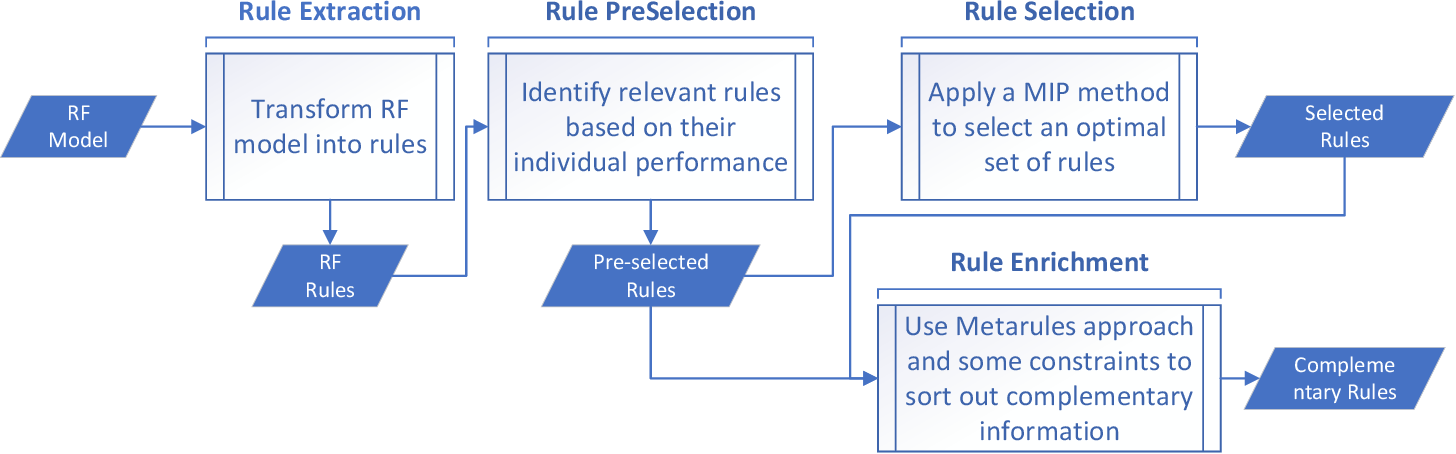}
\caption{Forest-ORE framework for interpreting RF}
\label{fig:approach} 
\end{figure*}

Machine learning (ML) interpretability is required in several real-world applications \citep{ carrizosaMathematicalOptimizationClassification2021, dasTaxonomySurveyInterpretable2020}, such as in healthcare, law, and security, because of different aspects \citep{haddouchiAssessingInterpretationCapacity2018a}. The first aspect is related to trust in ML models. A good prediction performance is not sufficient to make a model trustworthy. To be accepted and deployed, the model should be sufficiently proven accurate via intelligible explanation. The second aspect concerns the need to take action based on the ML model. Indeed, the wide use of ML models by domain experts is mainly owing to the models’ ability to uncover new knowledge. This knowledge should be interpretable so that a model can be analyzed, approved, and refined in a decision-making system. Another aspect is the consideration of regulatory constraints. ML interpretability is a serious concern in regulated fields \citep{goodmanEuropeanUnionRegulations2017} and applications affecting people's lives. In such fields of application, decisions made by a model have to be highly interpretable so that they conform with regulations and provide explanations in the case of individuals complaints \citep{blanco-justiciaMachineLearningExplainability2020}.
\\Random Forest \citep{breimanRandomForests2001} is one of 
the most performant predictive models used today \citep{lundbergLocalExplanationsGlobal2020}. Its success is due to the diversity of its collection of trees, which makes it robust against overfitting \citep{breimanCLASSIFICATIONREGRESSIONTREES1884,sagiEnsembleLearningSurvey2018a}. 
In addition, the RF building process is considered an intelligible and user-friendly approach \citep{liawClassificationRegressionRandomForest2002,maissaehaddouchiSurveyMethodsTools2019}. RF is also a flexible method in the sense that it can solve several types of statistical data analysis \citep{cutlerRANDOMFORESTSCLASSIFICATION2007}, and it is suitable for tasks dealing with high-dimensional features space and small samples\citep{biauRandomForestGuided2016}. It can handle big data via parallelization as well \citep{chenParallelRandomForest2017}. However, RF produces a black box model because of its hundreds of deep decision trees. 
\\Interestingly, the interpretability of the RF model has been addressed by many researchers. Those proposing a representative rule ensemble consider it key to efficient comprehensibility and communication. The work proposed in this article adheres to this vision but differs from the other approaches in different aspects. This method uses a mixed-integer optimization program to tackle different parameters that affect the choice of an interpretable rule ensemble for RF. These parameters concern the predictive performance, the coverage, and the complexity of the final model (size of the rule ensemble, rule lengths, and rule overlaps). To the best of our knowledge, this is the first time the search for an interpretable rule ensemble for RF has considered all the parameters above simultaneously, or is solved using a mixed-integer optimization. Furthermore, besides providing an optimal rule ensemble competitive in predictive performance with RF, this method is concerned with unveiling the knowledge that can be lost in the quest for reduction and concision when forming the optimal set of rules. It also allows monitoring of the rule selection process, which can provide flexibility for posthoc analysis. Finally, this approach delivers interesting metrics that can be used to generate a graphical representation of the final model.
\\This method, Forest-ORE (ORE for Optimal Rule Ensemble), is divided into four stages. The first stage extracts the RF rule ensemble. The second reduces the rule ensemble size. It reserves the rules with good individual predictive quality based on some fixed parameter thresholds. The third stage applies a mixed-integer programming (MIP) method to build the optimal subset of rules. Finally, the fourth stage unveils other complementary information by using the metarules approach combined with some selective criteria. Figure \ref{fig:approach} illustrates these four stages.

This method is illustrated via a simulated data set, and its robustness is assessed over 36 benchmark datasets. Empirical results show that the resulting model is competitive with RF in predictive performance (measured through different metrics such as accuracy, precision, and recall) and provides a rule ensemble enabling an excellent trade-off between predictive performance and interpretability.
\\The rest of this paper is organized as follows. Section 2 reviews the relevant literature to present work associated with RF interpretability. Section 3 explains our methodology for interpreting RF. Section 4 presents experiments using this methodology through an illustrative example and a comparative analysis over benchmarking datasets. Section 5 discusses the results. Finally, Section 6 provides the conclusion.
\section{Related Work}
\label{sec:2}
We present in this section several methods and tools used to interpret RF. They are organized using a classification similar to the one adopted in \citep{ maissaehaddouchiSurveyMethodsTools2019,ariaComparisonInterpretativeProposals2021}. We also present closest works to ours that aim at optimizing discovered rules without necessarily aiming at interpreting RF models.
\subsection{Insights derived from RF internal processing}
In addition to providing prediction, the RF algorithm delivers supplementary outputs learned during its building process that help interpret its results. The most commonly used are ``variables importance plots'', ``partial dependency plots'', and ``proximity plots''. 
Variables importance plots \citep{breimanRandomForests2001} and partial dependency plots \citep{friedmanGreedyFunctionApproximation2000,breimanWALDLECTUREII2002} help users understand which features are important for predictions. Nevertheless, they do not reveal existing variable interactions and can be biased in the case of correlated variables. Proximity plots \citep{breimanWALDLECTUREII2002} on the other hand, are used to identify data clusters and outliers learned by RF. However, proximity plots frequently look similar or irrespective of the data, which raises doubts about their usefulness \citep{tibshirani_valerie_nodate}. 
Many researchers provided enhanced tools and interactive visualization interfaces based on the concepts above \citep{quachInteractiveRandomForests2012, ehrlingerGgRandomForestsVisuallyExploring2015, m.jonesEdarfExploratoryData2016b, beckettRfvizInteractiveVisualization2018, zhaoIForestInterpretingRandom2019, paluszyFacultyMathematicsInformatics2017,plonskiVisualizingRandomForest2014, golinoVisualizingRandomForest2014, tanTreeSpacePrototypes2020, parr_partial_2021}.
\subsection{Methods based on RF post processing (post hoc methods)}
Many contributions in the literature enhancing RF interpretability use post hoc methods. “Size reduction” methods aim to reduce the number of RF trees. “Rule extraction” methods 
select representative rules from RF models. Finally, “local explanation” methods provide local interpretation of RF predictions.
\subsubsection{Size reduction}
Some authors have proposed methods that reduce the size of the RF model. Authors such as \citep{latinneLimitingNumberTrees2001a, vanasscheSeeingForestTrees2008, bernardSelectionDecisionTrees2008, zhangSearchSmallestRandom2009, yangMarginOptimizationBased2012, khanEnsembleOptimalTrees2020, adnanOptimizingNumberTrees2016} have developed approaches for extracting a reduced subset of decision trees that can compete in predictive performance with a large RF model. Reducing the size would make exploring the tree paths easier; however, the final model could remain a black box model depending on the number and depth of the trees.
\subsubsection{Local explanation}
This class of methods is focused on local explanations of RF predictions. ``FeatureContribution'' \citep{palczewskaInterpretingRandomForest2013} and ``ForestFloor'' \citep{wellingForestFloorVisualizations2016} are methods determining the influence of each predictive variable on each individual predicted instance. ``ForestFloor'' also provides an interesting graphical representation that can be used to explore the prediction models decomposition in 2D-3D features space. ``LionForests'' \citep{mollas_conclusive_2022} uses unsupervised learning techniques and an enhanced similarity metric to process local interpretation of RF predictions. Other methods are developed to explain any black box model predictions locally, including RF (agnostic approaches) \citep{baehrensHowExplainIndividual2010, singhProgramsBlackBoxExplanations2016, ribeiroWhyShouldTrust2016b, guidottiLocalRuleBasedExplanations2018}.
Methods providing local explanations to a specific prediction are useful in real-world applications, especially those dealing with legal and moral constraints. Nevertheless, they generally do not allow a global overview of the RF prediction model or the discovery of patterns in the data.

\subsubsection{Rule Extraction} 
Several authors have developed frameworks extracting a reduced set of rules to approximate the tree ensemble. In most papers, authors define a rule as the combination of conditions from a root node to a leaf node in a tree. They sometimes use pruning methods to reduce the length of the rules. ``inTrees'' \citep{dengInterpretingTreeEnsembles2019b} uses a complexity-guided condition selection method to tackle the trade-off among the frequency, the error, and the length of the rules.``ExtractingRuleRF'' \citep{phungExtractingRuleRF2015a} forms a set of ranked and weighted rules based on a greedy approach. ``SIRUS'' \citep{benardSIRUSStableInterpretable2020} extract the most significant rules from a slightly modified random forest based on their probability of occurrence. ``RF+HC'' \citep{mashayekhiRuleExtractionRandom2015d} employs a hill-climbing method to build an ensemble of rules significantly reduced in size. ``defragTrees'' \citep{haraMakingTreeEnsembles2017} derives a Bayesian model selection algorithm that optimizes the surrogate model while maintaining the prediction performance. ``ForEx++'' \citep{mdnasimadnanForExNewFramework2017} builds a high-quality rule ensemble based on different averaged metrics. ``MIRCO'' \citep{birbilRuleCoveringInterpretation2020a} uses a mathematical programming approach to minimize the total impurity and the number of the selected rules. ``OptExplain'' \citep{zhangExtractingOptimalExplanations2021} extracts rules based on logical reasoning, sampling and optimization.
\\Alternatively, some authors have considered all the tree nodes as candidate rule.``RuleFit'' \citep{friedmanPredictiveLearningRule2008},``Node Harvest'' \citep{meinshausenNodeHarvest2010a}, and ``RF+SGL'' \citep{mashayekhiRuleExtractionDecision2017c} select from the RF nodes, the most important rules predicting outcomes. These three methods mainly differ in their rule selection processes and final model representations. ``Node Harvest''  gets predictions based on averaging a weighted rule list, ``RuleFit'' uses regularized linear regression to perform predictions, and ``RF+SGL'' performs predictions based on a heuristic search method and a sparse group lasso method. 
\\Rule extraction methods probably provide the most interpretable outputs compared to the other kind of methods. Their if-then semantics, built with intelligible features, are similar to natural thinking, provided that the length and the number of rules are acceptable. These methods can be very helpful in practical applications requiring interpretability, especially if they are easily applicable. For instance, the ``inTrees'' framework have been identified as very useful in many fields \citep{khalidBlackBoxTransparentComputational2015, gargettModelingInteractionSensory2015, narayananSSDFailuresDatacenters2016, miraboutalebiFattyAcidMethyl2016, eskandarianComprehensiveDataMining2017, wangDiscoveryCelltypeSpecific2018, keIntegratingGutMicrobiome2021, ghannamMachineLearningApplications2021, casiraghiExplainableMachineLearning2020}. Nevertheless, these approaches do not generally provide insights about existing overlaps or interactions between rules. They also often prioritize predictive performance over interpretability coverage. Indeed, the selecting-rule process is stopped when the performance is not improved, and as a result, the interpretability coverage is sometimes not optimized. In addition, the final rule ensemble is sometimes made of many overlapping rules, and an instance can be a member of multiple rules. This issue can distort the overall overview of the final model.
\\\\In this work, we propose interpreting the RF model via rule extraction. We consider this class of approaches key to effective natural comprehensibility and communication for local and global interpretation (as long as their size and overlaps are not very large). The rules can concisely inform users about the population prototypes, the important variables, and their relationships. They can also be used to inspect specific predictions. This work differs from other rule extraction-based approaches that interpret RF models, in several aspects. Unlike others, this method considers several parameters that influence the choice of an interpretable rule ensemble for RF simultaneously. These parameters are formalized through an MIP problem and concern the predictive performance, the interpretability coverage, and the complexity of the final model (size of the rule ensemble, rule lengths, and rule overlaps). 
\\\\There are other recent works that are related to ours in the sense that they use mathematical programming to tackle the trade-off between the predictive performance and the interpretability of a rule ensemble. These works generally use a rule generator (such as an association rule mining technique) to build the initial collection of rules and use optimization techniques to elect a set of rules that satisfies a function objective and some constraints. We report in the following some of the closest work to ours that aims at optimizing discovered rules without necessarily aiming at interpreting RF models.
\subsection{Rule learning methods aiming at the optimization of discovered rules}
Mathematical programming-based rule learning methods differ mainly in the formulation of the optimization problem, the optimization approach used, and the structure of the resulting rule ensemble.
\\\cite{dash_boolean_2018} proposed a method for learning Boolean rules through an integer program that optimally trades the classification accuracy for rule simplicity. They minimize the Hamming loss of the rule set through the objective function and bound the total complexity of the rule set through a constraint (they defined the complexity of a rule as a fixed cost of one plus the number of conditions in the rule). 
BRL \citep{letham_interpretable_2015} and its faster successor SBRL \citep{yang_scalable_2017} build bayesian rule lists (BRL) for binary classification by learning accurate probabilistic rule lists from pre-mined conditions while prioritizing lists with few rules and short conditions \citep{molnar_interpretable_nodate}. 
The learning process in the methods above does not consider the coverage of the rules and their overlaps. \\
\cite{akyuz_discovering_2021} developed a linear programming-based rule learning framework to build a reduced set of rules for multi-class classification. Their optimization problem aims to find a set of rule weights such that the sum of the total classification error (hinge loss) and the total rule cost is minimized. The weights inform about the importance of each classification rule. The cost relates to rules' attributes, such as rule length or false negatives. The coverage aspect is tackled by adding a constraint that ensures that all the instances are covered by at least one rule. However, the rule overlapping is not taken into consideration.
\cite{lakkaraju_interpretable_2016} developed a decision set learning algorithm for multi-class classification that selects from a pre-mined set of rules (using the frequent itemset mining approach \citep{agrawalFastAlgorithmsMining}), accurate, short, and non-overlapping rules that cover the whole feature space and consider small classes. Their objective function optimizes using a smooth local search algorithm, a weighted sum of seven terms that consider the trade-off between the accuracy and interpretability of the rules. The weights are non-negative tuning parameters that scale the relative influence of the different terms. In \citep{lakkaraju_interpretable_2016}, the optimization problem does not explicitly optimize the global accuracy. It uses properties (precision and recall) that encourage per-rule accuracy. 
\\\\The structure connecting the discovered rules is also an important factor that should be considered when interpreting the rules  \citep{lakkaraju_interpretable_2016,dash_boolean_2018}.
\\Ordered decision lists, such as in \citep{letham_interpretable_2015,agrawalFastAlgorithmsMining, dengInterpretingTreeEnsembles2019b}, are made of ordered rule sets where a rule applies only when none of the preceding rules apply. This ordered structure made of a list of if-then else statements can be appreciated in practical applications (such as in disease diagnostics) where the nature of human thinking follows a similar reasoning. However, this structure increases the difficulty of interpreting the rules because interpreting each new rule in the list requires understanding all the preceding rules\citep{lakkaraju_interpretable_2016}. This limitation can be a real drawback in multi-class classification if important classes are described by rules that appear further down in the list.
\\On the other hand, decision sets are collections of rules that can be considered in any order. Each instance is labeled using a majority vote over its covering rules. This structure is key to interpretability, provided that the rules are accurate and the overlap between rules is acceptable for human understandability. The coverage aspect of the collection of rules is also important, especially in a multi-class classification issue, where important classes represent minorities that can be assigned to a default rule if the coverage is not optimized. Organizing the data space based on decision sets can reveal data prototypes that can be explained by one or a small subset of rules. 
\\A user study \citep{lakkaraju_interpretable_2016} showed that humans can reason much more accurately about the decision boundaries of a decision set than those of an ordered decision list.
\\Weighted rule lists, such as in \citep{friedmanPredictiveLearningRule2008,akyuz_discovering_2021,azmi_interpretable_2019}, are organized according to a structure that assigns a weight to each rule. The weight reflects the importance of the rule's contribution to the final decision. This structure can reveal a real pattern in the decision process but can sometimes suffer from multiple overlapping rules with different weights. 
\\\\The Forest-ORE approach addresses multi-class problems. It takes into consideration both the individual predictive performance of each rule and the global predictive performance of the rule ensemble. It also considers the complexity of the rule ensemble via its size and the length of the rules. Finally, it takes into consideration the overlapping and coverage aspects during the selection of the optimal set of rules. The objective function in Forest-ORE method does not explicitly optimize the global accuracy of the rule ensemble. It optimizes the size of the rule ensemble and the quality of the individual rules (the quality of a rule is measured through its confidence, coverage, and length). Instead, Forest-ORE uses the accuracy of the initial rule set pre-mined from the RF model, to set a lower bound for the global accuracy of the final rule ensemble, as a constraint in the optimization problem. However, this parameter can be tuned by the user to investigate a better or slightly worst level of accuracy if it allows a gain in interpretability. The coverage ratio is also tackled through a constraint that we set by default to a value near to 1. The idea behind targeting an almost perfect coverage instead of a perfect coverage is that sometimes there are some noisy instances and outliers (generally a small portion of the data) that are hard to predict and explain. Trying to cover these instances can negatively influence the final model. However, this parameter is also a tuning parameter for the model. The user can experiment and assess how changing its value affects the final model. The overlapping aspect is also approached differently than in similar methods. We consider two kinds of overlaps. The first one is the overall overlap as in other similar works. The second one applies directly to each instance. We set an upper bound for the number of rules covering each instance (via an additional constraint). We consider this aspect important for issues that require inspecting individual instances. The default value is set to three. However, one can investigate other values for this parameter.
\\ We structure the final rule ensemble of Forest-ORE into an unordered decision set. We also experiment in this work an alternative version of Forest-ORE structured into an ordered rule list. 
\\In addition to providing an optimized rule ensemble competitive in predictive performance with RF, the Forest-ORE method enriches this set of rules with complementary rules that can afford supplementary knowledge. It also allows for tuning of the rule selection process, thus providing flexibility for model debugging or posthoc analysis. Lastly, this approach produces several metrics that can, in particular, be used to plot a graphical representation of the rule ensemble.

\section{The Forest-ORE method}
\label{secMethods}
In this article, we consider classification problems in which all descriptive attributes are categorical. However, we can also tackle regression issues and various attribute types via discretization \citep{garciaSurveyDiscretizationTechniques2013}. The advantage of using categorical variables is that the predictive variables space is split into predefined subspaces that can be scrutinized afterward to find filled locations that contain groups of instances. Most of the time, the studied instances are concentrated in some locations in the attribute space, which could be interesting to explore further.
In the following sections, we will present the four main blocs constituting our framework, namely: 1) Rule Extraction, 2) Rule PreSelection, 3) Rule Selection, 4) Rule Enrichment. 
\subsection{Rule Extraction}
In the first bloc of our framework, we extract all the RF rules. Each rule delimits a sub-region on the attribute space, defined by a condition $Cond$  and a target class $Y$. Algorithm ~\ref{alg:ALG1} summarizes the rule extraction process. 

\begin{algorithm}[h]
        \caption{Rule Extraction \newline \textit{Let RF represent the Random Forest model. In an RF tree, $Cond_{i\to j}$ denotes the condition from node $i$ to node $j$ and $Y_i$ denotes the target class in a node $i$. $Parent(i)$ designs the parent node of a node $i$,   $Cond_{Le}$ denotes a path condition from root node $Ro$ to leaf node $Le$, and $Y_{Le}$ designs the path target class.  DataRules represents the rule ensemble data frame.}}
        \label{alg:ALG1}
        \begin{algorithmic}
        \Require $RF$
        \Ensure  $DataRules$
 	\State  \textbf{Initialization}: $DataRules\gets null$
    		 \ForEach  { tree in $RF$}
 		\ForEach  {tree path linking root node $Ro$ to leaf node $Le$}
            	\State      $Y_{Le} \gets leaf~node~class$
            	\State       $i \gets parent(Le)$ 
		\State      $Cond_{Le} \gets Cond_{i\to Le}$
			\While { $i\not= Ro$  }
			\State {$Cond_{Le} \gets Cond_{Le} \cap  Cond_{parent(i)\to i}$  }
			\State {$i \gets parent(i)$ }
		 	\EndWhile
		\State     Add $Cond_{Le}$ and $Y_{Le}$   to $DataRules$
		 \EndFor
            \EndFor 
        \end{algorithmic}
 \end{algorithm}

\noindent One can use pruning methods to extract more concise rules, such as those related in \citep{liuIntegratingClassificationAssociation1999, bayardoConstraintbasedRuleMining1999, bayDetectingGroupDifferences2001, dengCBCAssociativeClassifier2014}. We do not tackle this issue in this article.

\subsection{Rule PreSelection}
\label{PRE RS}
One should know that the size of the rule ensemble extracted in the first stage is typically large. Since the rules extracted from RF concern different samples (bootstrap sampling), applying them to the population as a whole reveals many poorly performant rules. Due to that, we introduce a rule preselection stage to mine a reduced set of interesting rules. The purpose is to reserve only a set of rules with good individual predictive quality. 
The predictive quality is measured through 4 metrics: class coverage, confidence, number of attributes, and number of levels. Given a population of $n$ instances, and given a rule $R$ defined by a condition $Cond$ and a target class $Yclass$ as follows ``$R:Cond\Rightarrow  Yclass$", the coverage of the rule $R$  ``$cov(R)$" represents the number of instances satisfying $cond$ divided by $n$. The class coverage of the rule $R$ ``$class\_cov(R)$" represents the number of instances satisfying $cond$ divided by $n_c$ the size of the population belonging to $Yclass$. Considering this kind of coverage assists in keeping rules representing minority classes and avoids over-fitting rules for majority classes (especially when facing unbalanced data). The confidence of the rule $R$ ``$conf(R)$" represents the number of instances satisfying $cond$ and $Yclass$ divided by the number of instances satisfying $Cond$. The number of attributes ``$att\_nbr(R)$" represents the number of variables used in $cond$, and the number of levels ``$lev\_nbr(R)$"  represents the number of modalities used in $cond$. As example, the number of attributes for the following rule ``$A \in \{A3,A4\} ~~\&~~B \in \{B3,B4\} \Rightarrow Y=C1$" equals 2 and its number of levels equals 4.
We also use Jaccard distance to measure the similarity between each pair of rules. The Jaccard index is a well-known technique used to measure the similarity between two sets \citep{fletcherComparingSetsPatterns2018}, and is defined as the size of the intersection divided by the size of the union of the two sets.
Then, to filter the weak rules from the RF rule ensemble, we proceed as summarized in Algorithm \ref{alg:Preselection}.
\\\\The output of the ``Preselected Rules'' stage is constituted of: 
\begin{itemize}
\item $RuleMetrics$: data frame of the preselected rule metrics in the format illustrated in Appendix ~\ref{appendix:PSo}. The types of information provided for each rule are the rule $id$, its confidence ($Conf.$), its coverage ($Cov.$), the number of attributes used ($Att.~nbr$), the number of the levels used ($Lev.~nbr$), the number of attributes scaled to a value between 0 and 1 ($Att.~nbr\_S$), the number of the levels scaled to a value between 0 and 1 ($Lev.~nbr\_S$), the variables used in the condition ($Attributes$), and the predicted target ($Ypred$).
\item$CovOk$: data frame ($n$ rows and $m$ columns) providing the correct coverage of the preselected rules. It is a binary data frame where $CovOk[i,j] = 1$ if rule $j$ covers row $i$ and predicts it correctly, and 0 otherwise.
\item$CovNok$: data frame providing the incorrect coverage of the preselected rules. It is a binary data frame where $CovNok[i,j] = 1$ if rule $j$ covers row $i$ and does not predict it correctly, and 0 otherwise.
\\$CovOk$ and $CovNok$ are computed by comparing the rule predictions and the target values in the training data set. Their format is illustrated in Appendix ~\ref{appendix:PSo}.
\\
\end{itemize}
\begin{algorithm}[h]
\caption{Rule Preselection \newline \textit{Let $RFR$ denote RF rules and $Data$ the training dataset. Let $min\_conf$ and $min\_class\_cov$ denote the lower limits  for rules confidence and class coverage. Let $max\_len$ and  $max\_simil$ denote the upper limits for rule length and similarity. Let $PSR$ denote the resulting Preselected Rules, and $PSRS$ denote similar rules removed. }}
\label{alg:Preselection}
\begin{algorithmic}
\Require $RFR, Data, min\_conf , min\_class\_cov, max\_len,$ $max\_simil$
\Ensure $PSR, PSRS$
\State \textbf{Initialization:} $PSR \gets RFR$ and $PSRS \gets null$
\State $RR\gets redund(PSR)$ ~\Comment{$redund(D)$: function extracting redundant rules in a set $D$}
\State $PSR\gets PSR - RR$
\State $PSR\gets\{R \in PSR \mid len\left(R\right) \leq max\_len \}$ 
 \ForEach {$R \in PSR$ }
\State  Compute $~~R_{conf}=conf(R)$ and $R_{ class\_cov}=class\_cov(R)$
\EndFor 
\State  $PSR\gets\{R \in PSR \mid R_{ class\_cov} \geq min\_class\_cov~and~R_{conf} \geq min\_conf  \}$ 
\State Compute $Mat_{simil}$, the $k*k$ matrix of rules' pairwise similarity 
\ForEach {$row~~ i~~in ~~Mat_{simil}$ }
\State $S_{simil_{i}} \gets\{R \in PSR \mid Mat_{simil}\left[i,j\right] \geq max\_simil  , ~~j=1...k\}$ 
\EndFor
\State $S\gets \{S_{simil_{i}}: i =1...k\}$ 
\ForEach {$S_{simil} \in S$}
\State $Best_{conf} \gets \{argmax(conf(R)) \mid R \in S_{simil} \}$
\State $Best_{cov} \gets \{argmax(cov(R)) \mid R \in Best_{conf} \}$
\State $Best_{att} \gets \{argmin(att\_nbr(R)) \mid R \in Best_{cov} \}$
\State $ R_{best} \gets \{argmin(lev\_nbr(R)) \mid R \in Best_{att} \}$
\State  $PSRS\gets PSRS\cup \{S_{simil}-\{R_{best}\}\}$
\State  $PSR\gets PSR- \{S_{simil}-\{R_{best}\}\}$
\EndFor
\end{algorithmic}
\end{algorithm}
 \subsection{Rule Selection}
\label{RS}
Once the preselected rules are extracted, we apply an optimization method to form an optimal collection of rules.
\subsubsection{Problem Description}
This optimization problem involves determining the optimal set of rules for interpreting the RF model while considering diverse individual and collaborative factors: predictive performance, coverage, and complexity (size of the rule ensemble, rule lengths, and rule overlaps).
The objective is to build a set of rules that cover our population in an intelligible way and reserve a predictive performance comparable to RF performance.
We tackle this problem by setting an ensemble of objectives and constraints:
\begin{enumerate}
\item Minimize the size of the final rule ensemble. The final rule ensemble should cluster our population into subgroups, and one or more rules should explain each one. Minimizing the size of the final ensemble is then essential to ensure an easy understanding of this clustering.
\item Maximize the individual contribution of each rule in the quality of the ensemble. Choosing rules with high coverage contributes to minimizing the number of rules and avoiding overfitting. Moreover, choosing rules with high confidence contributes to raising the confidence in the representativity of each subgroup. It also increases the predictive accuracy of the final rule ensemble.
\item Minimize the complexity of the final rule ensemble. This complexity is measured in terms of the number of variables and levels used in each rule. If two rules have similar coverage and confidence, the rule with a smaller number of variables would be preferred as being simpler to interpret. The rule with the smaller number of levels would also be preferred as the most concise. It is a kind of pruning.
\item Reserve a predictive performance comparable to RF predictive performance.
\item Maximize the coverage of the final rule ensemble. We are interested in maximizing the rate of the population concerned with interpretability. The remaining data not covered by the final rule ensemble will be mapped to a default rule representing its majority target class.
\item Minimize the rule overlaps to avoid blurring the overall overview of the final ensemble. Each member of the population should not belong to more than a fixed number of rules. In addition, it is interesting to limit the overall overlap. \\
\end{enumerate}
We have formulated this problem as a mixed-integer programming (MIP) model. The first three points above were expressed in the objective function. The remaining points were addressed as constraints.
This model was implemented in the Gurobi Python API, and then solved using the Gurobi Optimizer (Free academic license). This optimization problem is described in the following section.

\subsubsection{Problem Formulation}
\textbf{Input data}\\
\begin{itemize}[label={},leftmargin=0.4cm]
\item Let $RuleMetrics$ represent the preselected rule data frame, where lines refer to the preselected rules and columns their attributes. 
Let $CovOk$ and $CovNok$ be the data frames of correct and incorrect coverage. Finally, Let $init\_error$ be the RF prediction error in the training data.\\
\end{itemize}
\textbf{Sets and Indices}\\
\begin{itemize}[label={},leftmargin=0.4cm]
\item \textbf{\textit{i}} $\in I=\{1,2,...,n\}$: Index of instances.
\item \textbf{\textit{j}} $\in J=\{1,2,...,m\}$: Index of preselected rules.\\
\end{itemize}
\textbf{Parameters}\\
\begin{itemize}[label={},leftmargin=0.4cm]
\item \textbf{\textit{confidence[j]}} $\in \left[T_{conf},1\right]$: confidence of the rule $j$. $T_{conf}>$ 0 : rule confidence threshold.
\item \textbf{\textit{coverage[j]}} $\in \left[T_{cov} ,1\right]$: coverage of the rule $j$. $T_{cov}>$ 0 : rule coverage threshold.
\item \textbf{\textit{att\_ratio[j]}} $\in \left(0 ,1\right]$: the size of the attributes used in rule $j$  (scaled).
\item \textbf{\textit{levels\_ratio[j]}} $\in \left(0 ,1\right]$: the size of the modalities used in rule $j$  (scaled).
\item \textbf{\textit{CovOk[i,j]}} $\in \{0 ,1\}$: correct coverage. $CovOk\left[i,j\right] = 1$ if rule $j$ covers instance $i$ and predicts it correctly, and 0 otherwise.
\item \textbf{\textit{CovNok[i,j]}} $\in \{0 ,1\}$: incorrect coverage. $ CovNok\left[i,j\right] = 1$ if rule $j$ covers instance $i$ and does not predict it correctly, and 0 otherwise.
\item \textbf{\textit{init\_error}}: RF error in prediction. 
\item \textbf{\textit{\boldmath{$w_0, w_1, w_2, w_3$}}} $\in \left[0 ,1\right]$: weights used in the objective function. Default parameters used are 1, 1, 0.1, and 0.05, respectively.
\item \textbf{\textit{maxcover}} $\in \{1,2,\dots,10\}$: the upper bound for the number of rules to which an instance can belong to. The default parameter is 3.
\item \textbf{\textit{maxoverlap}} $\in \left[0,1\right]$: the upper bound for the overall overlap ratio (ratio of instances bellonging to 2 rules or more). The default parameter is 0.5.
\item \textbf{\textit{alpha}} $\in \left[0,1\right]$: the upper bound for the loss in overall accuracy compared to the RF accuracy. The default parameter is 0.01. 
\item \textbf{\textit{beta}} $\in \left[0,1\right]$: the upper bound for the loss in overall coverage compared to the coverage of the preselected rules (initial coverage equals 1). The default parameter is 0.05.
\item \textbf{\textit{n}} $\in \mathbb{N}$:  the size of the population.
\item \textbf{\textit{m}} $\in \mathbb{N} $:  the size of the preselected rules.\\
 \end{itemize}
\textbf{Decision Variables}\\
\begin{itemize}[label={},leftmargin=0.4cm]
\item \textbf{\textit{is\_selected[j]}} $\in \{0 ,1\}$:  takes value 1 if we select rule $j$, and 0 otherwise.
\item \textbf{\textit{is\_covered[i]}}  $\in \{0 ,1\}$:  takes value 1 if instance $i$ is covered by at least one rule, and 0 otherwise.
\item \textbf{\textit{is\_error[i]}} $\in \{0 ,1\}$: takes value 0 if instance $i$ is correctely predicted ( if the sum of rules that predict it correctly is strictly greater than the sum of rules that mispredict it), and 1 otherwise.
\item \textbf{\textit{is\_overlap[i]}} $\in \{0 ,1\}$: takes value 1 if instance $i$ belongs to 2 or more rules, and 0 otherwise.\\
 \end{itemize}
\textbf{Objective Function}\\
\\Minimize the weighted trade-off between the number of rules and their quality.
\begin{center}
\newmaketag
\begin{small}
\begin{equation}
\label{Obj.Func.}
\begin{aligned}
\textit{Min.:} &\sum_{j \in J} is\_selected\bigl[j\bigr]\times \biggl( 1 +w_0\times \bigl(1-confidence\bigl[j\bigr]\bigr)  \\ &+  w_1\times \bigl(1-coverage\bigl[j\bigr]\bigr)  + w_2\times att\_ratio\bigl[j\bigr] \\ &+ w_3\times levels\_ratio\bigl[j\bigr] \biggr)
\end{aligned}
\end{equation}
\end{small}
\end{center}
The first component of the objective function minimizes the size of the rule ensemble that will organize and cluster the data space.  The second term minimizes the cumulative error in prediction of the rule ensemble. This term encourages rules with high confidence. This aspect is important because each rule is intended to represent and explain a cluster in the data. Its importance is monitored through the weight  $w_0$. Setting a high value to  $w_0$  is expected to guarantee the generation of trustworthy rules that final users could accept. 
The third term maximizes the cumulative coverage of the rule ensemble. This term encourages rules with high coverage and is monitored through the weight $w_1$. Setting a high value to $w_1$ is expected to minimize the number of rules and avoid overfitting rules that the second term of the objective function could prioritize. The fourth term minimizes the cumulative rule lengths, and the fifth one minimizes the cumulative sum of levels used in the rules. These two terms encourage concise rules with few variables and few levels. 
In the default setting, we give equal importance to the three first components ($weight$ = 1) and less importance to the remaining components ($w_2 = 0.1, w_3 = 0.05$). Doing so, we expect that the solution to the optimization problem will be essentially guided by the three first components and that the remaining two components will serve to refine the choice of the rules. Given two rules with similar confidence and coverage, the rule with fewer variables will be chosen. Similarly, if two rules have similar confidence, coverage, and length, the rule with fewer levels will be chosen.
\\\\
\textbf{Constraints}
\\\\Let $P_i$  be:
\newmaketag
\begin{small}
\begin{equation}
\sum_{j\in J} is\_selected\left[j\right]\times\left(CovOk\left[i,j\right]-CovNok\left[i,j\right]\right)
\end{equation}
\end{small}  
\\and let $C_i$  be:
\newmaketag
\begin{small}
\begin{equation}
\sum_{j\in J} is\_selected\left[j\right]\times\left(CovOk\left[i,j\right]+CovNok\left[i,j\right]\right)
\end{equation}
\end{small}
\\
\begin{itemize}[label={}]
\item \textbf{Maxcover constraint}: \\Each item is covered by at most $maxcover$ rules.\\
 \end{itemize}
\newmaketag
\begin{small}
\begin{equation}
\label{C2}
C_i\le maxcover\ \ \ \ \ \ \forall\ i\in I
\vspace{2ex}
\end{equation}
\end{small}

\begin{itemize}[label={}]
\item \textbf{Error constraints}: \\The selected rules error in prediction does not exceed the initial error + $alpha$ (see Proof 1 in Appendix \ref{appendix:MIP}).\\
 \end{itemize}
\newmaketag
\begin{small}
\begin{equation}
\label{C3}
P_i\le maxcover\times\left(1-is\_error\left[i\right]\right)\forall i\in I
\end{equation}
\end{small}

\newmaketag
\begin{small}
\begin{equation}
\label{C4}
P_i\geq 1-is\_error\left[i\right]\times \left(1+maxcover\right)\forall i\in I
\end{equation}
\end{small}

\newmaketag
\begin{small}
\begin{equation}
\label{C5}
\begin{aligned}
\sum_{i\in I}{is\_error\left[i\right]} - \left(1-is\_covered\left[i\right]\right)\le&\\ \left( init\_error +alpha\right)\times\sum_{i\in I}{is\_covered\left[i\right]}
\end{aligned}
\vspace{2ex}
\end{equation}
\end{small}

\begin{itemize}[label={}]
\item \textbf{Min coverage constraints}: \\The selected rules cover at least $100\times(1-$beta$)\%$ of the population (see Proof 2 in Appendix \ref{appendix:MIP}).\\
 \end{itemize}
\newmaketag
\begin{small}
\begin{equation}
\label{C6}
C_i\le maxcover\times is\_covered\left[i\right]\forall i\in I
\end{equation}
\end{small}

\newmaketag
\begin{small}
\begin{equation}
\label{C7}
C_i\geq is\_covered\left[i\right]\forall i\in I
\end{equation}
\end{small}

\newmaketag
\begin{small}
\begin{equation}
\label{C8}
\sum_{i\in I}{is\_covered\left[i\right] \geq n\times \left(1-beta\right)}
\vspace{2ex}
\end{equation}
\end{small}

\begin{itemize}[label={}]
\item \textbf{Max overall overlap constraints}: \\The rate of items belonging to more than 2 rules does not exceed $maxoverlap$ (see Proof 3 in Appendix \ref{appendix:MIP}).\\
 \end{itemize}
\newmaketag
\begin{small}
\begin{equation}
\label{C9}
C_i\le1-is\_overlap\left[i\right] \times \left(1 - maxcover\right)\forall i\in I
\end{equation}
\end{small}

\newmaketag
\begin{small}
\begin{equation}
\label{C10}
C_i\geq2\times is\_overlap\left[i\right]\forall i\in I
\end{equation}
\end{small}

\newmaketag
\begin{small}
\begin{equation}
\label{C11}
\sum_{i\in I}{is\_overlap\left[i\right]} \le maxoverlap\times\sum_{i\in I}{is\_covered\left[i\right]}
\end{equation}
\end{small}
\\ As mentioned before, the objective function in Forest-ORE method does not optimize the global accuracy but only the quality of the individual rules. Instead, we use the accuracy of the initial pre-mined set of rules from RF to set a lower bound for the accuracy to target for the final rule ensemble (Error constraints). The objective is to maintain a predictive performance comparable to RF predictive performance. In addition, we set the lower bound for the coverage ratio (Min coverage constraints) to a value near to one to guarantee that the targeted accuracy concerns the entire data. Doing so does not guarantee that we obtain the optimal value for the predictive performance. Instead, we tolerate an amount of loss in performance to balance between accuracy and interpretability. This being said, the constraints related to accuracy and coverage can be tuned by the users to investigate a better or slightly worst level of performance if it allows a gain in interpretability. 
\\The overlapping constraints consider two kinds of overlaps. The first one is the overall overlap as in other similar works (Max overall overlap constraints). The second one applies directly to each instance (Max cover constraint). It fixes an upper bound for the number of overlapping rules for each instance. We consider that this aspect is important for issues that require inspecting individual instances. The default value for this upper bound is set to three. However, one can investigate other values for this parameter.
\\\\By formulating these objectives and constraints as an optimization problem, we will generate an ensemble of rules optimally matching our setting. However, there will be many competitive rules that can provide additional information. The ``Rule Enrichment'' stage (section \ref{RE}) will reveal these complementary rules. The rule enrichment is a facultative stage that aims to add new information to the built rule ensemble. This stage can be useful for users because it allows for interpreting the finding from different facets.

\subsection{Rule Enrichment}
\label{RE}
Since we noted that there were many competitive rules at the ``Rule PreSelection'' and ``Rule Selection'' stages, sometimes with entirely different sets of descriptive variables, we thought it would be interesting to shed light on those that could add complementary information. The purpose of doing so is to reveal additional rules that are applied to each selected rule subspace. 
\newline An interesting methodology finding such rules relationships is the Metarules approach proposed in \citep{berradoUsingMetarulesOrganize2007}. It consists of organizing and grouping rules by exploring their mutual relationship and containment. This methodology allows for the discovery of a collection of independent rules' subsets. Each subset forms a cluster of rules that could be summarized based on a graphical representation and analyst preferences. Metarules uses the Association Rules Mining (ARM) approach to find these relationships. 
\newline \\Let $MetaR$ represent the $nxm$ rule matrix where each line refers to an instance, and each column refers to a rule from the collection of preselected rules. This matrix links, for each instance, the rules covering it, as illustrated in table \ref{tab:meta}. Line 1, for example, means that the conditions of the rules $R_1$, $R_2$, and $R_4$ are applied to instance 1. In the Metarules methodology, each line from $MetaR$ is mapped to a transaction where each rule is considered an item. As example, line 1 from table ~\ref{tab:meta} is mapped to the transaction: $\{R_1,R_2,R_4\}$. The ARM approach is then applied to the $n$ transactions in order to find the one-way association rules. The one-way association rules takes the format $R_i\mathrm{\to}R_j$ and is called a metarule. 
The quality of the containment in the Metarules approach is monitored through the ARM confidence and support. The support of a metarule is computed by dividing the number of instances satisfying $R_i$ and $R_j$ by the total number of instances. The confidence is computed by dividing the number of instances satisfying $R_i$ and $R_j$ by the number of instances where $R_i$ is applied. To ensure a quasi-total containment, we fix the ARM minimum confidence to a value near to 1. Accordingly, we guarantee that the $R_i$ is not a generic rule applied to other regions different from the one delimited by the rule $j$. As for the support, it is recommended to set it to a value that avoids over-fitting.
\\It should be noted that the Metarules approach leads, in general, to the discovery of a large number of metarules unveiling containments between the rules. In our work, we use the metarules approach combined with other constraints to select a reduced number of metarules. We first extract rules interacting with the selected rules via the metarules approach. We then select the ones providing new information to each selected rule. Since each rule $R_j$ is a combination of (variable, values) pairs that defines a subregion of the attribute space, the idea is to search rules defining the same subregion and using a set of variables different from the ones used in the rule $R_j$. From these rules, we choose the best ones based on the rate of their intersections with the rule $R_j$, confidence values, coverage values, and the number of attributes used. We define the intersection between the rules $R_i$ and $R_j$ ``$intersect(R_i,R_j)$" as the size of the set of instances covered by the rules $R_i$ and $R_j$ divided by the size the set of instances covered by the rule $R_j$. 

\begin{table}[htb]
  \centering
\caption{Simplified illustration of the Metarule matrix. Rows represent data instances, and columns represent rules.}
\resizebox{0.5\textwidth}{!}{%
     \begin{tabular}{rcccccrc}
    \toprule
          & \textbf{R\textsubscript{1}} & \textbf{R\textsubscript{2}} & \textbf{R\textsubscript{3}} & \textbf{R\textsubscript{4}} & \textbf{R\textsubscript{5}} & \multicolumn{1}{c}{\textbf{R\textsubscript{6}}} & \textbf{R\textsubscript{7}} \\
    \midrule
    \textbf{1} & $R_1$    & $R_2$    &       & $R_4$    &       & \multicolumn{1}{c}{} &  \\
    \textbf{2} &       & $R_2$    &       & $R_4$    &       & \multicolumn{1}{c}{} &  \\
    \textbf{3} &       & $R_2$    & $R_3$    &       & $R_5$    & \multicolumn{1}{c}{} &  \\
         \bottomrule
    \end{tabular}%
}
   \label{tab:meta}%
\end{table}%
\noindent Algorithm ~\ref{alg:enrichment} describes the “Rule enrichment” steps. The dataframe of complementary rules takes the format illustrated in Appendix ~\ref{appendix:RE}.  
\begin{algorithm}[htbp]
        \caption{Rule Enrichment \newline \textit{ Let $PSR$ denote the Preselected rules, $PSRS$ the similar rules removed (ref Algorithm \ref{alg:Preselection}), $SR$ the Selected Rules, and $Data$ the training dataset. Let $CR$ denote the set of complementary rules that will be returned. Let fix minimum confidence and minimum support for association rules mining ($arm\_minconf , arm\_minsup$). }}
        \label{alg:enrichment}
        \begin{algorithmic}
        \Require $PSR, PSRS, SR, arm\_minconf , arm\_minsup$ 
        \Ensure $ CR$
       \State  \textbf{Initialization:} $CR \gets null, PSR \gets PSR \cup PSRS$
       \State Compute $MetaR$, the Metarules matrix ~\Comment{rows represent instances, and columns represent $PSR$ rules (see Table \ref{tab:meta})} 
        \State Convert $MetaR$ to transactions $Meta_{trans}$
       \State Apply association rule mining to $Meta_{trans}$ to discover the rules applied to the subspaces covered by SR. Each discovered metarule is constrained to be an expression of the form $ R_i\mathrm{\to}R_j$ where~~$R_i \in  PSR$ and  $R_j \in SR$. 
        \ForEach {$R_j \in $ SR}
		\State Extract $Att_{R_j}$ the list of attibutes used in $R_j$ 
		\State Extract $Meta_{R_j}$ the set of $R_j$ metarules 
		\State $RM \gets \{R \in Meta_{R_j} \mid Att_{R} = Att_{R_j} \}$
            	\State $Meta_{R_j} \gets Meta_{R_j} - RM$
		\State Extract $U_{att} = unique( \{Att_{R} \mid R \in Meta_{R_j} \})$
            	\ForEach {$Att \in U_{att}$} 
		\State $Rules_{Att} \gets \{R \in Meta_{R_j} \mid Att_{R} = Att \}$
		\State $Best_{intersect} \gets \{argmax(intersect(R,R_j)) \mid R \in Rules_{Att} \}$
		\State $Best_{conf} \gets \{argmax(conf(R)) \mid R \in Best_{intersect} \}$
		\State $Best_{cov} \gets \{argmax(cov(R)) \mid R \in Best_{conf} \}$
		\State $R_s \gets \{argmin(att\_nbr(R)) \mid R \in Best_{cov} \}$
		\State $CR\gets CR\cup \{R_s\}$ 
            	 \EndFor
      \EndFor    		
        \end{algorithmic}
 \end{algorithm}

\noindent \\To test our methodology, we will at first illustrate it through its application to a simulated XOR dataset, and then show its effectiveness on several benchmark datasets.

\section{Experiments}
\label{secResults}
In this section, we present the performance of Forest-ORE compared to RF, RPART (Recursive Partitioning and Regression Trees) \citep{breimanCLASSIFICATIONREGRESSIONTREES1884}, STEL (Simplified Tree Ensemble Learner) \citep{dengInterpretingTreeEnsembles2019b}, RIPPER (Repeated Incremental Pruning to Produce Error Reduction) \citep{cohen_fast_1995}, and SBRL (Scalable Bayesian Rule Lists) \citep{yang_scalable_2017}. We first describe our validation procedure. Then, we apply this procedure to an illustrative example. Finally, we compare the performance of the methods mentioned above over 36 benchmarking datasets. The implementation and the computational work are done using the R language and environment for statistical computing \citep{rcoreteamLanguageEnvironmentStatistical2019}, the Python programming language \citep{pythoncoreteamPythonDynamicOpen2019}, and Gurobi Optimizer Software (with free academic licence) \citep{gurobioptimizationllcGurobiOptimizerReference2021}. The code, the data files, and the resulting files for the benchmark reported in this paper are available via GitHub (refer to Section ``Declarations").

\subsection{Experimental Set Up}
\label{secProcedure}
In this study, we compare Forest-ORE to RF as a baseline for the predictive performance. We use RPART because it is one of the most well-known interpretable methods. Furthermore, we have chosen STEL because of its popularity, in recent years, for solving practical issues requiring interpretability of RF (section \ref{sec:2}). In \citep{dengInterpretingTreeEnsembles2019b}, the author proposed two outputs for interpreting RF models. The first output is a reduced set of rules obtained by applying a complexity-guided regularized random forest method to RF rules (GRRFR). The second output is an ordered list of rules formed by applying a greedy algorithm to GRRFR. This ordered list is called the Simplified Tree Ensemble Learner (STEL).The prediction of an instance using STEL is made based on the first ordered rule whose condition is applied to the instance. We also included in the comparative study the following well-known rule-learning algorithms for classification. We compare the Forest-ORE algorithm to the SBRL method, which is known in recent years, for producing very condensed rule sets. We also compare it to RIPPER, a state-of-the-art rule learning method. The two algorithms produce ordered rule lists. 
\\Concerning Forest-ORE, we present the performance of the optimal set of rules (section \ref{RS}), referred to as Forest-ORE, and the performance of the preselected rules (section \ref{PRE RS}), referred to as Pre-Forest-ORE. The prediction in these two cases is made based on rules majority voting. The data not covered by the ORE is predicted using a default class label. For our experiments, we compute the default class label based on the training data used to build the ORE. If the ORE covers the entire data, we assign the majority class label in the data to the default class label. Otherwise, we assign the majority class label in the remaining data not covered by the ORE to the default class label. However, other choices of default class labels can be easily implemented, such as those reported in \citep{lakkaraju_interpretable_2016}. We also present the performance of a combination between Forest-ORE and STEL (referred to as Forest-ORE + STEL), in which we apply the greedy algorithm used to form STEL to Forest-ORE rules. The result is an ordered list of rules.  
\\\\We compare these classifiers based on their predictive performance, their interpretability coverage, and the complexity of their resulting models. The interpretability coverage corresponds to the coverage of the rule ensemble on the testing sets, and the complexity is measured through the model size (total number of rules), and the length of the rules.
\\The predictive performance is assessed through accuracy, macro precision, macro recall, and Cohen’s kappa measures. Accuracy (the ratio of correctly predicted instances to the total instances) is the most widely used measure to compare classifier performance but is not sufficient in the case of imbalanced data \citep{haiboheLearningImbalancedData2009}. Macro precision (known as a measure of exactness) and macro recall (known as a measure of completeness) inform about how well the classifiers perform regarding each class. Cohen’s Kappa \citep{cohenCoefficientAgreementNominal1960b} is a statistical measure used to compare multi-class and imbalanced class data. It is known as a measure of reliability. It informs about how well a classifier is performing compared to the performance of a classifier that simply guesses at random according to the frequency of each class. Cohen’s kappa ranges from -1 to 1. According to Landis and Koch \citep{landisMeasurementObserverAgreement1977}, values less than 0 indicate that the classifier is useless, while values ranging between 0 and 0.20 qualify its usefulness as slight, those between 0.21 and 0.40 as fair, those between 0.41 and 0.60 as moderate, those between 0.61 and 0.80 as substantial, and those between 0.81 and 1 as almost perfect. 
\\We also use the fidelity metric to assess how well the explanations provided by the methods that interpret RF approximates the predictions of the RF model. We measure the fidelity on all instances, instances correctly predicted by RF, and instances incorrectly predicted by RF. 
\\In addition, we use the Friedman test to compare the different algorithms over the benchmark datasets. We rank all the algorithms on each data set and prediction metric and use the mean ranks in the Friedman test to reject the null hypothesis of no difference among the algorithms \citep{ benavoli_should_2016}.  We then use the Wilcoxon statistical signed-rank test \citep{wilcoxonIndividualComparisonsRanking1945} with a level of significance equal to 0.05, as post-hoc test, to establish pairwise significant differences. We perform pairwise comparisons of the prediction metrics between each pair of classifiers. The results are then summarized by counting the times each classifier outperforms, ties, and underperforms compared to the other discretizers. 
\\\\We have set Forest-ORE's default parameters to the following. Those for the rule preselection stage are  rule minimum class coverage = 0.025, minimum confidence = 0.51, maximum number of descriptive variables used in each rule = 6, and rule similarity threshold = 0.95. Those for the optimization stage are $MaxCover$ = 3, $MaxOverlap$ = 0.5, $Alpha$= 0.01, and $Beta$= 0.025. The other classifiers have been run using their default parameters. The number of trees for RF has been set to 100.
\\We have used a 10-fold Monte Carlo cross-validation procedure \citep{dubitzkyFundamentalsDataMining2007}. This procedure creates 10 random splits of the dataset into training (70\%) and testing sets (30\%). The relative ratio of the target classes is respected during the splitting process. At each splitting round, we fit the classifiers to the training set and compute the metrics previously described over the training and testing sets.

\subsection{Illustrative example}
\label{example}
\begin{figure}[htbp]
\centering
  \includegraphics[width=0.5\textwidth]{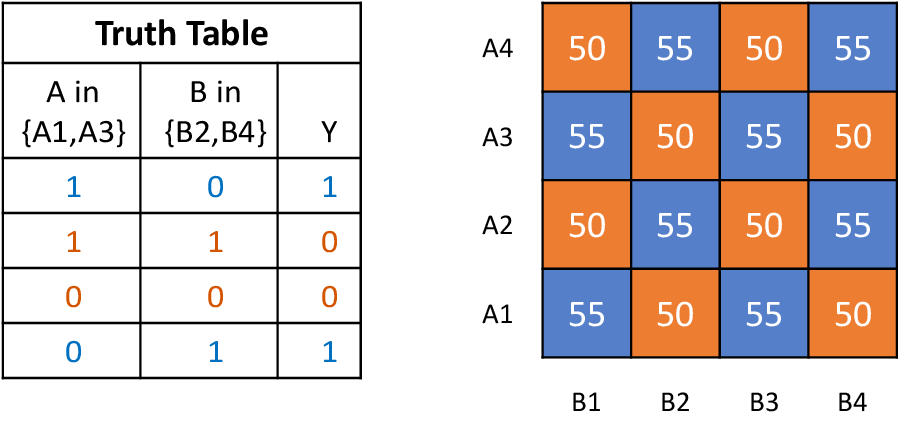}
\caption{XOR dataset: On the left: XOR truth table. On the right: the number in each box refers to the number of instances respecting the condition defined by the box and the color refers to the box target class.}
\vspace{-2ex}
\label{fig:XOR_data}       
\end{figure}

Here, we illustrate the Forest-ORE processing through an example. We have simulated the XOR dataset presented in figure \ref{fig:XOR_data} as follows. This dataset contains 840 instances, 2 categorical descriptive attributes $A=\{A1,A2,A3,A4\}$ and $B=\{B1,B2,B3,B4\}$, and two classes $Y=\{0,1\}$ (blue and orange). To make things more complex, we have added a third descriptive attribute $C=\{C1,C2\}$ defined by the following constraint: If $A \in \{A3,A4\} ~\&~ B \in \{B3,B4\}$ then $C=C1$, Else $C=C2$. $C$ does not play a role in predicting the target class; nevertheless, adding this kind of relationship between attributes alters the accuracy of RF (RF error rate equals on average 0 before adding $C$ but equals, on average, 10\% after adding $C$). 

\noindent \\Table \ref{tab:XORperf} reports the mean and standard error of different metrics on the classification of the XOR dataset. These metrics concern the testing sets. Tables \ref{tab:SRForest_ORE}, \ref{tab:SRSTEL},  \ref{tab:SRCART}, \ref{tab:SRRIPPER}, and \ref{tab:SRSBRL} report the rule ensembles resulting from applying Forest-ORE, STEL, RPART, RIPPER, and SBRL methods. As shown in table ~\ref{tab:XORperf}, the accuracy of the classifiers is 100\%, except the accuracy of the RF model, which is 90\% on average (RF predictive performance was altered after adding variable $C$).
\\The difference between Forest-ORE and the other classifiers with similar accuracy concerns the size, the coverage, and the length of the rules. Forest-ORE has covered the entire data space (coverage 100\%) using four rules (Table \ref{tab:SRForest_ORE}). STEL has selected two significant rules that cover around 49\% of the data, representing class $“1”$ (Table \ref{tab:SRSTEL}), and a default rule for the class $“0”$. RIPPER selected eight rules that cover around 48\% of the data space, where each rule represents a small region of the class $“1”$ (see Table \ref{tab:SRRIPPER} and Figure \ref{fig:XOR_data}).  SBRL produced an ordered list of eight rules covering 75\% of the data. However, the interpretation of SBRL classification becomes somewhat difficult from the fifth rule in the list (see Table \ref{tab:SRSBRL}). 
\\The length of the rules is two on average for Forest-ORE and STEL. These methods use only the variables $A$ and $B$.  The length exceeds two on average for RPART and RIPPER because of the use of the $C$ variable, which in fact, has no role in the classification of the data. The average rule length is smaller for SBRL because it tended to use only one variable in several rules. Using ordered lists allows for reducing the number of variables used as we go further in the list. The interpretation of a rule should, however, consider all the variables used in its preceding rules. The expression of a specific rule in the list is in fact, the intersection between this rule conditions with the negation of all the preceding conditions. Thus, the rule length in an ordered list does not reflect the real length of the rule. 
\\\\The preselected stage has reduced the size of the rules by 93\% on average (from 631 for RF to 42 for Pre-Forest-ORE). This reduction stage is important for the optimization stage.
\begin{table*}[htb]
  \centering
  \caption{Performance results of classifying the XOR dataset}
\resizebox{0.9\textwidth}{!}{%
\centering
     \begin{tabular}{lrrrrrrrrrr}
    \toprule
          & \multicolumn{2}{p{5.56em}}{Total rules} & \multicolumn{2}{p{5.725em}}{Total rules per class} & \multicolumn{2}{p{4.78em}}{Rule length} & \multicolumn{2}{p{5.56em}}{Coverage} & \multicolumn{2}{p{5.56em}}{Accuracy} \\
    \midrule
    method & \multicolumn{1}{c}{Mean} & \multicolumn{1}{c}{SE} & \multicolumn{1}{c}{Mean} & \multicolumn{1}{c}{SE} & \multicolumn{1}{c}{Mean} & \multicolumn{1}{c}{SE} & \multicolumn{1}{c}{Mean} & \multicolumn{1}{c}{SE} & \multicolumn{1}{c}{Mean} & \multicolumn{1}{c}{SE} \\
    \midrule
    RF    & 631.2 & 7.31  & 315.6 & 3.65  & 2.3   & 0.01  & 1.00  & 0.00  & 0.90  & 0.03 \\
    Pre-Forest-ORE & 42.0  & 0.42  & 21.0  & 0.21  & 2.2   & 0.02  & 1.00  & 0.00  & 1.00  & 0.00 \\
    Forest-ORE & 4.0   & 0.00  & 2.0   & 0.00  & 2.0   & 0.00  & 1.00  & 0.00  & 1.00  & 0.00 \\
    Forest-ORE+STEL & 2.0   & 0.00  & 1.0   & 0.00  & 2.0   & 0.00  & 0.52  & 0.00  & 1.00  & 0.00 \\
    STEL  & 2.0   & 0.00  & 1.0   & 0.00  & 2.0   & 0.00  & 0.49  & 0.01  & 1.00  & 0.00 \\
    RPART & 10.0  & 1.30  & 5.0   & 0.65  & 2.3   & 0.14  & 1.00  & 0.00  & 1.00  & 0.00 \\
    SBRL  & 8.0   & 0.00  & 4.0   & 0.00  & 1.5   & 0.00  & 0.75  & 0.01  & 1.00  & 0.00 \\
    RIPPER & 8.0   & 0.00  & 4.0   & 0.00  & 2.1   & 0.03  & 0.48  & 0.00  & 1.00  & 0.00 \\
    \bottomrule
    \end{tabular}%
}
  \label{tab:XORperf}%
\vspace{2ex}
\end{table*}%

\begin{table*}[h]
  \centering
  \caption{Selected rules  provided by applying Forest-ORE to the XOR dataset}
\resizebox{1\textwidth}{!}{%
\centering
    \begin{tabular}{rrrrrrlll}
    \toprule
    \multicolumn{1}{l}{id} & \multicolumn{1}{l}{confidence} & \multicolumn{1}{l}{coverage} & \multicolumn{1}{l}{class\_coverage} & \multicolumn{1}{l}{att. nbr} & \multicolumn{1}{l}{lev. nbr} & cond.  & Ypred & att. \\
    \midrule
    9     & 1.00  & 0.24  & 0.50  & 2     & 4     & X[,1] in \{A1,A3\} \& X[,2] in \{B1,B3\} & `1'   & V1,V2 \\
    11    & 1.00  & 0.26  & 0.49  & 2     & 4     & X[,1] in \{A1,A3\} \& X[,2] in \{B2,B4\} & `0'   & V1,V2 \\
    26    & 1.00  & 0.27  & 0.51  & 2     & 4     & X[,1] in \{A2,A4\} \& X[,2] in \{B1,B3\} & `0'   & V1,V2 \\
    34    & 1.00  & 0.24  & 0.50  & 2     & 4     & X[,1] in \{A2,A4\} \& X[,2] in \{B2,B4\} & `1'   & V1,V2 \\
    \bottomrule
    \end{tabular}%
}
  \label{tab:SRForest_ORE}%
\vspace{-2ex}
\end{table*}%
\begin{table*}[h!]
  \centering
\caption{Selected rules  provided by applying STEL to the XOR dataset}
\resizebox{0.5\textwidth}{!}{%
    \begin{tabular}{rrrrrr}
    \toprule
          & len & freq & err & condition & pred \\
    \midrule
    \textbf{1} & 2     & 0.24  & 0     &  X[,1] in \{A2,A4\} \& X[,2] in \{B2,B4\} & `1' \\
    \textbf{2} & 2     & 0.23  & 0     & X[,1] in \{A1,A3\} \& X[,2] in \{B1,B3\}  & `1' \\
    \textbf{3} & 1     & 0.52  & 0     & X[,1]==X[,1] & `0' \\
    \bottomrule
    \end{tabular}%
}
\label{tab:SRSTEL}%
\vspace{-2ex}
\end{table*}%

\begin{table*}[h!]
  \centering
  \caption{Selected rules  provided by applying RPART to the XOR dataset}
\resizebox{0.5\textwidth}{!}{%
\centering
    \begin{tabular}{rr}
    \toprule
    pred  & condition \\
    \midrule
    0     & when V1 is A4 \& V2 is B3     \\
    0     & when V1 is A1 or A4 \& V2 is B4 \& V3 is C2 \\
    0     & when V1 is A4 \& V2 is B1 \& V3 is C2 \\
    0     & when V1 is A1 \& V2 is B2 \& V3 is C2 \\
    0     & when V1 is A2 or A3 \& V2 is B3 \& V3 is C2 \\
    0     & when V1 is A3 \& V2 is B2 \& V3 is C2 \\
    0     & when V1 is A2 or A3 \& V2 is B1 or B4 \& V3 is C1 \\
    0     & when V1 is A2 \& V2 is B1 \& V3 is C2 \\
    1     & when V1 is A1 \& V2 is B3     \\
    1     & when V1 is A4 \& V2 is B2 \& V3 is C2 \\
    1     & when V1 is A1 \& V2 is B1 \& V3 is C2 \\
    1     & when V1 is A1 or A4 \& V2 is B1 or B2 or B4 \& V3 is C1 \\
    1     & when V1 is A2 \& V2 is B2 \& V3 is C2 \\
    1     & when V1 is A2 or A3 \& V2 is B2 or B3 \& V3 is C1 \\
    1     & when V1 is A2 \& V2 is B4 \& V3 is C2 \\
    1     & when V1 is A3 \& V2 is B1 or B4 \& V3 is C2 \\
    \bottomrule
    \end{tabular}%
}
  \label{tab:SRCART}%
\vspace{-2ex}
\end{table*}%

\begin{table*}[h!]
    \centering
  \caption{Selected rules  provided by applying RIPPER to the XOR dataset}

\resizebox{0.48\textwidth}{!}{%
\centering
    \begin{tabular}{l}
    \toprule
    Rules \\
    \midrule
    (V1 = A2) and (V2 = B2) $\Rightarrow$ Y=1 (35.0/0.0) \\
    (V2 = B1) and (V1 = A3) $\Rightarrow$ Y=1 (39.0/0.0) \\
    (V2 = B4) and (V1 = A2) $\Rightarrow$ Y=1 (38.0/0.0) \\
    (V1 = A1) and (V2 = B1) $\Rightarrow$ Y=1 (37.0/0.0) \\
    (V3 = C1) and (V2 = B4) and (V1 = A4) $\Rightarrow$ Y=1 (36.0/0.0) \\
    (V2 = B3) and (V1 = A3) $\Rightarrow$ Y=1 (34.0/0.0) \\
    (V1 = A1) and (V2 = B3) $\Rightarrow$ Y=1 (30.0/0.0) \\
    (V2 = B2) and (V1 = A4) $\Rightarrow$ Y=1 (31.0/0.0) \\
     $\Rightarrow$ Y=0 (308.0/0.0) \\
    \bottomrule
   \end{tabular}%
}
  \label{tab:SRRIPPER}%
\vspace{-2ex}
\end{table*}%

\begin{table*}[h!]
    \centering
  \caption{Selected rules  provided by applying SBRL to the XOR dataset}
\resizebox{0.35\textwidth}{!}{%
\centering
    \begin{tabular}{clc}
    \toprule
    id\_rule & cond  & positive proba \\
    \midrule
    26    & {V1=A3,V2=B3} & 0.972 \\
    6     & {V1=A1,V2=B3} & 0.969 \\
    22    & {V1=A3,V2=B1} & 0.976 \\
    2     & {V1=A1,V2=B1} & 0.974 \\
    44    & {V2=B1} & 0.012 \\
    49    & {V2=B3} & 0.013 \\
    42    & {V1=A4} & 0.986 \\
    20    & {V1=A2} & 0.987 \\
    0     &       & 0.007 \\
    \bottomrule
    \end{tabular}%
}
  \label{tab:SRSBRL}%
\vspace{-2ex}
\end{table*}%

\noindent \\In addition to the fact that our framework covers 100\% of our population through four accurate rules defining four clusters, the ``Rule enrichment'' stage allowed the discovery of the $C$ variable influence on each cluster (example in Table \ref{tab:MEtaruleXOR}). Thus, this stage successfully revealed the relevant interactions adding at least one new piece of information to each cluster. This kind of interaction can be crucial in some practical applications even if it does not bring any improvement in terms of prediction accuracy. We can cite, as an example, the case related in \citep{caruanaIntelligibleModelsHealthCare2015b}, about the use of a rule-based system for predicting the death risk due to pneumonia in order to more accurately identify patients who require hospitalization. The learned model assigned a lower risk of dying for asthma patients. Obviously, this rule is counterintuitive, but it reports a real pattern in the data. In fact, asthmatic patients who come to the hospital because of pneumonia are usually admitted directly to the intensive care unit to receive an aggressive treatment, which lowers their risk of dying compared to the general population. This latest information was not provided in the model because it does not improve the prediction performance. Since rules are often chosen to be as concise as possible and are often pruned, much information, sometimes necessary for interpretability, is not revealed. Hence the interest of the ``Rule Enrichment'' stage that we propose in this article.

\begin{table*}[htb]
  \centering
  \caption{Illustration of ``Rule enrichment'' for the XOR dataset}
\resizebox{1\textwidth}{!}{%
\centering
    \begin{tabular}{C{0.4in}C{0.37in}L{2.3in}C{0.37in}C{0.4in}C{0.5in}C{0.4in}C{0.5in}C{0.4in}C{0.4in}C{0.5in}}
    \toprule
    baserule & idRule & condition & Ypred & intersect baserule & confidence & coverage & class coverage & att. nbr & lev. nbr & var. used \\
    \midrule
    \textbf{26} & \textbf{26} & \textbf{X[,1] in \{A2,A4\} \& X[,2] in \{B1,B3\}} & \textbf{`0'} & \textbf{1.00} & \textbf{1.00} & \textbf{0.27} & \textbf{0.51} & \textbf{2} & \textbf{4} & \textbf{V1,V2} \\
    26    & 40    & X[,1] in \{A2,A4\} \& X[,2] in \{B1,B3\} \& X[,3] in \{C2\} & `0'   & 0.74  & 1.00  & 0.20  & 0.38  & 3     & 5     & V1,V2,V3 \\
    \textbf{34} & \textbf{34} & \textbf{X[,1] in \{A2,A4\} \& X[,2] in \{B2,B4\}} & \textbf{`1'} & \textbf{1.00} & \textbf{1.00} & \textbf{0.24} & \textbf{0.50} & \textbf{2} & \textbf{4} & \textbf{V1,V2} \\
    34    & 20    & X[,1] in \{A2,A4\} \& X[,2] in \{B2,B4\} \& X[,3] in \{C2\} & `1'   & 0.74  & 1.00  & 0.18  & 0.37  & 3     & 5     & V1,V2,V3 \\
    \textbf{11} & \textbf{11} & \textbf{X[,1] in \{A1,A3\} \& X[,2] in \{B2,B4\}} & \textbf{`0'} & \textbf{1.00} & \textbf{1.00} & \textbf{0.26} & \textbf{0.49} & \textbf{2} & \textbf{4} & \textbf{V1,V2} \\
    11    & 25    & X[,1] in \{A1,A3\} \& X[,2] in \{B2,B4\} \& X[,3] in \{C2\} & `0'   & 0.76  & 1.00  & 0.20  & 0.37  & 3     & 5     & V1,V2,V3 \\
    \textbf{9} & \textbf{9} & \textbf{X[,1] in \{A1,A3\} \& X[,2] in \{B1,B3\}} & \textbf{`1'} & \textbf{1.00} & \textbf{1.00} & \textbf{0.24} & \textbf{0.50} & \textbf{2} & \textbf{4} & \textbf{V1,V2} \\
    9     & 41    & X[,1] in \{A1,A3\} \& X[,2] in \{B1,B3\} \& X[,3] in \{C2\} & `1'   & 0.76  & 1.00  & 0.18  & 0.38  & 3     & 5     & V1,V2,V3 \\
    \bottomrule
    \end{tabular}%
}
  \label{tab:MEtaruleXOR}%
\end{table*}%

\begin{figure}[h]
\vspace{-2ex} 
\centering
  \includegraphics[width=0.55\textwidth]{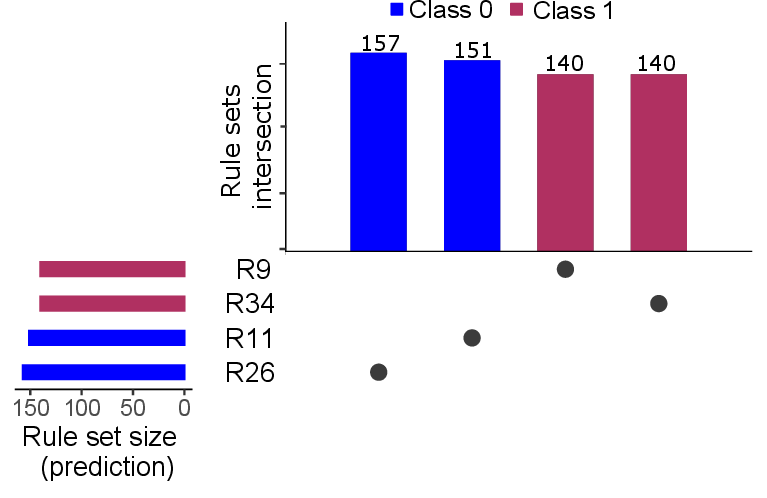}
\caption{Upset visualization on XOR rule sets}
\label{fig:upsetXOR} 
\vspace{-2ex}      
\end{figure}

\noindent In order to visualize the overlaps between the sets defined by the selected rules, one can use approaches available in literature, such as VennEuler \citep{wilkinsonExactApproximateAreaProportional2012}, Upset \citep{lexUpSetVisualizationIntersecting2014b} and Radial sets \citep{alsallakhRadialSetsInteractive2013}. Venn diagram is the most intuitive tool for visualizing intersections and looking at what is shared between groups. However, as the number of sets increases, Venn diagram becomes complex and hard to interpret \citep{ho_what_2021}. Upset and Radial sets are more suitable for multiple overlapping sets. Figure \ref{fig:upsetXOR} uses the Upset method to visualize rules overlaps on the XOR dataset. The horizontal bar chart on the bottom left side shows the distribution of the instances over the rules. Their color reflects the rule classification. The vertical bar chart on the top right side shows the distribution of the instances depending on the combination of rules to which they belong. Their color reflects the true class of the instances. In the example of the XOR dataset, there is no overlap between the rule sets. We give the example of the Mushroom dataset (Appendix \ref{appendix:upsetmushroom}) to show an example of overlaps. 

\noindent In the following tables and figures, we evaluate the performance of the different classifiers over 36 benchmarking datasets.

\subsection{Empirical results}
\label{Experiments}
The benchmarking datasets are available in UCI Machine Learning \citep{Dua:2019} and Keel datasets \citep{j.alcala-fdezKEELDataMiningSoftware2011} repositories. Table \ref{tab:bench.data} gives the total number of instances, descriptive variables, and target classes  for each dataset. Datasets with continuous descriptive variables are discretized by applying ForestDisc discretizer \citep{maissae_novel_2022}. This discretizer has shown excellent results compared with state-of-the-art discretizers.
\\ Since SBRL is developed for binary classification, we have divided the comparative study into two parts. The first one is devoted to binary classification and concerns 19 datasets, and the second to multi-class classification and concerns the remaining 17 datasets. 

\begin{figure}[H]
\vspace{-2ex}  
\centering
  \includegraphics[width=0.75\textwidth]{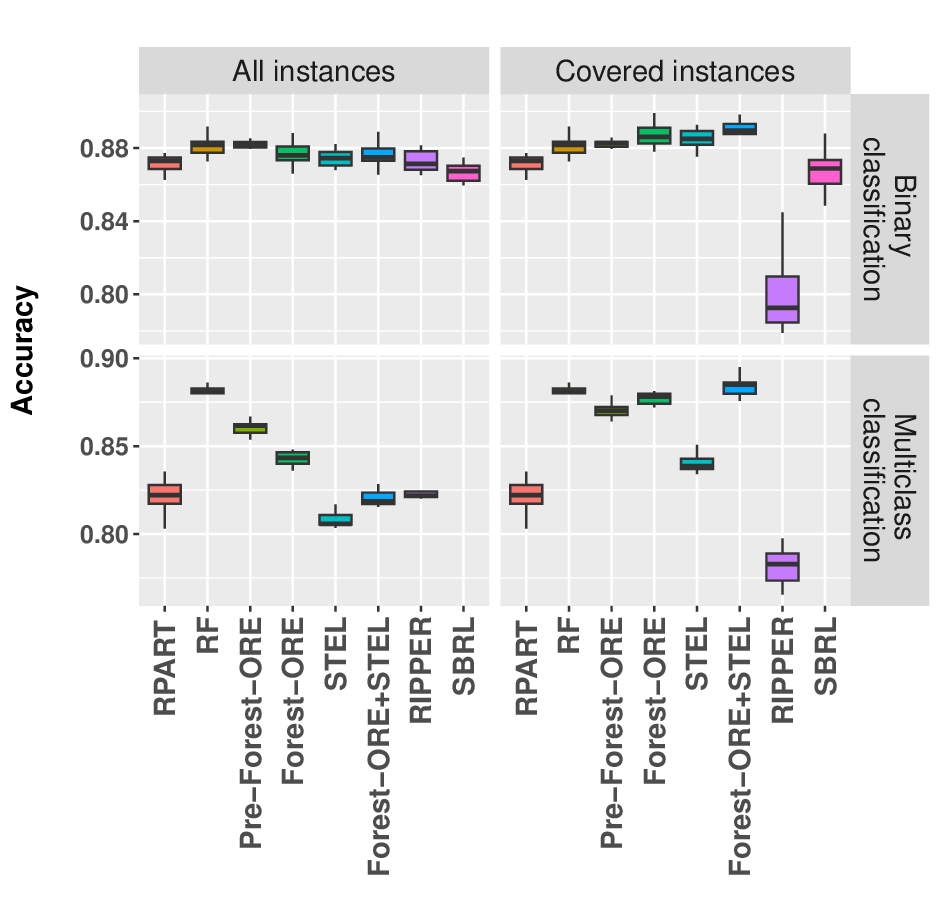}
\caption{Boxplots of the mean accuracy on the testing sets over the benchmark datasets. On the left: Global accuracy. On the right: Accuracy on the testing sets covered by the classifier rules. }
\label{fig:Accuracy} 
\vspace{-2ex}      
\end{figure}

\begin{figure}[H]
\centering
  \includegraphics[width=0.75\textwidth]{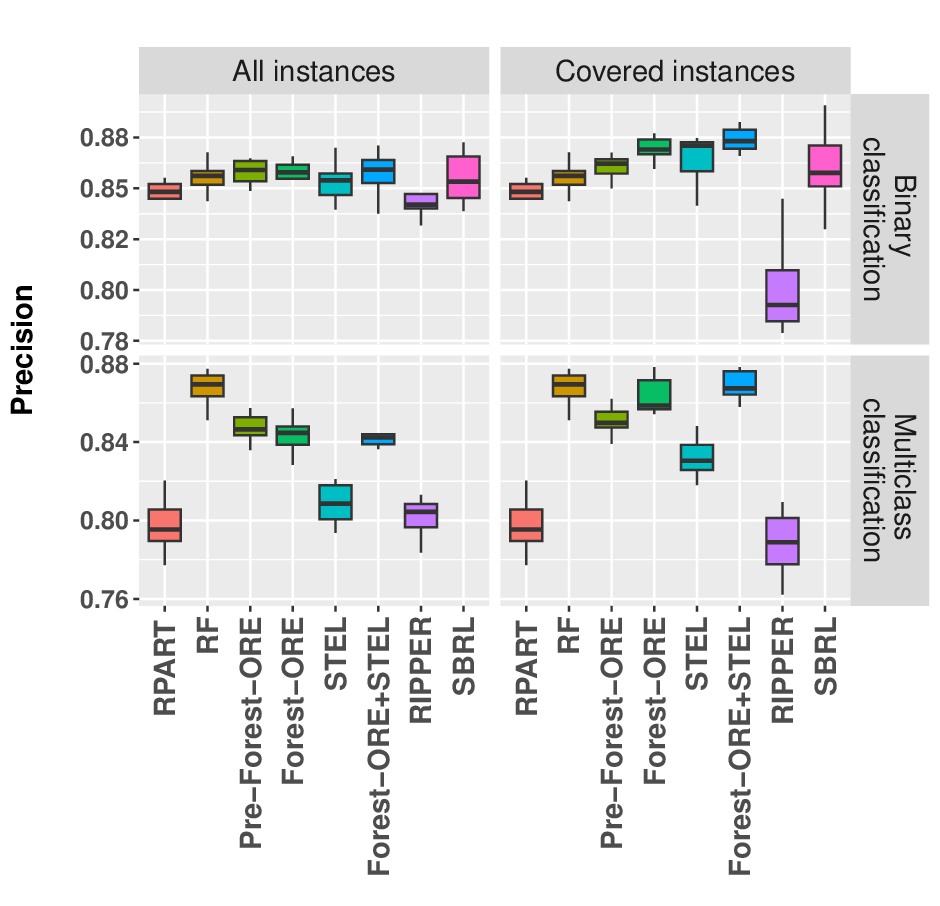}
\caption{Boxplots of the mean macro precision on the testing sets over the benchmark datasets. On the left: Global  macro precision. On the right:  macro precision on the testing sets covered by the classifier rules. }
\label{fig:macroPrecision}     
\vspace{-2ex}  
\end{figure}

\begin{figure}[H]
\centering
  \includegraphics[width=0.75\textwidth]{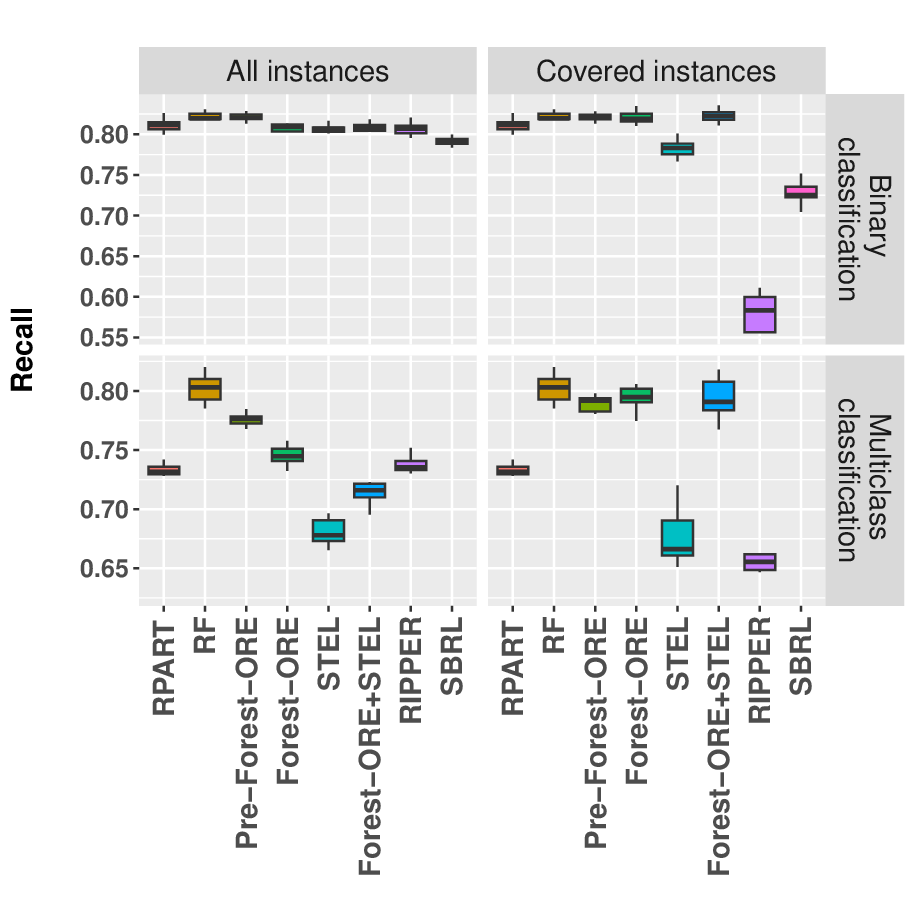}
\caption{Boxplots of the mean macro recall on the testing sets over the benchmark datasets. On the left: Global  macro recall. On the right:  macro recall on the testing sets covered by the classifier rules. }
\label{fig:macroRecall}     
\vspace{-2ex}  
\end{figure}

\begin{figure}[H]
\centering
  \includegraphics[width=0.75\textwidth]{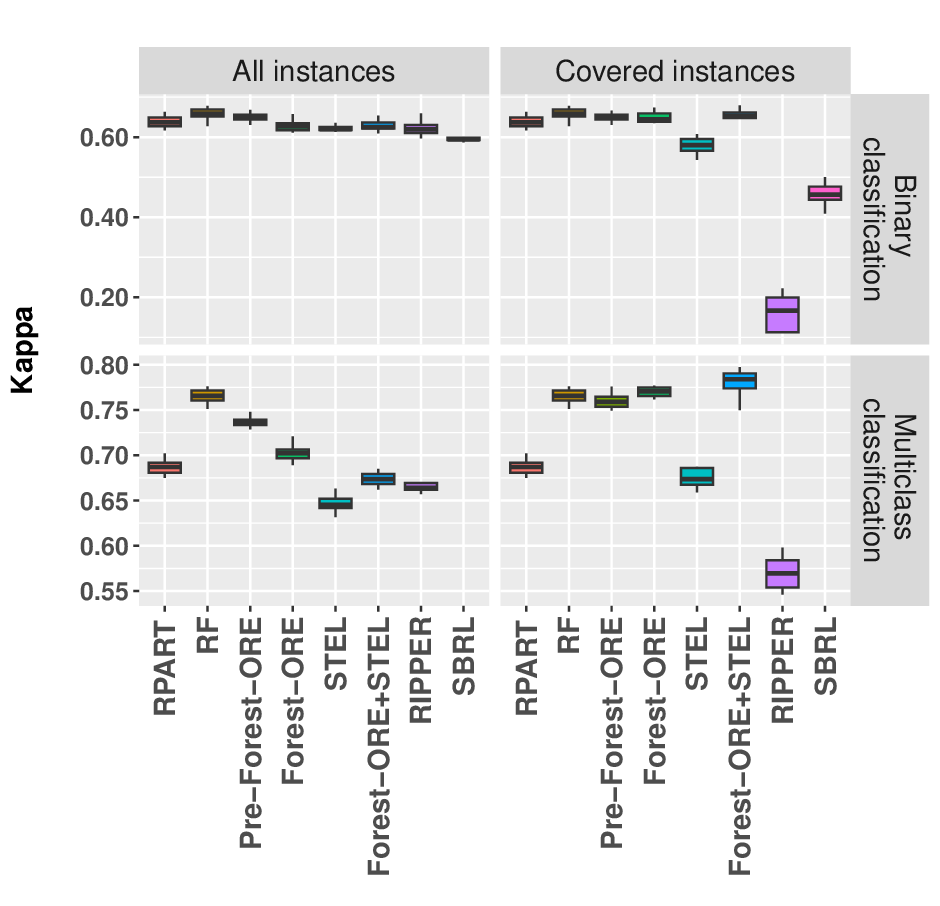}
\caption{Boxplots of the mean Kappa on the testing sets over the benchmark datasets. On the left: Global  Kappa. On the right:  Kappa on the testing sets covered by the classifier rules. }
\label{fig:Kappa}    
\vspace{-2ex}   
\end{figure}

\begin{figure}[htbp]
\centering
  \includegraphics[width=0.75\textwidth]{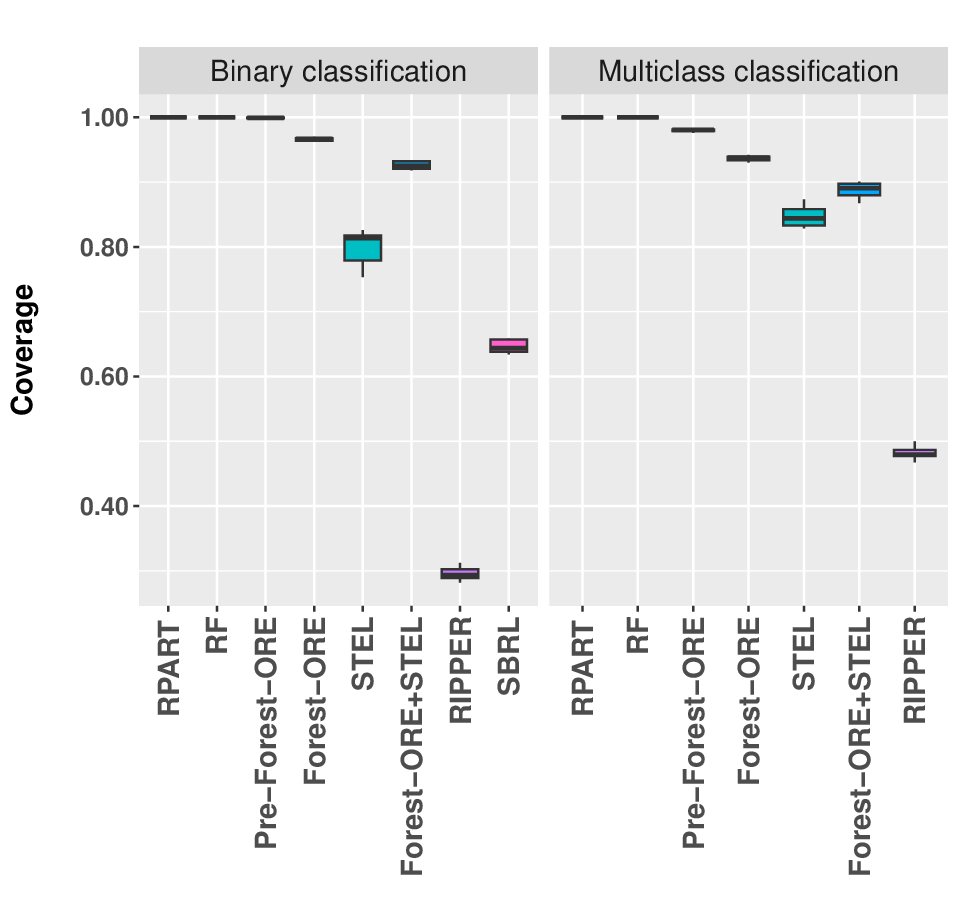}
\caption{Boxplots of the mean coverage on the testing sets over the benchmark datasets.}
\label{fig:coverage}       
\end{figure}

\begin{table*}[htb]
  \centering
  \caption{Benchmark datasets}
\resizebox{0.95\textwidth}{!}{%
    \begin{tabular}{lL{0.85in}C{0.65in}C{0.65in}C{0.4in}rlL{0.85in}C{0.65in}C{0.65in}C{0.4in}}
    \toprule
          & \textbf{Dataset} & \textbf{NB instances} & \textbf{NB attributes} & \textbf{NB classes} &       &       & \textbf{Dataset} & \textbf{NB instances} & \textbf{NB attributes} & \textbf{NB classes} \\
    \midrule
    1     & ANNEAL & 898   & 38    & 5     &       & 19     & MUSHROOM & 8124  & 22    & 2 \\
    2     & APPENDICITIS & 106   & 7     & 2     &       & 20     & MUTAGENESIS & 1618  & 11    & 2 \\
    3     & AUTO  & 205   & 25    & 6     &       & 21     & NEWTHYROID & 215   & 5     & 3 \\
    4     & BANANA & 5300  & 2     & 2     &       & 22     & PAGEBLOCKS & 5473  & 10    & 5 \\
    5     & BANKNOTE & 1372  & 4     & 2     &       & 23     & PHONEME & 5404  & 5     & 2 \\
    6     & BRCANCER & 699   & 9     & 2     &       & 24     & TEXTURE & 5500  & 40    & 11 \\
    7     & CAR   & 1728  & 6     & 4     &       & 25     & THYROID & 7200  & 21    & 3 \\
    8     & CRYOTHERAPY & 90    & 6     & 2     &       & 26     & TICTACTOE & 958   & 9     & 2 \\
    9     & DERMA & 366   & 34    & 6     &       & 27    & TITATNIC & 2201  & 3     & 2 \\
    10    & ECOLI & 336   & 7     & 8     &       & 28    & VERTEBRAL & 310   & 6     & 3 \\
    11    & GLASS & 214   & 9     & 6     &       & 29    & VOTE  & 435   & 16    & 2 \\
    12    & HABERMAN & 306   & 3     & 2     &       & 30    & VOWEL & 990   & 13    & 11 \\
    13    & HEART & 270   & 13    & 2     &       & 31    & WDBC  & 569   & 30    & 2 \\
    14    & HYPOTHYROID & 3163  & 24    & 2     &       & 32    & WILT  & 4839  & 5     & 2 \\
    15    & INDIAN & 583   & 10    & 2     &       & 33    & WINE  & 178   & 13    & 3 \\
    16    & ION   & 351   & 33    & 2     &       & 34   & WIRELESS & 2000  & 7     & 4 \\
    17    & IRIS  & 150   & 4     & 3     &       & 35    & WISCONSIN & 683   & 9     & 2 \\
    18    & MAMMOGRAPHIC & 961   & 5     & 2     &       & 36    & XOR   & 840   & 3     & 2 \\
    \bottomrule
    \end{tabular}%
}
  \label{tab:bench.data}%
\end{table*}%
\noindent \\Figures \ref{fig:Accuracy}, \ref{fig:macroPrecision},  \ref{fig:macroRecall}, and \ref{fig:Kappa} show the variation of the average accuracy, macro precision, macro recall, and kappa on the testing sets over the benchmark datasets. Tables \ref{tab:Accuracy}, \ref{tab:macroPrecision},  \ref{tab:macroRecall}, and \ref{tab:Kappa} display the Wilcoxon signed-rank test scoring in the accuracy, macro precision, macro recall, and Kappa on the testing sets. Each figure and table show the results on binary classification on top, and below, they show results on multiclass classification. On the right, they report the results on the covered instances, and on the left, the global results (on all instances). In the case of the global results, not covered instances are predicted using the default rule provided by each classifier. 
\\When it comes to the global results, RF outperforms the other classifiers, followed by Pre-Forest-ORE and Forest-ORE. If we consider only the covered instances, Forest-ORE, Forest-ORE+STEL, and RF outperform the other classifiers. This result suggests that Forest-ORE performs poorly on the uncovered data (else rule). The differences in the case of multiclass classification are more important than those in the case of binary classification.
\\\\We have annexed in Appendix \ref{appendix:resultsperdata} the average accuracy and coverage metrics and their ranking per dataset and classification issue. More detailed results can be found on the GitHub repository reserved for this study (refer to Section "Declarations").
The Friedman test on these ranking results rejected the null hypothesis of no difference among the compared algorithms. The p-value is below $ 4.51E-05$ for all the compared results (see Appendix ~\ref{appendix:resultsperdata}).
\\The Wilcoxon signed-rank test confirms the differences in the predictive performance especially, in the case of multiclass classification. If we consider the sum of wins and ties, the differences are more important in multiclass classification than in binary classification. 
\\Based on the results of the Kappa measure, and according to Landis and Koch classification \citep{ landisMeasurementObserverAgreement1977}, the usefulness of the classification processed by RF, PreForest-ORE, Forest-ORE, STEL, Forest-ORE+STEL, and RPART methods, on covered instances is deemed substantial, whereas the usefulness of the classification processed by the other methods is considered moderate.”
\\\\Table \ref{tab:fidelity} reports the fidelity metric measured for the methods that are intended to approximate the RF model. The fidelity is measured on the testing sets. It is also broken down according to whether the instances are correctly or incorrectly predicted by RF. It is more interesting to mimic the RF model on correctly predicted instances than on incorrectly predicted instances. This figure also reports the coverage measure to show the rate of explainable instances. Based on this table, Forest ORE enables, on average 95\% of agreement with RF on the predicted covered instances. This rate is more important when the instances are correctly predicted by the RF model (97\%). The agreement of Forest-ORE with RF in inexplainable instances (which represent, on average 5\% of the instances) is around 55\%. This result confirms the previous result that suggests that Forest-ORE performs poorly on uncovered data (predicted using the else rule). This issue will be analyzed in future work to improve the overall predictive performance of Forest-ORE. Forest-ORE performs better than STEL regarding the fidelity metric. The agreement of STEL with the RF model is, on average, around 92\% on the covered instances, and the rate of explainable instances is about 82\%. STEL performs better than Forest-ORE on inexplainable instances.
\\\\Figure \ref{fig:coverage} and Table \ref{tab:lRlength} report the average rules' coverage (on testing sets) and complexity over the benchmark datasets. Complexity concerns the number of rules per target class and the number of attributes used per rule. 
Based on the reported results, RPART produces the best trade-off between coverage and complexity. Forest-ORE enables the second-best trade-off. It covers, on average, 95\% of the instances with an average of 3.8 rules per target class and 3.3 attributes per rule. STEL, SBRL, and RIPPER tend to produce less complex models but worse coverage results. It is not surprising that ordered rule lists are less complex than unordered rule lists. As previously reported in the illustrative example, the expression of a specific rule in an ordered rule list is in fact the intersection between this rule conditions with the negation of all the preceding conditions. Thus, the rule length in an ordered list does not reflect the real length of the rule.

\begin{table}[htb]
\vspace{-2ex}
  \centering
 \caption{Wilcoxon signed rank test scoring in accuracy on the testing sets. On the left: Global scores. On the right: the scores on the testing sets covered by the classifier rules.}
\resizebox{0.9\textwidth}{!}{%
\centering
    \begin{tabular}{|c|lcccc|clccc|}
    \toprule
          & \multicolumn{4}{c}{\textbf{All instances}} &       &       & \multicolumn{4}{c|}{\textbf{Covered instances}} \\
\cmidrule{2-11}    \multicolumn{1}{|c|}{} & \textbf{Method} & \textbf{Wins} & \textbf{Ties} & \textbf{Losses} &       &       & \textbf{Method} & \textbf{Wins} & \textbf{Ties} & \textbf{Losses} \\
    \midrule
    \multirow{8}[2]{*}{\begin{sideways}Binary classification\end{sideways}} & RF    & 7     & 0     & 0     &       &       & Forest-ORE+STEL & 7     & 0     & 0 \\
          & Pre-Forest-ORE & 6     & 0     & 1     &       &       & Forest-ORE & 4     & 2     & 1 \\
          & Forest-ORE & 1     & 4     & 2     &       &       & RF    & 3     & 3     & 1 \\
          & Forest-ORE+STEL & 1     & 4     & 2     &       &       & STEL  & 3     & 3     & 1 \\
          & STEL  & 1     & 4     & 2     &       &       & Pre-Forest-ORE & 2     & 2     & 3 \\
          & RIPPER & 1     & 4     & 2     &       &       & SBRL  & 1     & 3     & 3 \\
          & RPART & 0     & 5     & 2     &       &       & RPART & 1     & 1     & 5 \\
          & SBRL  & 0     & 1     & 6     &       &       & RIPPER & 0     & 0     & 7 \\
    \midrule
    \multicolumn{1}{r}{} &       &       &       &       & \multicolumn{1}{r}{} &       &       &       &       & \multicolumn{1}{r}{} \\
    \midrule
    \multirow{7}[2]{*}{\begin{sideways}Multi. classification\end{sideways}} & RF    & 6     & 0     & 0     &       &       & Forest-ORE+STEL & 6     & 0     & 0 \\
          & Pre-Forest-ORE & 5     & 0     & 1     &       &       & RF    & 4     & 1     & 1 \\
          & Forest-ORE & 3     & 1     & 2     &       &       & Forest-ORE & 4     & 1     & 1 \\
          & Forest-ORE+STEL & 2     & 1     & 3     &       &       & Pre-Forest-ORE & 3     & 0     & 3 \\
          & RPART & 1     & 3     & 2     &       &       & STEL  & 2     & 0     & 4 \\
          & RIPPER & 0     & 2     & 4     &       &       & RPART & 1     & 0     & 5 \\
          & STEL  & 0     & 1     & 5     &       &       & RIPPER & 0     & 0     & 6 \\
    \bottomrule
    \end{tabular}%
}
  \label{tab:Accuracy}%
\end{table}%

\begin{table}[htb]
  \centering
 \caption{Wilcoxon signed rank test scoring in the macro precision on the testing sets. On the left: Global scores. On the right: the scores on the testing sets covered by the classifier rules.}
\resizebox{0.9\textwidth}{!}{%
\centering
    \begin{tabular}{|c|lcccc|clccc|}
   \toprule
          & \multicolumn{4}{c}{\textbf{All instances}} &       &       & \multicolumn{4}{c|}{\textbf{Covered instances}} \\
\cmidrule{2-11}    \multicolumn{1}{|c|}{} & \textbf{Method} & \textbf{Wins} & \textbf{Ties} & \textbf{Losses} &       &       & \textbf{Method} & \textbf{Wins} & \textbf{Ties} & \textbf{Losses} \\
    \midrule
    \multirow{8}[2]{*}{\begin{sideways}Binary classification\end{sideways}} & RF    & 6     & 1     & 0     &       &       & Forest-ORE+STEL & 6     & 1     & 0 \\
          & Pre-Forest-ORE & 2     & 4     & 1     &       &       & Forest-ORE    & 3     & 3     & 1 \\
          & Forest-ORE & 1     & 5     & 1     &       &       & RF & 3     & 3     & 1 \\
          & Forest-ORE+STEL & 1     & 5     & 1     &       &       & SBRL  & 2     & 5     & 0 \\
          & SBRL  & 0     & 7     & 0     &       &       & STEL  & 2     & 4     & 1 \\
          & STEL  & 0     & 6     & 1     &       &       & Pre-Forest-ORE & 2     & 2     & 3 \\
          & RIPPER & 0     & 5     & 2     &       &       & RPART & 1     & 0     & 6 \\
          & RPART & 0     & 3     & 4     &       &       & RIPPER & 0     & 0     & 7 \\
    \midrule
    \multicolumn{1}{r}{} &       &       &       &       & \multicolumn{1}{r}{} &       &       &       &       & \multicolumn{1}{r}{} \\
    \midrule
    \multirow{7}[2]{*}{\begin{sideways}Multi. classification\end{sideways}} & RF    & 6     & 0     & 0     &       &       & Forest-ORE+STEL & 5     & 1     & 0 \\
          & Pre-Forest-ORE & 3     & 2     & 1     &       &       & RF    & 4     & 2     & 0 \\
          & Forest-ORE & 3     & 2     & 1     &       &       & Forest-ORE & 4     & 1     & 1 \\
          & Forest-ORE+STEL & 3     & 2     & 1     &       &       & Pre-Forest-ORE & 3     & 0     & 3 \\
          & RPART & 0     & 2     & 4     &       &       & STEL  & 2     & 0     & 4 \\
          & STEL  & 0     & 2     & 4     &       &       & RPART & 0     & 1     & 5 \\
          & RIPPER & 0     & 2     & 4     &       &       & RIPPER & 0     & 1     & 5 \\
    \bottomrule
    \end{tabular}%
}
  \label{tab:macroPrecision}%
\end{table}%

\begin{table}[htb]
  \centering
 \caption{Wilcoxon signed rank test scoring in the macro recall on the testing sets. On the left: Global scores. On the right: the scores on the testing sets covered by the classifier rules.}
\resizebox{0.9\textwidth}{!}{%
\centering
    \begin{tabular}{|c|lcccc|clccc|}
     \toprule
          & \multicolumn{4}{c}{\textbf{All instances}} &       &       & \multicolumn{4}{c|}{\textbf{Covered instances}} \\
\cmidrule{2-11}    \multicolumn{1}{|c|}{} & \textbf{Method} & \textbf{Wins} & \textbf{Ties} & \textbf{Losses} &       &       & \textbf{Method} & \textbf{Wins} & \textbf{Ties} & \textbf{Losses} \\
    \midrule
    \multirow{8}[2]{*}{\begin{sideways}Binary classification\end{sideways}} & RF    & 6     & 1     & 0     &       &       & RF    & 4     & 3     & 0 \\
          & Pre-Forest-ORE & 6     & 1     & 0     &       &       & Pre-Forest-ORE & 4     & 3     & 0 \\
          & Forest-ORE & 1     & 4     & 2     &       &       & Forest-ORE & 4     & 3     & 0 \\
          & Forest-ORE+STEL & 1     & 4     & 2     &       &       & Forest-ORE+STEL & 4     & 3     & 0 \\
          & RPART & 1     & 4     & 2     &       &       & RPART & 2     & 1     & 4 \\
          & STEL  & 1     & 4     & 2     &       &       & STEL  & 2     & 1     & 4 \\
          & RIPPER & 1     & 4     & 2     &       &       & SBRL  & 1     & 0     & 6 \\
          & SBRL  & 0     & 0     & 7     &       &       & RIPPER & 0     & 0     & 7 \\
    \midrule
    \multicolumn{1}{r}{} &       &       &       &       & \multicolumn{1}{r}{} &       &       &       &       & \multicolumn{1}{r}{} \\
    \midrule
    \multirow{7}[2]{*}{\begin{sideways}Multi. classification\end{sideways}} & RF    & 6     & 0     & 0     &       &       & RF    & 6     & 0     & 0 \\
          & Pre-Forest-ORE & 5     & 0     & 1     &       &       & Pre-Forest-ORE & 3     & 2     & 1 \\
          & Forest-ORE & 2     & 2     & 2     &       &       & Forest-ORE & 3     & 2     & 1 \\
          & RPART & 1     & 3     & 2     &       &       & Forest-ORE+STEL & 3     & 2     & 1 \\
          & RIPPER & 1     & 3     & 2     &       &       & RPART & 2     & 0     & 4 \\
          & Forest-ORE+STEL & 1     & 2     & 3     &       &       & STEL  & 0     & 1     & 5 \\
          & STEL  & 0     & 0     & 6     &       &       & RIPPER & 0     & 1     & 5 \\
    \bottomrule
    \end{tabular}%
}
  \label{tab:macroRecall}%
\end{table}%

\begin{table}[htb]
\vspace{-2ex}
  \centering
 \caption{Wilcoxon signed rank test scoring in the Kappa score on the testing sets. On the left: Global scores. On the right: the scores on the testing sets covered by the classifier rules.}
\resizebox{0.9\textwidth}{!}{%
\centering
  \begin{tabular}{|c|lcccc|clccc|}
  \toprule
          & \multicolumn{4}{c}{\textbf{All instances}} &       &       & \multicolumn{4}{c|}{\textbf{Covered instances}} \\
\cmidrule{2-11}    \multicolumn{1}{|c|}{} & \textbf{Method} & \textbf{Wins} & \textbf{Ties} & \textbf{Losses} &       &       & \textbf{Method} & \textbf{Wins} & \textbf{Ties} & \textbf{Losses} \\
    \midrule
    \multirow{8}[2]{*}{\begin{sideways}Binary classification\end{sideways}} & RF    & 7     & 0     & 0     &       &       & RF    & 5     & 2     & 0 \\
          & Pre-Forest-ORE & 6     & 0     & 1     &       &       & Forest-ORE & 4     & 3     & 0 \\
          & Forest-ORE & 1     & 4     & 2     &       &       & Forest-ORE+STEL & 4     & 3     & 0 \\
          & Forest-ORE+STEL & 1     & 4     & 2     &       &       & Pre-Forest-ORE & 4     & 2     & 1 \\
          & RPART & 1     & 4     & 2     &       &       & RPART & 2     & 1     & 4 \\
          & STEL  & 1     & 4     & 2     &       &       & STEL  & 2     & 1     & 4 \\
          & RIPPER & 1     & 4     & 2     &       &       & SBRL  & 1     & 0     & 6 \\
          & SBRL  & 0     & 0     & 7     &       &       & RIPPER & 0     & 0     & 7 \\
    \midrule
    \multicolumn{1}{r}{} &       &       &       &       & \multicolumn{1}{r}{} &       &       &       &       & \multicolumn{1}{r}{} \\
    \midrule
    \multirow{7}[2]{*}{\begin{sideways}Multi. classification\end{sideways}} & RF    & 6     & 0     & 0     &       &       & Forest-ORE+STEL & 5     & 1     & 0 \\
          & Pre-Forest-ORE & 5     & 0     & 1     &       &       & RF    & 4     & 2     & 0 \\
          & Forest-ORE & 3     & 1     & 2     &       &       & Forest-ORE & 3     & 2     & 1 \\
          & RPART & 2     & 2     & 2     &       &       & Pre-Forest-ORE & 3     & 1     & 2 \\
          & Forest-ORE+STEL & 2     & 1     & 3     &       &       & RPART & 1     & 1     & 4 \\
          & STEL  & 0     & 1     & 5     &       &       & STEL  & 1     & 1     & 4 \\
          & RIPPER & 0     & 1     & 5     &       &       & RIPPER & 0     & 0     & 6 \\
    \bottomrule
    \end{tabular}%
}
  \label{tab:Kappa}%
\end{table}%

\begin{table*}[h!]
\centering
\caption{Fidelity of explanations}
\resizebox{1\textwidth}{!}{%
    \begin{tabular}{rlcccccccc}
    \toprule
          &       & \multicolumn{2}{c}{\textbf{Coverage}} & \multicolumn{2}{C{1in}}{\textbf{Fidelity on all instances}} & \multicolumn{2}{C{1.2in}}{\textbf{Fidelity on instances correctly predicted by RF}} & \multicolumn{2}{C{1.2in}}{\textbf{Fidelity on instances incorrectly predicted by RF}} \\
    \midrule
          &       & Mean  & SE    & Mean  & SE    & Mean  & SE    & Mean  & SE \\
\cmidrule{3-10}    \multicolumn{1}{r}{\multirow{3}[2]{*}{\textbf{Pre-Forest-ORE}}} & all instances & 1.000 & 0.000 & 0.952 & 0.002 & 0.967 & 0.001 & 0.803 & 0.007 \\
          & covered instances & 0.990 & 0.000 & 0.956 & 0.002 & 0.971 & 0.001 & 0.805 & 0.008 \\
          & not covered instances & 0.010 & 0.000 & 0.604 & 0.054 & 0.642 & 0.052 & 0.571 & 0.056 \\
    \midrule
    \multicolumn{1}{r}{\multirow{3}[2]{*}{\textbf{Forest-ORE}}} & all instances & 1.000 & 0.000 & 0.932 & 0.002 & 0.951 & 0.001 & 0.741 & 0.010 \\
          & covered instances & 0.952 & 0.001 & 0.955 & 0.001 & 0.970 & 0.001 & 0.803 & 0.007 \\
          & not covered instances & 0.048 & 0.001 & 0.545 & 0.012 & 0.559 & 0.017 & 0.496 & 0.014 \\
    \midrule
    \multicolumn{1}{r}{\multirow{3}[2]{*}{\textbf{Forest-ORE+STEL}}} & all instances & 1.000 & 0.000 & 0.917 & 0.002 & 0.937 & 0.001 & 0.725 & 0.012 \\
          & covered instances & 0.908 & 0.002 & 0.959 & 0.001 & 0.972 & 0.001 & 0.824 & 0.009 \\
          & not covered instances & 0.092 & 0.002 & 0.549 & 0.014 & 0.568 & 0.017 & 0.478 & 0.021 \\
        \midrule
    \multicolumn{1}{r}{\multirow{3}[2]{*}{\textbf{STEL}}} & all instances & 1.000 & 0.000 & 0.908 & 0.001 & 0.927 & 0.001 & 0.738 & 0.009 \\
          & covered instances & 0.823 & 0.004 & 0.925 & 0.003 & 0.944 & 0.002 & 0.755 & 0.011 \\
          & not covered instances & 0.177 & 0.004 & 0.767 & 0.010 & 0.789 & 0.013 & 0.650 & 0.018 \\
    \bottomrule
    \end{tabular}%
     }
\label{tab:fidelity}%
\end{table*}%

\begin{table*}[h!]
\centering
\caption{Complexity of explanations. Complexity concerns the number of rules per target class and the number of attributes used per rule.}
\resizebox{0.75\textwidth}{!}{%
  \begin{tabular}{|c|lrrrr|rrr}
    \toprule
    \multicolumn{1}{r}{} &       & \multicolumn{3}{c}{Number of rules per target class} & \multicolumn{1}{c}{} & \multicolumn{3}{c}{Number of attributes per rule} \\
\cmidrule{3-9}    \multicolumn{1}{r}{} &       & Average & Min   & Max   & \multicolumn{1}{r}{} & Average & Min   & Max \\
    \midrule
    \multirow{8}[2]{*}{\begin{sideways}Binary classification\end{sideways}} & RPART & 3.2   & 1.1   & 10.7  &       & 2.5   & 1.3   & 3.2 \\
          & RF    & 1221.7 & 315.6 & 2984.7 &       & 4.4   & 1.3   & 7.9 \\
          & Pre-Forest-ORE & 212.7 & 21    & 431.4 &       & 3.9   & 1.4   & 5.4 \\
          & Forest-ORE & 4.4   & 1.5   & 14.1  &       & 3.1   & 1.8   & 4.4 \\
          & STEL  & 3.2   & 0.9   & 8.6   &       & 2.2   & 1.4   & 3.2 \\
          & Forest-ORE+STEL & 4     & 1     & 11.7  &       & 3.1   & 1.9   & 4.3 \\
          & RIPPER & 2.7   & 0.2   & 6.3   &       & 2.1   & 1.6   & 2.8 \\
          & SBRL  & 2.1   & 0.6   & 5.3   &       & 1.6   & 1.2   & 2 \\
    \midrule
    \multicolumn{1}{r}{} &       &       &       &       & \multicolumn{1}{r}{} &       &       &  \\
    \midrule
    \multirow{7}[2]{*}{\begin{sideways}Multi. classification\end{sideways}} & RPART & 1.8   & 0.8   & 4.1   &       & 3.8   & 2.3   & 4.7 \\
          & RF    & 931.9 & 241.3 & 3219.9 &       & 5.2   & 1.1   & 9.5 \\
          & Pre-Forest-ORE & 115.7 & 19    & 320.1 &       & 4.1   & 1.1   & 5.4 \\
          & Forest-ORE & 3.3   & 1.3   & 10    &       & 3.5   & 1.5   & 4.9 \\
          & STEL  & 1.6   & 0.6   & 2.9   &       & 2.4   & 1.3   & 3.7 \\
          & Forest-ORE+STEL & 2.3   & 1.1   & 4.7   &       & 3.4   & 1.6   & 4.8 \\
          & RIPPER & 2.8   & 1     & 6.6   &       & 2.2   & 1.2   & 3.5 \\
    \bottomrule
    \end{tabular}%
   }
\label{tab:lRlength}%
\end{table*}%

\subsection{Ablation Studies}
\label{abl_study}
In this section, we analyze the effect of the different weights used in the function objective (Formula \ref{Obj.Func.}) on the quality of the set of rules and their predictive performance and coverage. We also analyze the impact of suppressing the preselection stage on these measures. The study includes 20 datasets; ten of them concern binary classification, and ten concern multi-class classification. 
\\Similarly to the ablation methodology used in \citep{lakkaraju_interpretable_2016}, we obtain the first four ablation models by excluding the weights from the objective function one at a time. The fifth model is obtained by suppressing the preselection stage from Forest-ORE processing. Ablation models are described in Table \ref{tab:Abl_models}.
\\Table \ref{tab:Abl_results} shows that “abl\_conf” and “abl\_cov” models induce a decrease in the average rule confidence and average rule coverage. This result demonstrates that excluding $W_0 $ and $W_1$ weights lowers the quality of the rules. Similar observations can be made about excluding $W_2 $ and $W_3$, which induce an increase in the number of attributes and modalities per rule. This result demonstrates that excluding the $W_2 $ and $W_3$ increase the complexity of the rule ensemble.
\\Table \ref{tab:Abl_perfresults}, reports the average results on the predictive performance. This table shows that the predictive performance of “abl\_lenght” and “abl\_preselect” are slightly better than “no\_abl” model. However, The “abl\_lenght” model induces an increase in the complexity of the model (increase in the length of the rules). On the other hand, the “abl\_preselect” model induces a increase in the computational time (Table \ref{tab:Abl_exetime}). 
\\ These results show, on the one hand, that each term in the objective function contributes to improving the quality of the final rule ensemble. On the other hand, it demonstrates the importance of the preselection stage. Using the preselection stage induces a very small loss in predictive performance (0.003 on average accuracy), but an important gain in computational time (divised by 4.7 on average).

\begin{table*}[h]
\centering
\caption{Description of the ablation models}
\resizebox{0.9\textwidth}{!}{%
    \begin{tabular}{p{9em}p{5.5em}p{18em}p{4em}}
    \toprule
    Ablation model & Abbrev. & Description & Setting \\
    \midrule
    No high confidence & abl\_conf & is obtained by excluding the term which encourages rules with high confidence  & $W_0 = 0$ \\
    No high coverage & abl\_cov & is obtained by excluding the term that encourages rules with high coverage & $W_1 =0 $ \\
    No reduced length & abl\_length & is obtained by excluding the term which encourages rules with few attributes & $W_2 = 0$ \\
    No reduced modalities & abl\_mod & is obtained by excluding the term which encourages rules with few levels & $W_3 = 0$ \\
    No preselection stage & abl\_preselect & is obtained by excluding the preselection stage, which is intended to reduce the size of the initial set of rules by selecting the best RF rules, based on their individual performance & Remove preselection stage \\
    \bottomrule
    \end{tabular}%
   }
\label{tab:Abl_models}%
\end{table*}%

\begin{table*}[h!]
\centering
\caption{Results of the ablation on the quality of the rules}
\resizebox{0.9\textwidth}{!}{%
  \begin{tabular}{lcccccccccc}
    \toprule
          & \multicolumn{2}{p{6.56em}}{Confidence} & \multicolumn{2}{p{6.56em}}{Coverage} & \multicolumn{2}{p{6.56em}}{Class coverage} & \multicolumn{2}{p{6.56em}}{Number of variables} & \multicolumn{2}{p{6.56em}}{Number of levels} \\
    \midrule
    abl\_model & Mean  & SE    & Mean  & SE    & Mean  & SE    & Mean  & SE    & Mean  & SE \\
    \midrule
 no\_abl & 0.930 & 0.005 & 0.190 & 0.007 & 0.485 & 0.015 & 3.016 & 0.068 & 6.986 & 0.309 \\
    abl\_conf & 0.912 & 0.006 & 0.196 & 0.007 & 0.501 & 0.015 & 2.969 & 0.066 & 6.827 & 0.293 \\
    abl\_cov & 0.932 & 0.005 & 0.177 & 0.007 & 0.460 & 0.015 & 2.978 & 0.068 & 6.663 & 0.301 \\
    abl\_length & 0.930 & 0.005 & 0.191 & 0.007 & 0.487 & 0.015 & 3.089 & 0.069 & 7.230 & 0.324 \\
    abl\_mod & 0.930 & 0.005 & 0.190 & 0.007 & 0.487 & 0.015 & 3.031 & 0.068 & 7.103 & 0.312 \\
    abl\_preselect & 0.928 & 0.005 & 0.197 & 0.008 & 0.510 & 0.016 & 3.002 & 0.074 & 6.971 & 0.325 \\
    \bottomrule
    \end{tabular}%
    }
\label{tab:Abl_results}%
\end{table*}%

\begin{table*}[h!]
\centering
\caption{Results of the ablation on the predictive performance of the rules}
\resizebox{0.7\textwidth}{!}{%
    \begin{tabular}{lcccccccc}
    \toprule
          & \multicolumn{2}{c}{Accuracy} & \multicolumn{2}{c}{F1 score} & \multicolumn{2}{c}{Fidelity} & \multicolumn{2}{c}{Coverage} \\
\cmidrule{2-9}          & Mean  & SE    & Mean  & SE    & Mean  & SE    & Mean  & SE \\
    \midrule
    no\_abl & 0.859 & 0.003 & 0.825 & 0.004 & 0.952 & 0.002 & 0.958 & 0.001 \\
    abl\_conf & 0.860 & 0.004 & 0.826 & 0.005 & 0.945 & 0.003 & 0.958 & 0.002 \\
    abl\_cov & 0.860 & 0.003 & 0.826 & 0.004 & 0.955 & 0.002 & 0.955 & 0.002 \\
    abl\_length & 0.862 & 0.003 & 0.827 & 0.004 & 0.954 & 0.002 & 0.957 & 0.001 \\
    abl\_mod & 0.860 & 0.003 & 0.826 & 0.004 & 0.954 & 0.002 & 0.958 & 0.001 \\
    abl\_preselect & 0.862 & 0.003 & 0.828 & 0.004 & 0.951 & 0.002 & 0.956 & 0.004 \\
    \bottomrule
    \end{tabular}%
   }
\label{tab:Abl_perfresults}%
\end{table*}%

  \begin{table*}[h!]
\centering
\caption{Results of the ablation on the exection time}
\resizebox{1\textwidth}{!}{%
       \begin{tabular}{lcccccccccccc}
    \toprule
          & \multicolumn{4}{c}{Size of the set of rules} & \multicolumn{8}{c}{Execution time (s)} \\
\cmidrule{2-13}          & \multicolumn{2}{c}{Initial set} & \multicolumn{2}{c}{Final set} & \multicolumn{2}{c}{Extract/preselect rules} & \multicolumn{2}{c}{Prepare opt inputs} & \multicolumn{2}{c}{Build opt. model} & \multicolumn{2}{c}{Run Opt.} \\
\cmidrule{2-13}          & Mean  & SE    & Mean  & SE    & Mean  & SE    & Mean  & SE    & Mean  & SE    & Mean  & SE \\
    \midrule
    no\_abl & 445   & 6     & 9.76 & 0.46 & 10.52 & 0.05  & 15.17 & 0.05  & 28.05 & 0.10   & 24.31 & 2.98 \\
    abl\_conf & 445   & 6     & 9.77  & 0.46  & 10.52 & 0.05  & 15.17 & 0.05  & 28.09 & 0.10   & 36.00    & 4.07 \\
    abl\_cov & 445   & 6     & 9.78  & 0.46  & 10.52 & 0.05  & 15.17 & 0.05  & 28.10  & 0.09  & 18.30  & 1.34 \\
    abl\_length & 445   & 6     & 9.79  & 0.46  & 10.52 & 0.05  & 15.17 & 0.05  & 28.08 & 0.09  & 22.20  & 2.83 \\
    abl\_mod & 445   & 6     & 9.80  & 0.47  & 10.52 & 0.05  & 15.17 & 0.05  & 28.14 & 0.09  & 25.82 & 2.74 \\
    abl\_preselect & 1978  & 20.00    & 9.13  & 0.42  & 6.49  & 0.03  & 140.36 & 1.80   & 155.12 & 0.31  & 62.13 & 8.82 \\
    \bottomrule
    \end{tabular}%
   }
\label{tab:Abl_exetime}%
\end{table*}%

\section{Discussion}
\begin{figure}[h]
\centering
  \includegraphics[width=1\textwidth]{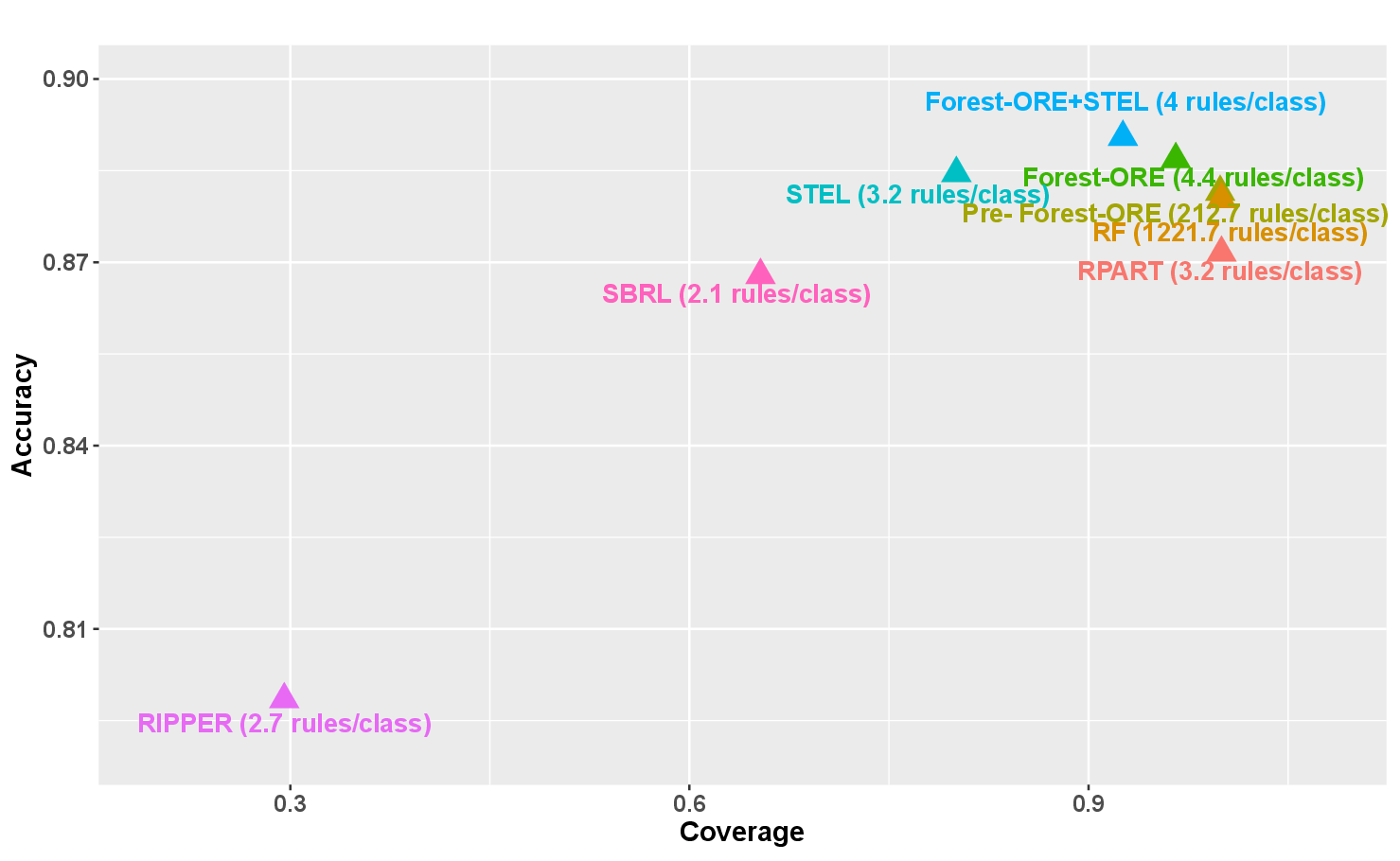}
\caption{Trade-off between the accuracy and the coverage of the rule set explaining the data (case of binary classification). }
\label{fig:accuracyvscoverage_bin}
 \vspace{-2ex}      
\end{figure}

\begin{figure}[H]
\centering
  \includegraphics[width=1\textwidth]{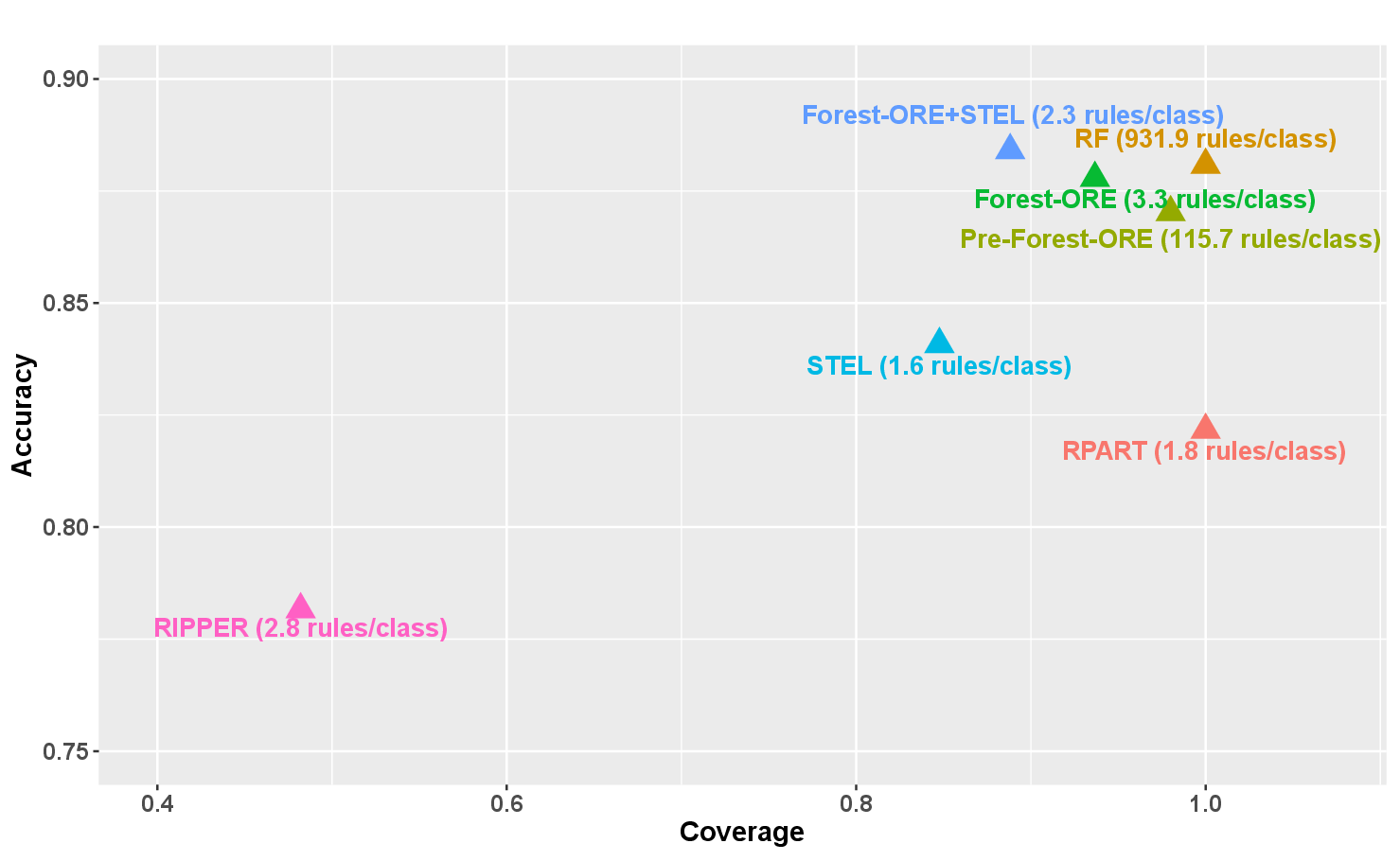}
\caption{Trade-off between the accuracy and the coverage of the rule set explaining the data (case of multiclass classification). }
\label{fig:accuracyvscoverage_multi}
\vspace{-2ex}       
\end{figure}

The empirical analysis provided in section \ref{Experiments} shows that Forest-ORE enables an excellent  trade-off between the predictive performance, the coverage of the model, and its complexity. Figures \ref{fig:accuracyvscoverage_bin} and \ref {fig:accuracyvscoverage_multi} emphasize this result.
These figures show the position of the different classifiers regarding the trade-off between the average accuracy and the average coverage of the rule set explaining the data. Classifiers near the top right corner of the graph should be preferred for being the most accurate and the most complete in terms of the rate of data explained. In addition, the size of the rule set explaining the different classes should be small to enable interpretation.

However, Forest-ORE is not competitive with the other classifiers in computational time. This can be a drawback if the interpretability of RF results is required online or when processing large datasets.
Appendix \ref{appendix:exetime} reports the execution time spent by Forest-ORE over the benchmarking datasets. 
Forest-ORE's execution time is broken down into five parts. The first part concerns the rule extraction task, and the second the rule preselection task. The third part concerns the preparation of the inputs for the optimization step. Finally, the last parts report the execution time spent on building and executing the optimization task. In this first version of Forest-ORE, we did not focus on optimizing the computational time or parallelizing tasks. An analysis of the time complexity of the three first parts can easily show that some parameters should be considered while optimizing their computational time. These parameters are the max-depth and the max-leaf nodes in RF trees, the number of RF trees, and the rules max-length. The last parts concern solving an MIP problem that is theoretically NP-hard. However, by using the Gurobi solver, we could find feasible solutions in acceptable execution times for most benchmark datasets. In addition, the results suggest that MIP running time is problem-dependent: we did not observe a strong relationship between the MIP running time and the size of the datasets or the number of its descriptive attributes.

To conclude, choosing the best method for interpreting Random Forest remains problem-dependent. We recommend, in practice, comparing and combining different methods interpreting RF models. For instance, we have tested in this work combining Forest-ORE and STEL, which reduces the size and the length of the final rule ensemble and produces the best performance in some cases.

\section{Conclusion}
In this paper, we have proposed a new framework that aims to improve the interpretability of the random forest model. We first reviewed the relevant literature to present different approaches associated with RF interpretability. Various methods are reported in the literature, and those involving a representative rule ensemble are considered key to efficient interpretability and communication. The methodology that has been proposed in this article adheres to this vision. It is divided into four stages. The first stage extracts RF rules. The second reduces RF rule ensemble size based on rules' individual predictive performance and complexity. The third stage uses a mixed-integer optimization approach to select an optimal set of rules based on the trade-off between the rules' collective performance, coverage, and complexity. Finally, the fourth stage uses the metarules approach to provide complementary information to the rules formed in the third stage. 
\\This method has been illustrated through a simulated dataset and its robustness has been assessed over 36 benchmark datasets. The results show that this method outperforms RF and the other classifiers in predicting the covered instances. In addition, this method enables the best trade-off between predictive performance, rule set coverage, rule set size, and length of the rules. 
\\In future work, we plan to generalize this approach to other tree or rule ensembles and tackle the regression issues. We also aim to improve the approach's computational time and offer an ergonomic user interface for manipulating the different parameters associated with this method and visualizing the different results.

\section*{Declarations}

\begin{itemize}
\item Funding: Not applicable
\item Conflict of interest/Competing interests: On behalf of all authors, the corresponding author states that there is no conflict of interest.
\item Ethics approval: Not applicable
\item Consent to participate: Not applicable
\item Consent for publication: Not applicable
\item Availability of data and materials: All data used are publicly available in UCI Machine Learning repository and Keel datasets repository.
\item Code availability: The implementation and the computational work are done using the R language and environment for statistical computing, the Python programming language, and the Gurobi Optimizer Software (with a free academic licence). The code, the data files, and the resulting files of the benchmark reported in the article are available via Github: \url{https://github.com/HMAISSAE/Forest-ORE\_Bench\_2023}.
\item Authors\textsc{\char13} contributions: All authors contributed to the study conception and design. Material preparation and data collection and analysis were performed by Haddouchi Maissae. The first draft of the manuscript was written by Haddouchi Maissae and all authors commented on previous versions of the manuscript. All authors read and approved the final manuscript.
\item Human and Animal Ethics: Not applicable
\end{itemize}




\bibliography{biblio2023}

\begin{thebibliography}{95}
\providecommand{\natexlab}[1]{#1}
\providecommand{\url}[1]{{#1}}
\providecommand{\urlprefix}{URL }
\providecommand{\doi}[1]{\url{https://doi.org/#1}}
\providecommand{\eprint}[2][]{\url{#2}}
 \bibcommenthead

\bibitem[{Adnan and Islam(2016)}]{adnanOptimizingNumberTrees2016}
Adnan MN, Islam MZ (2016) Optimizing the number of trees in a decision forest
  to discover a subforest with high ensemble accuracy using a genetic
  algorithm. Knowledge-Based Systems 110:86--97.
  \doi{10.1016/j.knosys.2016.07.016}

\bibitem[{Agrawal et~al(1994)Agrawal, Srikant, Road, and
  Jose}]{agrawalFastAlgorithmsMining}
Agrawal R, Srikant R, Road H, et~al (1994) Fast {{Algorithms}} for {{Mining
  Association Rules}}. Proceedings of the 20th VLDB Conference Santiago, Chile
  p~13

\bibitem[{Akyuz and Birbil(2021)}]{akyuz_discovering_2021}
Akyuz MH, Birbil SI (2021) Discovering {Classification} {Rules} for
  {Interpretable} {Learning} with {Linear} {Programming}. ArXiv:2104.10751 [cs,
  stat]

\bibitem[{Alsallakh et~al(2013)Alsallakh, Aigner, Miksch, and
  Hauser}]{alsallakhRadialSetsInteractive2013}
Alsallakh B, Aigner W, Miksch S, et~al (2013) Radial {{Sets}}: Interactive
  {{Visual Analysis}} of {{Large Overlapping Sets}}. IEEE Transactions on
  Visualization and Computer Graphics 19(12):2496--2505.
  \doi{10.1109/TVCG.2013.184}

\bibitem[{Aria et~al(2021)Aria, Cuccurullo, and
  Gnasso}]{ariaComparisonInterpretativeProposals2021}
Aria M, Cuccurullo C, Gnasso A (2021) A comparison among interpretative
  proposals for {{Random Forests}}. Machine Learning with Applications
  6:100,094. \doi{10.1016/j.mlwa.2021.100094}

\bibitem[{Azmi et~al(2019)Azmi, Runger, and Berrado}]{azmi_interpretable_2019}
Azmi M, Runger GC, Berrado A (2019) Interpretable regularized class association
  rules algorithm for classification in a categorical data space. Information
  Sciences 483:313--331. \doi{10.1016/j.ins.2019.01.047}

\bibitem[{Baehrens et~al(2010)Baehrens, Schroeter, Harmeling, Kawanabe, Hansen,
  and {ller}}]{baehrensHowExplainIndividual2010}
Baehrens D, Schroeter T, Harmeling S, et~al (2010) How to {{Explain Individual
  Classification Decisions}} p~29

\bibitem[{Bay and Pazzani(2001)}]{bayDetectingGroupDifferences2001}
Bay SD, Pazzani MJ (2001) Detecting {{Group Differences}}: Mining {{Contrast
  Sets}}. Data Mining and Knowledge Discovery 5(3):213--246.
  \doi{10.1023/A:1011429418057}

\bibitem[{Bayardo et~al(1999)Bayardo, Agrawal, and
  Gunopulos}]{bayardoConstraintbasedRuleMining1999}
Bayardo R, Agrawal R, Gunopulos D (1999) Constraint-based rule mining in large,
  dense databases. In: Proceedings 15th {{International Conference}} on {{Data
  Engineering}} ({{Cat}}. {{No}}.{{99CB36337}}). {IEEE}, {Sydney, NSW,
  Australia}, pp 188--197, \doi{10.1109/ICDE.1999.754924}

\bibitem[{Beckett(2018)}]{beckettRfvizInteractiveVisualization2018}
Beckett C (2018) Rfviz: An {{Interactive Visualization Package}} for {{Random
  Forests}} in {{R}}. All {{Graduate Plan B}} and Other {{Reports}} 1335

\bibitem[{B{\'e}nard et~al(2020)B{\'e}nard, Biau, Da~Veiga, and
  Scornet}]{benardSIRUSStableInterpretable2020}
B{\'e}nard C, Biau G, Da~Veiga S, et~al (2020) {{SIRUS}}: Stable and
  {{Interpretable RUle Set}} for {{Classification}}

\bibitem[{Benavoli et~al(2016)Benavoli, Corani, and
  Mangili}]{benavoli_should_2016}
Benavoli A, Corani G, Mangili F (2016) Should {We} {Really} {Use} {Post}-{Hoc}
  {Tests} {Based} on {Mean}-{Ranks}? Journal of Machine Learning Research 17
  (2016) p~10

\bibitem[{Bernard et~al(2008)Bernard, Heutte, and
  Adam}]{bernardSelectionDecisionTrees2008}
Bernard S, Heutte L, Adam S (2008) On the selection of decision trees in
  {{Random Forests}} pp 302--307. \doi{10.1109/IJCNN.2009.5178693}

\bibitem[{Berrado and Runger(2007)}]{berradoUsingMetarulesOrganize2007}
Berrado A, Runger GC (2007) Using metarules to organize and group discovered
  association rules. Data Mining and Knowledge Discovery 14(3):409--431.
  \doi{10.1007/s10618-006-0062-6}

\bibitem[{Biau and Scornet(2016)}]{biauRandomForestGuided2016}
Biau G, Scornet E (2016) A random forest guided tour. TEST 25(2):197--227.
  \doi{10.1007/s11749-016-0481-7}

\bibitem[{Birbil et~al(2020)Birbil, Edali, and
  Yuceoglu}]{birbilRuleCoveringInterpretation2020a}
Birbil SI, Edali M, Yuceoglu B (2020) Rule {{Covering}} for {{Interpretation}}
  and {{Boosting}}. arXiv:200706379 [cs, stat]
  {\href{https://arxiv.org/abs/2007.06379}{{https://arxiv.org/abs/arXiv:2007.06379}}}
  {[cs, stat]}

\bibitem[{{Blanco-Justicia} et~al(2020){Blanco-Justicia}, {Domingo-Ferrer},
  Mart{\'i}nez, and
  S{\'a}nchez}]{blanco-justiciaMachineLearningExplainability2020}
{Blanco-Justicia} A, {Domingo-Ferrer} J, Mart{\'i}nez S, et~al (2020) Machine
  learning explainability via microaggregation and shallow decision trees.
  Knowledge-Based Systems 194:105,532. \doi{10.1016/j.knosys.2020.105532}

\bibitem[{Breiman(2001)}]{breimanRandomForests2001}
Breiman L (2001) Random {{Forests}}. Machine Learning 45(1):5--32.
  \doi{10.1023/A:1010933404324}

\bibitem[{Breiman(2002)}]{breimanWALDLECTUREII2002}
Breiman L (2002) {{Wald lecture II looking inside the black box }} p~35

\bibitem[{Breiman et~al(1884)Breiman, Friedman, Olshen, and
  Stone}]{breimanCLASSIFICATIONREGRESSIONTREES1884}
Breiman L, Friedman JH, Olshen RA, et~al (1884) {{Classification And Regression
  Trees }}, the wadsworth statistics/probability series edn. Monterey, {{CA}} :
  Wadsworth \& {{Brooks}}/{{Cole Advanced Books}} \& {{Software}}, 1984. - 358
  p.

\bibitem[{Carrizosa et~al(2021)Carrizosa, {Molero-R{\'i}o}, and
  Romero~Morales}]{carrizosaMathematicalOptimizationClassification2021}
Carrizosa E, {Molero-R{\'i}o} C, Romero~Morales D (2021) Mathematical
  optimization in classification and regression trees. TOP 29(1):5--33.
  \doi{10.1007/s11750-021-00594-1}

\bibitem[{Caruana et~al(2015)Caruana, Lou, Gehrke, Koch, Sturm, and
  Elhadad}]{caruanaIntelligibleModelsHealthCare2015b}
Caruana R, Lou Y, Gehrke J, et~al (2015) Intelligible {{Models}} for
  {{HealthCare}}: Predicting {{Pneumonia Risk}} and {{Hospital}} 30-day
  {{Readmission}} pp 1721--1730. \doi{10.1145/2783258.2788613}

\bibitem[{Casiraghi et~al(2020)Casiraghi, Malchiodi, Trucco, Frasca,
  Cappelletti, Fontana, Esposito, Avola, Jachetti, Reese, Rizzi, Robinson, and
  Valentini}]{casiraghiExplainableMachineLearning2020}
Casiraghi E, Malchiodi D, Trucco G, et~al (2020) Explainable {{Machine
  Learning}} for {{Early Assessment}} of {{COVID}}-19 {{Risk Prediction}} in
  {{Emergency Departments}}. IEEE Access 8:196,299--196,325.
  \doi{10.1109/ACCESS.2020.3034032}

\bibitem[{Chen et~al(2017)Chen, Li, Tang, Bilal, Yu, Weng, and
  Li}]{chenParallelRandomForest2017}
Chen J, Li K, Tang Z, et~al (2017) A {{Parallel Random Forest Algorithm}} for
  {{Big Data}} in a {{Spark Cloud Computing Environment}}. IEEE Transactions on
  Parallel and Distributed Systems 28(4):919--933.
  \doi{10.1109/TPDS.2016.2603511},
  {\href{https://arxiv.org/abs/1810.07748}{{https://arxiv.org/abs/arXiv:1810.07748}}}

\bibitem[{Cohen(1960)}]{cohenCoefficientAgreementNominal1960b}
Cohen J (1960) A {{Coefficient}} of {{Agreement}} for {{Nominal Scales}}.
  Educational and Psychological Measurement 20(1):37--46.
  \doi{10.1177/001316446002000104}

\bibitem[{Cohen(1995)}]{cohen_fast_1995}
Cohen WW (1995) Fast {Effective} {Rule} {Induction}. In: Prieditis A, Russell S
  (eds) Machine {Learning} {Proceedings} 1995. Morgan Kaufmann, San Francisco
  (CA), p 115--123, \doi{10.1016/B978-1-55860-377-6.50023-2},
  \urlprefix\url{https://www.sciencedirect.com/science/article/pii/B9781558603776500232}

\bibitem[{Cutler et~al(2007)Cutler, Edwards, Beard, Cutler, Hess, Gibson, and
  Lawler}]{cutlerRANDOMFORESTSCLASSIFICATION2007}
Cutler DR, Edwards TC, Beard KH, et~al (2007) {{Random forests for
  classification in ecology }}. Ecology 88(11):2783--2792.
  \doi{10.1890/07-0539.1}

\bibitem[{Das et~al(2020)Das, Agarwal, Venugopal, Sheldon, and
  Shiva}]{dasTaxonomySurveyInterpretable2020}
Das S, Agarwal N, Venugopal D, et~al (2020) Taxonomy and {{Survey}} of
  {{Interpretable Machine Learning Method}} pp 670--677.
  \doi{10.1109/SSCI47803.2020.9308404}

\bibitem[{Dash et~al(2018)Dash, Gunluk, and Wei}]{dash_boolean_2018}
Dash S, Gunluk O, Wei D (2018) Boolean {Decision} {Rules} via {Column}
  {Generation}. In: Advances in {Neural} {Information} {Processing} {Systems},
  vol~31. Curran Associates, Inc.,
  \urlprefix\url{https://proceedings.neurips.cc/paper/2018/hash/743394beff4b1282ba735e5e3723ed74-Abstract.html}

\bibitem[{Deng(2019)}]{dengInterpretingTreeEnsembles2019b}
Deng H (2019) Interpreting tree ensembles with {{inTrees}}. International
  Journal of Data Science and Analytics 7(4):277--287.
  \doi{10.1007/s41060-018-0144-8}

\bibitem[{Deng et~al(2014)Deng, Runger, Tuv, and
  Bannister}]{dengCBCAssociativeClassifier2014}
Deng H, Runger G, Tuv E, et~al (2014) {{CBC}}: An associative classifier with a
  small number of rules. Decision Support Systems 59:163--170.
  \doi{10.1016/j.dss.2013.11.004}

\bibitem[{Dua and Graff(2017)}]{Dua:2019}
Dua D, Graff C (2017) {{UCI}} machine learning repository

\bibitem[{Dubitzky et~al(2007)Dubitzky, Granzow, and
  Berrar}]{dubitzkyFundamentalsDataMining2007}
Dubitzky W, Granzow M, Berrar DP (2007) Fundamentals of {{Data Mining}} in
  {{Genomics}} and {{Proteomics}}. {Springer Science \& Business Media}

\bibitem[{Ehrlinger(2015)}]{ehrlingerGgRandomForestsVisuallyExploring2015}
Ehrlinger J (2015) {{ggRandomForests}}: Visually {{Exploring}} a {{Random
  Forest}} for {{Regression}}. arXiv:150107196 [stat]
  {\href{https://arxiv.org/abs/1501.07196}{{https://arxiv.org/abs/arXiv:1501.07196}}}
  {[stat]}

\bibitem[{Eskandarian et~al(2017)Eskandarian, Bahrami, and
  Kazemi}]{eskandarianComprehensiveDataMining2017}
Eskandarian S, Bahrami P, Kazemi P (2017) A comprehensive data mining approach
  to estimate the rate of penetration: Application of neural network, rule
  based models and feature ranking. Journal of Petroleum Science and
  Engineering 156:605--615. \doi{10.1016/j.petrol.2017.06.039}

\bibitem[{Fletcher and Islam(2018)}]{fletcherComparingSetsPatterns2018}
Fletcher S, Islam MZ (2018) Comparing sets of patterns with the {{Jaccard}}
  index. Australasian Journal of Information Systems 22.
  \doi{10.3127/ajis.v22i0.1538}

\bibitem[{Friedman(2000)}]{friedmanGreedyFunctionApproximation2000}
Friedman JH (2000) Greedy {{Function Approximation}}: A {{Gradient Boosting
  Machine}}. Annals of Statistics 29:1189--1232

\bibitem[{Friedman and Popescu(2008)}]{friedmanPredictiveLearningRule2008}
Friedman JH, Popescu BE (2008) Predictive learning via rule ensembles. The
  Annals of Applied Statistics 2(3):916--954. \doi{10.1214/07-AOAS148}

\bibitem[{Garcia et~al(2013)Garcia, Luengo, S{\'a}ez, L{\'o}pez, and
  Herrera}]{garciaSurveyDiscretizationTechniques2013}
Garcia S, Luengo J, S{\'a}ez JA, et~al (2013) A {{Survey}} of {{Discretization
  Techniques}}: Taxonomy and {{Empirical Analysis}} in {{Supervised Learning}}.
  IEEE Transactions on Knowledge and Data Engineering 25(4):734--750.
  \doi{10.1109/TKDE.2012.35}

\bibitem[{Gargett and Barnden(2015)}]{gargettModelingInteractionSensory2015}
Gargett A, Barnden J (2015) Modeling the interaction between sensory and
  affective meanings for detecting metaphor. In: Proceedings of the {{Third
  Workshop}} on {{Metaphor}} in {{NLP}}. {Association for Computational
  Linguistics}, {Denver, Colorado}, pp 21--30

\bibitem[{Ghannam and Techtmann(2021)}]{ghannamMachineLearningApplications2021}
Ghannam R, Techtmann S (2021) Machine learning applications in microbial
  ecology, human microbiome studies, and environmental monitoring.
  Computational and Structural Biotechnology Journal 19.
  \doi{10.1016/j.csbj.2021.01.028}

\bibitem[{Golino and Gomes(2014)}]{golinoVisualizingRandomForest2014}
Golino HF, Gomes CMA (2014) Visualizing {{Random Forest}}'s {{Prediction
  Results}}. Psychology 05(19):2084--2098. \doi{10.4236/psych.2014.519211}

\bibitem[{Goodman and Flaxman(2017)}]{goodmanEuropeanUnionRegulations2017}
Goodman B, Flaxman S (2017) European {{Union}} regulations on algorithmic
  decision-making and a "right to explanation". AI Magazine 38(3):50.
  \doi{10.1609/aimag.v38i3.2741}

\bibitem[{Guidotti et~al(2018)Guidotti, Monreale, Ruggieri, Pedreschi, Turini,
  and Giannotti}]{guidottiLocalRuleBasedExplanations2018}
Guidotti R, Monreale A, Ruggieri S, et~al (2018) Local {{Rule}}-{{Based
  Explanations}} of {{Black Box Decision Systems}}. arXiv:180510820 [cs]
  {\href{https://arxiv.org/abs/1805.10820}{{https://arxiv.org/abs/arXiv:1805.10820}}}
  {[cs]}

\bibitem[{{Gurobi Optimization,
  LLC}(2021)}]{gurobioptimizationllcGurobiOptimizerReference2021}
{Gurobi Optimization, LLC} (2021) Gurobi {{Optimizer Reference Manual}}

\bibitem[{Haddouchi and
  Berrado(2018)}]{haddouchiAssessingInterpretationCapacity2018a}
Haddouchi H, Berrado A (2018) Assessing interpretation capacity in {{Machine
  Learning}}: A critical review pp 1--6. \doi{10.1145/3289402.3289549}

\bibitem[{Haddouchi and Berrado(2019)}]{maissaehaddouchiSurveyMethodsTools2019}
Haddouchi M, Berrado A (2019) A survey of methods and tools used for
  interpreting {{Random Forest}} pp 1--6. \doi{10.1109/ICSSD47982.2019.9002770}

\bibitem[{Haddouchi and Berrado(2022)}]{maissae_novel_2022}
Haddouchi M, Berrado A (2022) A novel approach for discretizing continuous
  attributes based on tree ensemble and moment matching optimization.
  International Journal of Data Science and Analytics
  \doi{10.1007/s41060-022-00316-1}

\bibitem[{{Haibo He} and Garcia(2009)}]{haiboheLearningImbalancedData2009}
{Haibo He}, Garcia E (2009) Learning from {{Imbalanced Data}}. IEEE
  Transactions on Knowledge and Data Engineering 21(9):1263--1284.
  \doi{10.1109/TKDE.2008.239}

\bibitem[{Hara and Hayashi(2017)}]{haraMakingTreeEnsembles2017}
Hara S, Hayashi K (2017) Making {{Tree Ensembles Interpretable}}: A {{Bayesian
  Model Selection Approach}}. arXiv:160609066 [stat]
  {\href{https://arxiv.org/abs/1606.09066}{{https://arxiv.org/abs/arXiv:1606.09066}}}
  {[stat]}

\bibitem[{Hastie et~al(2009)Hastie, Tibshirani, and
  Friedman}]{tibshirani_valerie_nodate}
Hastie T, Tibshirani R, Friedman J (2009) The elements of statistical learning:
  data mining, inference and prediction. Springer, 2 edition

\bibitem[{Ho et~al(2021)Ho, Tan, Sze, Wong, and Goh}]{ho_what_2021}
Ho SY, Tan S, Sze CC, et~al (2021) What can {Venn} diagrams teach us about
  doing data science better? International Journal of Data Science and
  Analytics 11(1):1--10. \doi{10.1007/s41060-020-00230-4}

\bibitem[{{J. Alcal\'a-Fdez} et~al(2011){J. Alcal\'a-Fdez}, Fernandez, Luengo,
  Derrac, Garc{\'i}a, S{\'a}nchez, and
  Herrera}]{j.alcala-fdezKEELDataMiningSoftware2011}
{J. Alcal\'a-Fdez}, Fernandez A, Luengo J, et~al (2011) {{KEEL Data}}-{{Mining
  Software Tool}}: Data {{Set Repository}}, {{Integration}} of algorithms and
  {{Experimental}} analysis {{Framework}}. Journal of Multiple-Valued Logic and
  Soft Computing 17:2-3 pp 255--287

\bibitem[{Ke(2021)}]{keIntegratingGutMicrobiome2021}
Ke S (2021) Integrating gut microbiome and host immune markers to understand
  the pathogenesis of {{Clostridioides}} difficile infection. Gut Microbes
  2021, VOL. 13:1--18. \doi{10.1080/19490976.2021.1935186}

\bibitem[{Khalid et~al(2015)Khalid, Tuszynski, Szlek, Jachowicz, and
  Mendyk}]{khalidBlackBoxTransparentComputational2015}
Khalid MH, Tuszynski PK, Szlek J, et~al (2015) From {{Black}}-{{Box}} to
  {{Transparent Computational Intelligence Models}}: A {{Pharmaceutical Case
  Study}} pp 114--118. \doi{10.1109/FIT.2015.30}

\bibitem[{Khan et~al(2020)Khan, Gul, Perperoglou, Miftahuddin, Mahmoud, Adler,
  and Lausen}]{khanEnsembleOptimalTrees2020}
Khan Z, Gul A, Perperoglou A, et~al (2020) Ensemble of optimal trees, random
  forest and random projection ensemble classification. Advances in Data
  Analysis and Classification 14(1):97--116. \doi{10.1007/s11634-019-00364-9}

\bibitem[{Lakkaraju et~al(2016)Lakkaraju, Bach, and
  Leskovec}]{lakkaraju_interpretable_2016}
Lakkaraju H, Bach SH, Leskovec J (2016) Interpretable {Decision} {Sets}: {A}
  {Joint} {Framework} for {Description} and {Prediction}. In: Proceedings of
  the 22nd {ACM} {SIGKDD} {International} {Conference} on {Knowledge}
  {Discovery} and {Data} {Mining}. ACM, San Francisco California USA, pp
  1675--1684, \doi{10.1145/2939672.2939874}

\bibitem[{Landis and Koch(1977)}]{landisMeasurementObserverAgreement1977}
Landis JR, Koch GG (1977) The {{Measurement}} of {{Observer Agreement}} for
  {{Categorical Data}}. Biometrics 33(1):159. \doi{10.2307/2529310}

\bibitem[{Latinne et~al(2001)Latinne, Debeir, and
  Decaestecker}]{latinneLimitingNumberTrees2001a}
Latinne P, Debeir O, Decaestecker C (2001) Limiting the {{Number}} of {{Trees}}
  in {{Random Forests}}. In: Goos G, Hartmanis J, {van Leeuwen} J, et~al (eds)
  Multiple {{Classifier Systems}}, vol 2096. {Springer Berlin Heidelberg},
  {Berlin, Heidelberg}, p 178--187, \doi{10.1007/3-540-48219-9_18}

\bibitem[{Letham et~al(2015)Letham, Rudin, McCormick, and
  Madigan}]{letham_interpretable_2015}
Letham B, Rudin C, McCormick TH, et~al (2015) Interpretable classifiers using
  rules and {Bayesian} analysis: {Building} a better stroke prediction model.
  The Annals of Applied Statistics 9(3). \doi{10.1214/15-AOAS848},
  \urlprefix\url{http://arxiv.org/abs/1511.01644}, arXiv:1511.01644 [cs, stat]

\bibitem[{Lex et~al(2014)Lex, Gehlenborg, Strobelt, Vuillemot, and
  Pfister}]{lexUpSetVisualizationIntersecting2014b}
Lex A, Gehlenborg N, Strobelt H, et~al (2014) {{UpSet}}: Visualization of
  {{Intersecting Sets}}. IEEE Transactions on Visualization and Computer
  Graphics 20(12):1983--1992. \doi{10.1109/TVCG.2014.2346248}

\bibitem[{Liaw and Wiener(2002)}]{liawClassificationRegressionRandomForest2002}
Liaw A, Wiener M (2002) Classification and {{Regression}} by {{randomForest}}
  2:5

\bibitem[{Liu et~al(1999)Liu, Hsu, and
  Ma}]{liuIntegratingClassificationAssociation1999}
Liu B, Hsu W, Ma Y (1999) Integrating {{Classification}} and {{Association Rule
  Mining}}. KDD-98 Proceedings p~7

\bibitem[{Lundberg et~al(2020)Lundberg, Erion, Chen, DeGrave, Prutkin, Nair,
  Katz, Himmelfarb, Bansal, and Lee}]{lundbergLocalExplanationsGlobal2020}
Lundberg SM, Erion G, Chen H, et~al (2020) From local explanations to global
  understanding with explainable {{AI}} for trees. Nature Machine Intelligence
  2(1):56--67. \doi{10.1038/s42256-019-0138-9}

\bibitem[{M.~Jones and J.~Linder(2016)}]{m.jonesEdarfExploratoryData2016b}
M.~Jones Z, J.~Linder F (2016) Edarf: Exploratory {{Data Analysis}} using
  {{Random Forests}}. The Journal of Open Source Software 1(6):92.
  \doi{10.21105/joss.00092}

\bibitem[{Mashayekhi and Gras(2015)}]{mashayekhiRuleExtractionRandom2015d}
Mashayekhi M, Gras R (2015) Rule {{Extraction}} from {{Random Forest}}: The
  {{RF}}+{{HC Methods}}. In: Kanade T, Kittler J, Kleinberg JM, et~al (eds)
  Advances in {{Artificial Intelligence}}, vol 3060. {Springer Berlin
  Heidelberg}, {Berlin, Heidelberg}, p 223--237, \doi{10}

\bibitem[{Mashayekhi and Gras(2017)}]{mashayekhiRuleExtractionDecision2017c}
Mashayekhi M, Gras R (2017) Rule {{Extraction}} from {{Decision Trees
  Ensembles}}: New {{Algorithms Based}} on {{Heuristic Search}} and {{Sparse
  Group Lasso Methods}}. International Journal of Information Technology \&
  Decision Making 16(06):1707--1727. \doi{10.1142/S0219622017500055}

\bibitem[{Md~Nasim~Adnan(2017)}]{mdnasimadnanForExNewFramework2017}
Md~Nasim~Adnan MZI (2017) {{ForEx}}++: A {{New Framework}} for {{Knowledge
  Discovery}} from {{Decision Forests}}. Australasian Journal of Information
  Systems

\bibitem[{Meinshausen(2010)}]{meinshausenNodeHarvest2010a}
Meinshausen N (2010) Node harvest. The Annals of Applied Statistics
  4(4):2049--2072. \doi{10.1214/10-AOAS367},
  {\href{https://arxiv.org/abs/0910.2145}{{https://arxiv.org/abs/arXiv:0910.2145}}}

\bibitem[{Miraboutalebi et~al(2016)Miraboutalebi, Kazemi, and
  Bahrami}]{miraboutalebiFattyAcidMethyl2016}
Miraboutalebi SM, Kazemi P, Bahrami P (2016) Fatty {{Acid Methyl Ester}}
  ({{FAME}}) composition used for estimation of biodiesel cetane number
  employing random forest and artificial neural networks: A new approach. Fuel
  166:143--151. \doi{10.1016/j.fuel.2015.10.118}

\bibitem[{Mollas et~al(2022)Mollas, Bassiliades, and
  Tsoumakas}]{mollas_conclusive_2022}
Mollas I, Bassiliades N, Tsoumakas G (2022) Conclusive local interpretation
  rules for random forests. Data Mining and Knowledge Discovery
  \doi{10.1007/s10618-022-00839-y}

\bibitem[{Molnar(2022)}]{molnar_interpretable_nodate}
Molnar C (2022) Interpretable {Machine} {Learning}, 2nd edn. A {Guide} for
  {Making} {Black} {Box} {Models} {Explainable}

\bibitem[{Narayanan et~al(2016)Narayanan, Vaid, Wang, Jeon, Sharma, Caulfield,
  Sivasubramaniam, Cutler, Liu, and
  Khessib}]{narayananSSDFailuresDatacenters2016}
Narayanan I, Vaid K, Wang D, et~al (2016) {{SSD Failures}} in {{Datacenters}}:
  What? when? and {{Why}}? pp 1--11. \doi{10.1145/2928275.2928278}

\bibitem[{Palczewska et~al(2013)Palczewska, Palczewski, Robinson, and
  Neagu}]{palczewskaInterpretingRandomForest2013}
Palczewska A, Palczewski J, Robinson RM, et~al (2013) Interpreting random
  forest models using a feature contribution method pp 112--119.
  \doi{10.1109/IRI.2013.6642461}

\bibitem[{Paluszy(2017)}]{paluszyFacultyMathematicsInformatics2017}
Paluszy A (2017) Faculty of {{Mathematics}}, {{Informatics}} and {{Mechanics}}.
  PhD thesis

\bibitem[{Parr and Wilson(2021)}]{parr_partial_2021}
Parr T, Wilson JD (2021) Partial dependence through stratification. Machine
  Learning with Applications 6:100,146. \doi{10.1016/j.mlwa.2021.100146}

\bibitem[{Phung et~al(2015)Phung, Chau, and Phung}]{phungExtractingRuleRF2015a}
Phung LTK, Chau VTN, Phung NH (2015) Extracting {{Rule RF}} in {{Educational
  Data Classification}}: From a {{Random Forest}} to {{Interpretable Refined
  Rules}} pp 20--27. \doi{10.1109/ACOMP.2015.13}

\bibitem[{P{\l}o{\'n}ski and
  Zaremba(2014)}]{plonskiVisualizingRandomForest2014}
P{\l}o{\'n}ski P, Zaremba K (2014) Visualizing {{Random Forest}} with
  {{Self}}-{{Organising Map}}. arXiv:14056684 [cs] 8468:63--71.
  \doi{10.1007/978-3-319-07176-3_6} {[cs]}

\bibitem[{{Python Core Team}(2019)}]{pythoncoreteamPythonDynamicOpen2019}
{Python Core Team} (2019) Python: A dynamic, open source programming language.
  Python Software Foundation

\bibitem[{Quach(2012)}]{quachInteractiveRandomForests2012}
Quach AT (2012) Interactive {{Random Forests Plots}}. All {{Graduate Plan B}}
  and Other {{Reports}} 134

\bibitem[{{R Core Team}(2019)}]{rcoreteamLanguageEnvironmentStatistical2019}
{R Core Team} (2019) R: A Language and Environment for Statistical Computing.
  {Vienna, Austria}

\bibitem[{Ribeiro et~al(2016)Ribeiro, Singh, and
  Guestrin}]{ribeiroWhyShouldTrust2016b}
Ribeiro MT, Singh S, Guestrin C (2016) "{{Why Should I Trust You}}?":
  Explaining the {{Predictions}} of {{Any Classifier}}. arXiv:160204938 [cs,
  stat]
  {\href{https://arxiv.org/abs/1602.04938}{{https://arxiv.org/abs/arXiv:1602.04938}}}
  {[cs, stat]}

\bibitem[{Sagi and Rokach(2018)}]{sagiEnsembleLearningSurvey2018a}
Sagi O, Rokach L (2018) Ensemble learning: A survey. WIREs Data Mining and
  Knowledge Discovery 8(4). \doi{10.1002/widm.1249}

\bibitem[{Singh et~al(2016)Singh, Ribeiro, and
  Guestrin}]{singhProgramsBlackBoxExplanations2016}
Singh S, Ribeiro MT, Guestrin C (2016) Programs as {{Black}}-{{Box
  Explanations}}. arXiv:161107579 [cs, stat]
  {\href{https://arxiv.org/abs/1611.07579}{{https://arxiv.org/abs/arXiv:1611.07579}}}
  {[cs, stat]}

\bibitem[{Tan et~al(2020)Tan, Soloviev, Hooker, and
  Wells}]{tanTreeSpacePrototypes2020}
Tan S, Soloviev M, Hooker G, et~al (2020) Tree {{Space Prototypes}}: Another
  {{Look}} at {{Making Tree Ensembles Interpretable}}. arXiv:161107115 [cs,
  stat] \doi{10.1145/3412815.3416893} {[cs, stat]}

\bibitem[{Van~Assche and Blockeel(2008)}]{vanasscheSeeingForestTrees2008}
Van~Assche A, Blockeel H (2008) Seeing the {{Forest Through}} the {{Trees}}.
  In: Blockeel H, Ramon J, Shavlik J, et~al (eds) Inductive {{Logic
  Programming}}, vol 4894. {Springer Berlin Heidelberg}, {Berlin, Heidelberg},
  p 269--279, \doi{10.1007/978-3-540-78469-2_26}

\bibitem[{Wang et~al(2018)Wang, Lin, and
  Ho}]{wangDiscoveryCelltypeSpecific2018}
Wang X, Lin P, Ho JWK (2018) Discovery of cell-type specific {{DNA}} motif
  grammar in cis-regulatory elements using random {{Forest}}. BMC Genomics
  19(S1). \doi{10.1186/s12864-017-4340-z}

\bibitem[{Welling et~al(2016)Welling, Refsgaard, Brockhoff, and
  Clemmensen}]{wellingForestFloorVisualizations2016}
Welling SH, Refsgaard HHF, Brockhoff PB, et~al (2016) Forest {{Floor
  Visualizations}} of {{Random Forests}}. arXiv:160509196 [cs, stat]
  {\href{https://arxiv.org/abs/1605.09196}{{https://arxiv.org/abs/arXiv:1605.09196}}}
  {[cs, stat]}

\bibitem[{Wilcoxon(1945)}]{wilcoxonIndividualComparisonsRanking1945}
Wilcoxon F (1945) Individual {{Comparisons}} by {{Ranking Methods}}. Biometrics
  Bulletin 1(6):80. \doi{10.2307/3001968}

\bibitem[{Wilkinson(2012)}]{wilkinsonExactApproximateAreaProportional2012}
Wilkinson L (2012) Exact and {{Approximate Area}}-{{Proportional Circular
  Venn}} and {{Euler Diagrams}}. IEEE Transactions on Visualization and
  Computer Graphics 18(2):321--331. \doi{10.1109/TVCG.2011.56}

\bibitem[{Yang et~al(2012)Yang, Lu, Luo, and
  Li}]{yangMarginOptimizationBased2012}
Yang F, Lu Wh, Luo Lk, et~al (2012) Margin optimization based pruning for
  random forest. Neurocomputing 94:54--63. \doi{10.1016/j.neucom.2012.04.007}

\bibitem[{Yang et~al(2017)Yang, Rudin, and Seltzer}]{yang_scalable_2017}
Yang H, Rudin C, Seltzer M (2017) Scalable {Bayesian} {Rule} {Lists}. In:
  Proceedings of the 34th {International} {Conference} on {Machine} {Learning}.
  PMLR, pp 3921--3930,
  \urlprefix\url{https://proceedings.mlr.press/v70/yang17h.html}, iSSN:
  2640-3498

\bibitem[{Zhang et~al(2021)Zhang, Hou, Huang, Shi, Bride, Dong, and
  Gao}]{zhangExtractingOptimalExplanations2021}
Zhang G, Hou Z, Huang Y, et~al (2021) Extracting {{Optimal Explanations}} for
  {{Ensemble Trees}} via {{Logical Reasoning}}. arXiv:210302191 [cs]
  {\href{https://arxiv.org/abs/2103.02191}{{https://arxiv.org/abs/arXiv:2103.02191}}}
  {[cs]}

\bibitem[{Zhang and Wang(2009)}]{zhangSearchSmallestRandom2009}
Zhang H, Wang M (2009) Search for the smallest random forest. Statistics and
  its interface 2(3):381

\bibitem[{Zhao et~al(2019)Zhao, Wu, Lee, and
  Cui}]{zhaoIForestInterpretingRandom2019}
Zhao X, Wu Y, Lee DL, et~al (2019) {{iForest}}: Interpreting {{Random Forests}}
  via {{Visual Analytics}}. IEEE Transactions on Visualization and Computer
  Graphics 25(1):407--416. \doi{10.1109/TVCG.2018.2864475}

\end{thebibliography}



\section*{Appendices}
\appendix
\section{Illustration of the rule preselection stage outputs}
\label{appendix:PSo}
Table \ref{tab:rulemetrics} illustrates row in the RuleMetrics data frame, and Table \ref{tab:rulecov} illustrates a row in CovOk/CovNok data frames.
\begin{table}[h]
\vspace{-2ex}
  \centering
  \caption{Illustration of a row in the RuleMetrics data frame}
\resizebox{0.7\textwidth}{!}{%
    \begin{tabular}{L{0.1in}C{0.15in}C{0.15in}C{0.15in}C{0.2in}C{0.2in}C{0.2in}C{0.25in}C{0.2in}L{0.8in}}
    \toprule
    \textbf{Id} & \textbf{Conf.} & \textbf{Cov.} & \textbf{Att. nbr} & \textbf{Lev. nbr} & \textbf{Att. nbr\_S} & \textbf{Lev. nbr\_S} & \textbf{Att.} & \textbf{Ypred} & \textbf{Condition} \\
    \midrule
    1     & 0.986 & 0.594 & 2     & 6     & 0.286 & 0.122 & V6,V7 & 2   & X[,6] in \{1,2\} \& X[,7] in \{1,2,3,5\} \\
    \bottomrule
    \end{tabular}%
}
  \label{tab:rulemetrics}%
\vspace{-3ex}    
\end{table}%

\begin{table}[h]
  \centering
  \caption{Illustration of a row in CovOk/CovNok data frames}
\resizebox{0.5\textwidth}{!}{%
    \begin{tabular}{cccccccccc}
    \toprule
    \textbf{R1} & \textbf{R2} & \textbf{R3} & \textbf{R4} & \textbf{R5} & \textbf{R6} & \textbf{R7} & \textbf{R8} & \textbf{...} & \textbf{Rm} \\
    \midrule
    1     & 0     & 0     & 0     & 1     & 0     & 0     & 1     & ...   & 0 \\
    \bottomrule
    \end{tabular}%
}
   \label{tab:rulecov}%
\vspace{-2ex}    
\end{table}%

\section{MIP proofs}
\label{appendix:MIP}
\subsection{Proof 1}
\label{appendix:proof1}
Constraints ~\ref{C3}~ and ~\ref{C4}~ ensure that we predict an instance correctly if the sum of rules that predict it correctly is strictly greater than the sum of rules that mispredict it (vote):
\begin{center}
\newmaketag
\begin{small}
\begin{equation}
\begin{aligned}
&is\_error\left[i\right] = 0 \iff P_i\geq 1 ~and~\\&is\_error\left[i\right] = 1 \iff P_i\le 0
\end{aligned}
\end{equation}
\end{small}
\end{center}
Given constraint ~\ref{C2}~ and since $CovOk\left[i,j\right]$, $CovNok\left[i,j\right]$, and $is\_selected\left[j\right]$ $\in\left[0,1\right]$
then, ~$-maxcover\le Pi\le\ +maxcover$~ and~ $0\le C_i\ \le maxcover$~. Thus:
\\When $is\_error[i] = 0$ : constraint ~\ref{C3}~ gives $ P_i \le maxcover$, which is always true, and constraint ~\ref{C4}~ gives $P_i\geq1$.
When $is\_error[i] = 1$ : constraint ~\ref{C3}~ gives$~ P_i \le 0$, and constraint ~\ref{C4}~ gives ~$ P_i\geq -maxcover$, which is always true.
Given equation ~\ref{C3}~ and ~\ref{C4}~, an instance is considered to have been predicted incorrectly in two cases: Case 1- The instance is covered, and the rules prediction (voting) does not match the reel target value, or there is a tie. Case 2- The instance is not covered. This is why, in equation ~\ref{C5}, we subtract the instances corresponding to case 2. 
\subsection{Proof 2}
\label{appendix:proof2}
Constraints \ref{C6} and \ref{C7} ensure that an instance i is covered if at least one rule covers it (correctly or incorrectly):
\begin{center}
\newmaketag
\begin{small}
\begin{equation}
\begin{aligned}
&is\_covered\left[i\right] = 1 \iff C_i \geq 1~and~\\ &is\_covered\left[i\right] = 0 \iff C_i\le 0
\end{aligned}
\end{equation}
\end{small}
\end{center}
Given that $0\le C_i\ \le maxcover$, then:
\\When $is\_covered\left[i\right] = 0$ :  constraint ~\ref{C6}~ gives $C_i \le 0$, and constraint ~\ref{C7}~ gives $Ci\geq0$, which is always true.
When $ is\_covered\left[i\right] = 1$ : constraint ~\ref{C6}~ gives $C_i \le maxcover$, which is always true, and constraint ~\ref{C7}~ gives $C_i\geq 1$.

\subsection{Proof 3}
\label{appendix:proof3}
Constraints ~\ref{C9}~ and ~\ref{C10}~ ensure that an instance i is considered overlapping if it belongs to two rules or more:
\begin{center}
\newmaketag
\begin{small}
\begin{equation}
\begin{aligned}
&is\_overlap\left[i\right] = 1 \iff C_i \geq 2~and~\\& is\_overlap\left[i\right] = 0 \iff C_i\le 1
\end{aligned}
\end{equation}
\end{small}
\end{center}
Given that $0\le C_i\ \le maxcover$:
\\When $is\_overlap\left[i\right] = 0$ :  constraint ~\ref{C9}~ gives $C_i \le 1$, and constraint ~\ref{C10}~ gives $Ci\geq0$, which is always true.
When $ is\_overlap\left[i\right] = 1$ : constraint ~\ref{C9}~ gives $C_i \le maxcover$, which is always true, and constraint ~\ref{C10}~ gives $C_i\geq 2$.

\section{Illustration of the rule enrichment stage output: complementary rules dataframe}
\label{appendix:RE}
Table \ref{tab:Enr.dat} illustrates a containment  between rule (id=102) and rule (id 97). The 1${}^{st}$ column is reserved for the selected rules IDs. The 2${}^{nd}$ column reports the complementary rules IDs. The 3${}^{rd}$ column is reserved for the conditions of the rules in the 2${}^{nd}$ column. The 5${}^{th}$ column reports the containment rate of the 2${}^{nd}$ column rule in the 1${}^{st}$ column rule. The remaining columns relate the characteristics of the 2${}^{nd}$ column rules. In this table, each bold line represents the characteristics of a selected rule. 

\begin{table*}[hbt]
  \centering
 \caption{Illustration of the complementary rules dataframe}
\resizebox{0.8\textwidth}{!}{%
     \begin{tabular}{R{0.15in}R{0.15in}C{2in}C{0.25in}C{0.35in}C{0.2in}C{0.3in}C{0.3in}C{0.3in}C{0.1in}}
         \toprule
    ID SR & ID Rule & Condition & Ypred & Intersect & Att. & Att. nbr & Lev. nbr & Conf. & Cov. \\
    \midrule
    102   & \textbf{102} & \textbf{X[,1] in \{A2,A4\} \& X[,2] in \{B2\}} & \textbf{1} & \textbf{1.00} & \textbf{V1,V2} & \textbf{2} & \textbf{3} & \textbf{1} & \textbf{0.25} \\
    102   & 97    & X[,3] in \{C2\} \& X[,5] in \{E1,E4\}  & 1   & 0.96  & V3,V5 & 2     & 3     & 1     & 0.19 \\
    \bottomrule
    \end{tabular}%
}

  \label{tab:Enr.dat}%
\end{table*}%

\section{Illustration of rules overlaps using the Upset method on the Mushroom dataset}
\label{appendix:upsetmushroom}
\begin{figure}[h!]
\centering
  \includegraphics[width=0.6\textwidth]{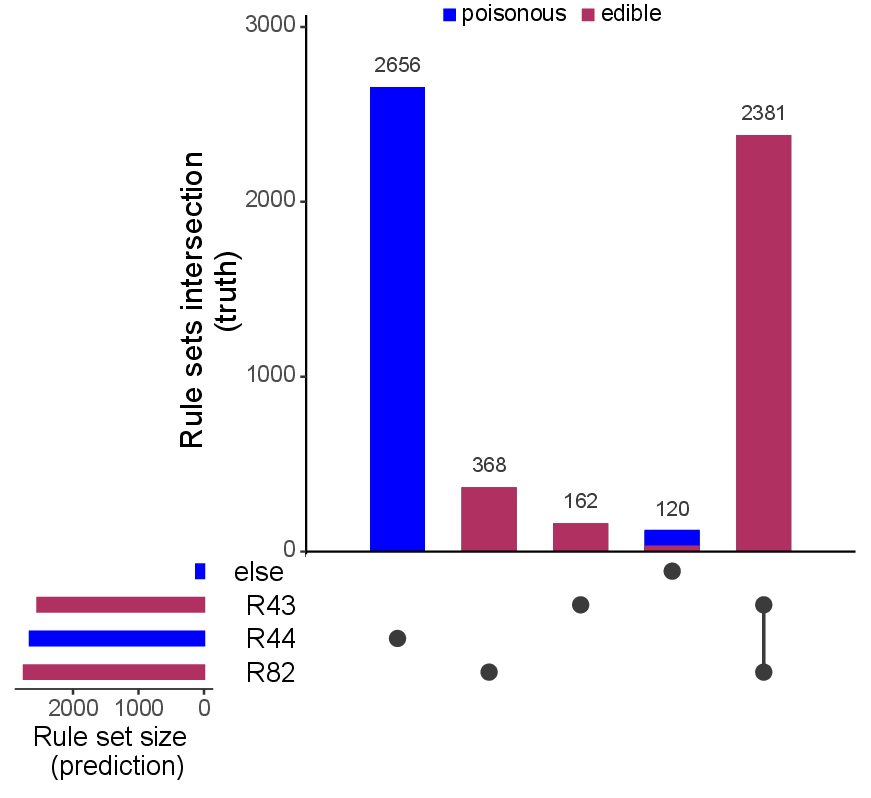}
\caption{Illustration of rules overlaps representation using the Upset method. Upset provides an efficient way to visualize intersections in the multiple sets explaining the Mushroom dataset. In this case, the overlap is between rules 82 et 43 and concerns the edible class. }
\label{fig:upsetMushroom}       
\end{figure}

We give the example of the Mushroom dataset (Figure \ref{fig:upsetMushroom}) to show an example of overlaps. The mushroom dataset includes 8124 instances and 23 descriptive variables. Table \ref{tab:SR_Mushroom_Forest_ORE} lists the selected rules resulting from applying Forest-ORE to the Mushroom dataset. In Figure \ref{fig:upsetMushroom}, the 2\textsuperscript{nd} vertical bar chart represents the size of instances applied exclusively to R82, the 3\textsuperscript{rd} to instances applied exclusively to R43, and the 5\textsuperscript{th} to the intersection between R43 and R82. This figure also shows that the ``else'' rule classifies instances as poisonous (horizontal bar chart), whereas, in fact, some of them are edible (vertical bar chart).
\begin{table*}[h!]
  \centering
  \caption{Selected rules provided by applying ``Forest-ORE'' to the Mushroom dataset}
\resizebox{1\textwidth}{!}{%
\centering
    \begin{tabular}{crrrrrp{11.5em}cc}
    \toprule
    \multicolumn{1}{l}{id} & \multicolumn{1}{l}{confidence} & \multicolumn{1}{l}{coverage} & \multicolumn{1}{l}{class\_coverage} & \multicolumn{1}{l}{att. nbr} & \multicolumn{1}{l}{lev. Nbr} & cond  & Ypred & att. \\
    \midrule
    43    & 1     & 0.45  & 0.86  & 4     & 18    & X[,5] in \{a,l,n\} \& \newline{}X[,15] in \{g,n,o,p,w\} \& \newline{}X[,18] in \{n,o\} \& \newline{}X[,20] in \{b,h,k,n,o,r,u,y\} & 'e'   & V5,V15,V18,V20 \\
   44    & 1     & 0.47  & 0.97  & 1     & 6     & \multicolumn{1}{l}{X[,5] in \{c,f,m,p,s,y\}} & 'p'   & V5 \\
   82    & 1     & 0.48  & 0.93  & 3     & 16    & X[,8] in \{b\} \& \newline{}X[,15] in \{b,e,g,n,o,p,w,y\} \& \newline{}X[,20] in \{b,k,n,o,u,w,y\} & 'e'   & V8,V15,V20 \\
     \bottomrule
    \end{tabular}%
}
  \label{tab:SR_Mushroom_Forest_ORE}%
\end{table*}%

\section{Results per dataset}
\label{appendix:resultsperdata}
Tables \ref{tab:acc_bin_all}, \ref{tab:acc_bin_cov}, \ref{tab:coverage_bin} , \ref{tab:acc_multi_all}, \ref{tab:acc_multi_cov} , and \ref{tab:coverage_multi}  report the average accuracy and coverage metrics and their ranking per dataset, and per classification issue. Table \ref{tab:Friedman_test} reports the p-value of the Friedman test on these average results.
\begin{table*}[h!]
\centering
\caption{Average accuracy metrics and their ranking per dataset on binary classification (all instances)}
\resizebox{0.9\textwidth}{!}{%
    \begin{tabular}{lllllllllll}
   \toprule
          & RPART & STEL  & Pre-Forest-ORE & Forest-ORE & Forest-ORE+STEL & RF    & RIPPER & SBRL \\
    \midrule
    APPENDICITIS & 0.863 (4) & 0.847 (2.5) & 0.875 (7) & 0.866 (5.5) & 0.866 (5.5) & 0.878 (8) & 0.847 (2.5) & 0.819 (1) \\
    BANKNOTE & 0.966 (7) & 0.964 (5) & 0.956 (2) & 0.963 (3.5) & 0.963 (3.5) & 0.970 (8) & 0.965 (6) & 0.945 (1) \\
    BRCANCER & 0.941 (3) & 0.952 (6) & 0.959 (7) & 0.948 (5) & 0.945 (4) & 0.974 (8) & 0.939 (2) & 0.937 (1) \\
    CRYOTHERAPY & 0.707 (1) & 0.715 (2) & 0.756 (5.5) & 0.741 (4) & 0.763 (7) & 0.767 (8) & 0.730 (3) & 0.756 (5.5) \\
    HABERMAN & 0.733 (8) & 0.727 (4) & 0.729 (5) & 0.731 (7) & 0.730 (6) & 0.721 (3) & 0.720 (2) & 0.716 (1) \\
    HEART & 0.806 (6) & 0.786 (3.5) & 0.833 (8) & 0.786 (3.5) & 0.778 (1) & 0.828 (7) & 0.779 (2) & 0.798 (5) \\
    HYPOTHYROID & 0.981 (6) & 0.981 (6) & 0.978 (3) & 0.976 (1) & 0.977 (2) & 0.983 (8) & 0.980 (4) & 0.981 (6) \\
    INDIAN & 0.709 (5) & 0.707 (3.5) & 0.717 (7.5) & 0.707 (3.5) & 0.700 (2) & 0.717 (7.5) & 0.693 (1) & 0.715 (6) \\
    ION   & 0.899 (5) & 0.916 (6) & 0.924 (7) & 0.895 (4) & 0.894 (3) & 0.927 (8) & 0.882 (2) & 0.802 (1) \\
    MAMMOGRAPHIC & 0.824 (7) & 0.820 (4) & 0.825 (8) & 0.821 (5) & 0.818 (3) & 0.822 (6) & 0.800 (2) & 0.761 (1) \\
    MUSHROOM & 0.995 (2) & 0.996 (4) & 0.998 (5.5) & 0.995 (2) & 0.995 (2) & 0.998 (5.5) & 1.000 (7.5) & 1.000 (7.5) \\
    MUTAGENESIS & 0.685 (8) & 0.670 (3) & 0.674 (4) & 0.677 (6) & 0.676 (5) & 0.683 (7) & 0.669 (2) & 0.665 (1) \\
    PHONEME & 0.761 (2) & 0.755 (1) & 0.771 (3) & 0.772 (4) & 0.774 (5) & 0.787 (6) & 0.814 (8) & 0.803 (7) \\
    TICTACTOE & 0.908 (1) & 0.985 (7) & 0.914 (2) & 0.974 (4.5) & 0.974 (4.5) & 0.932 (3) & 0.976 (6) & 1.000 (8) \\
    VOTE  & 0.953 (2) & 0.946 (1) & 0.958 (5) & 0.960 (6.5) & 0.960 (6.5) & 0.968 (8) & 0.957 (3.5) & 0.957 (3.5) \\
    WDBC  & 0.944 (4.5) & 0.947 (6) & 0.960 (7) & 0.944 (4.5) & 0.942 (2) & 0.964 (8) & 0.943 (3) & 0.909 (1) \\
    WILT  & 0.951 (5) & 0.946 (2.5) & 0.952 (6.5) & 0.946 (2.5) & 0.946 (2.5) & 0.952 (6.5) & 0.946 (2.5) & 0.953 (8) \\
    WISCONSIN & 0.937 (1) & 0.953 (5.5) & 0.963 (7) & 0.950 (3) & 0.951 (4) & 0.971 (8) & 0.940 (2) & 0.953 (5.5) \\
    XOR   & 1.000 (5) & 1.000 (5) & 1.000 (5) & 1.000 (5) & 1.000 (5) & 0.904 (1) & 1.000 (5) & 1.000 (5) \\
    \midrule
    \textbf{Total Rank} & \textbf{82.5} & \textbf{77.5} & \textbf{105} & \textbf{80} & \textbf{73.5} & \textbf{124.5} & \textbf{66} & \textbf{75} \\
    \bottomrule
\end{tabular}%
    }
\label{tab:acc_bin_all}%
\end{table*}%

\begin{table*}[h!]
\centering
\caption{Average accuracy metrics and their ranking per dataset on binary classification (covered instances)}
\resizebox{0.9\textwidth}{!}{%
    \begin{tabular}{lllllllllll}
     \toprule
          & RPART & STEL  & Pre-Forest-ORE & Forest-ORE & Forest-ORE+STEL & RF    & RIPPER & SBRL \\
    \midrule
    APPENDICITIS & 0.863 (4) & 0.847 (3) & 0.875 (7) & 0.870 (5) & 0.873 (6) & 0.878 (8) & 0.695 (1) & 0.835 (2) \\
    BANKNOTE & 0.966 (4) & 0.955 (2) & 0.956 (3) & 0.967 (5.5) & 0.967 (5.5) & 0.970 (8) & 0.968 (7) & 0.934 (1) \\
    BRCANCER & 0.941 (2) & 0.960 (4) & 0.959 (3) & 0.966 (6) & 0.963 (5) & 0.974 (8) & 0.914 (1) & 0.968 (7) \\
    CRYOTHERAPY & 0.707 (1) & 0.720 (2) & 0.756 (4) & 0.753 (3) & 0.783 (6) & 0.767 (5) & 0.809 (7) & 0.900 (8) \\
    HABERMAN & 0.733 (6) & 0.729 (4.5) & 0.729 (4.5) & 0.740 (7) & 0.744 (8) & 0.721 (3) & 0.460 (1) & 0.712 (2) \\
    HEART & 0.806 (4) & 0.788 (2) & 0.833 (7) & 0.811 (5) & 0.802 (3) & 0.828 (6) & 0.771 (1) & 0.846 (8) \\
    HYPOTHYROID & 0.981 (4) & 0.986 (7) & 0.979 (3) & 0.985 (6) & 0.989 (8) & 0.983 (5) & 0.818 (2) & 0.776 (1) \\
    INDIAN & 0.709 (3) & 0.735 (8) & 0.720 (6) & 0.717 (4.5) & 0.727 (7) & 0.717 (4.5) & 0.431 (1) & 0.579 (2) \\
    ION   & 0.899 (3) & 0.921 (4.5) & 0.924 (7) & 0.922 (6) & 0.921 (4.5) & 0.927 (8) & 0.840 (1) & 0.885 (2) \\
    MAMMOGRAPHIC & 0.824 (4) & 0.840 (8) & 0.825 (5) & 0.831 (6) & 0.837 (7) & 0.822 (3) & 0.780 (2) & 0.769 (1) \\
    MUSHROOM & 0.995 (1.5) & 0.995 (1.5) & 0.998 (3.5) & 1.000 (6.5) & 1.000 (6.5) & 0.998 (3.5) & 1.000 (6.5) & 1.000 (6.5) \\
    MUTAGENESIS & 0.685 (8) & 0.673 (3) & 0.674 (4) & 0.679 (6) & 0.678 (5) & 0.683 (7) & 0.539 (1) & 0.660 (2) \\
    PHONEME & 0.761 (2) & 0.817 (8) & 0.772 (3) & 0.781 (4) & 0.786 (5) & 0.787 (6) & 0.683 (1) & 0.802 (7) \\
    TICTACTOE & 0.908 (1) & 1.000 (7.5) & 0.914 (2) & 0.993 (5) & 0.995 (6) & 0.932 (3) & 0.972 (4) & 1.000 (7.5) \\
    VOTE  & 0.953 (2) & 0.963 (5) & 0.958 (3.5) & 0.964 (6.5) & 0.964 (6.5) & 0.968 (8) & 0.930 (1) & 0.958 (3.5) \\
    WDBC  & 0.944 (3) & 0.951 (4) & 0.960 (7) & 0.957 (6) & 0.955 (5) & 0.964 (8) & 0.915 (1) & 0.929 (2) \\
    WILT  & 0.951 (3.5) & 0.961 (7) & 0.953 (6) & 0.951 (3.5) & 0.972 (8) & 0.952 (5) & 0.466 (1) & 0.936 (2) \\
    WISCONSIN & 0.937 (2) & 0.958 (3) & 0.963 (4) & 0.965 (5) & 0.966 (6) & 0.971 (7) & 0.897 (1) & 0.977 (8) \\
    XOR   & 1.000 (5) & 1.000 (5) & 1.000 (5) & 1.000 (5) & 1.000 (5) & 0.904 (1) & 1.000 (5) & 1.000 (5) \\
    \midrule
    \textbf{Total Rank} & \textbf{63} & \textbf{89} & \textbf{87.5} & \textbf{101.5} & \textbf{113} & \textbf{107} & \textbf{45.5} & \textbf{77.5} \\
    \bottomrule
\end{tabular}%
    }
\label{tab:acc_bin_cov}%
\end{table*}%

\begin{table*}[h!]
\centering
\caption{Average coverage metrics and their ranking per dataset on binary classification}
\resizebox{0.9\textwidth}{!}{%
    \begin{tabular}{lllllllllll}
   \toprule
          & RPART & STEL  & Pre-Forest-ORE & Forest-ORE & Forest-ORE+STEL & RF    & RIPPER & SBRL \\
    \midrule
    APPENDICITIS & 1.000 (7) & 0.872 (3) & 1.000 (7) & 0.984 (5) & 0.941 (4) & 1.000 (7) & 0.103 (1) & 0.512 (2) \\
    BANKNOTE & 1.000 (7) & 0.449 (2) & 1.000 (7) & 0.980 (4.5) & 0.980 (4.5) & 1.000 (7) & 0.437 (1) & 0.778 (3) \\
    BRCANCER & 1.000 (7) & 0.918 (3) & 1.000 (7) & 0.956 (4.5) & 0.956 (4.5) & 1.000 (7) & 0.344 (1) & 0.791 (2) \\
    CRYOTHERAPY & 1.000 (7) & 0.885 (3) & 1.000 (7) & 0.930 (5) & 0.900 (4) & 1.000 (7) & 0.367 (1) & 0.415 (2) \\
    HABERMAN & 1.000 (7) & 0.931 (3) & 1.000 (7) & 0.975 (5) & 0.936 (4) & 1.000 (7) & 0.113 (1) & 0.570 (2) \\
    HEART & 1.000 (7) & 0.978 (5) & 1.000 (7) & 0.914 (3.5) & 0.914 (3.5) & 1.000 (7) & 0.416 (1) & 0.764 (2) \\
    HYPOTHYROID & 1.000 (7.5) & 0.983 (5) & 0.998 (6) & 0.972 (4) & 0.962 (3) & 1.000 (7.5) & 0.043 (1) & 0.059 (2) \\
    INDIAN & 1.000 (7.5) & 0.847 (3) & 0.987 (6) & 0.937 (5) & 0.908 (4) & 1.000 (7.5) & 0.131 (1) & 0.589 (2) \\
    ION   & 1.000 (7) & 0.942 (3) & 1.000 (7) & 0.948 (5) & 0.947 (4) & 1.000 (7) & 0.366 (1) & 0.542 (2) \\
    MAMMOGRAPHIC & 1.000 (7) & 0.926 (3) & 1.000 (7) & 0.965 (5) & 0.938 (4) & 1.000 (7) & 0.472 (1) & 0.811 (2) \\
    MUSHROOM & 1.000 (7) & 0.657 (2) & 1.000 (7) & 0.982 (4.5) & 0.982 (4.5) & 1.000 (7) & 0.482 (1) & 0.808 (3) \\
    MUTAGENESIS & 1.000 (7) & 0.974 (3) & 1.000 (7) & 0.979 (5) & 0.976 (4) & 1.000 (7) & 0.066 (1) & 0.756 (2) \\
    PHONEME & 1.000 (7.5) & 0.657 (2) & 0.997 (6) & 0.971 (5) & 0.952 (4) & 1.000 (7.5) & 0.295 (1) & 0.938 (3) \\
    TICTACTOE & 1.000 (7) & 0.333 (1) & 1.000 (7) & 0.963 (5) & 0.961 (4) & 1.000 (7) & 0.344 (2) & 0.653 (3) \\
    VOTE  & 1.000 (7) & 0.752 (3) & 1.000 (7) & 0.981 (4.5) & 0.981 (4.5) & 1.000 (7) & 0.398 (1) & 0.637 (2) \\
    WDBC  & 1.000 (7) & 0.948 (3) & 1.000 (7) & 0.972 (4.5) & 0.972 (4.5) & 1.000 (7) & 0.383 (1) & 0.623 (2) \\
    WILT  & 1.000 (7.5) & 0.755 (3) & 0.999 (6) & 0.977 (5) & 0.902 (4) & 1.000 (7.5) & 0.009 (1) & 0.715 (2) \\
    WISCONSIN & 1.000 (7) & 0.922 (3) & 1.000 (7) & 0.961 (4.5) & 0.961 (4.5) & 1.000 (7) & 0.369 (1) & 0.701 (2) \\
    XOR   & 1.000 (6.5) & 0.486 (2) & 1.000 (6.5) & 1.000 (6.5) & 0.524 (3) & 1.000 (6.5) & 0.476 (1) & 0.753 (4) \\
    \midrule
    \textbf{Total Rank} & \textbf{134.5} & \textbf{55} & \textbf{128.5} & \textbf{91} & \textbf{76.5} & \textbf{134.5} & \textbf{20} & \textbf{44} \\
    \bottomrule
\end{tabular}%
    }
\label{tab:coverage_bin}%
\end{table*}%

\begin{table*}[h!]
\centering
\caption{Average accuracy metrics and their ranking per dataset on multiclass classification (all instances)}
\resizebox{0.9\textwidth}{!}{%
 \begin{tabular}{llllllllll}
     \toprule
          & RPART & STEL  & Pre-Forest-ORE & Forest-ORE & Forest-ORE+STEL & RF    & RIPPER \\
    \midrule
    ANNEAL & 0.885 (3) & 0.861 (1) & 0.908 (7) & 0.887 (4) & 0.894 (5) & 0.907 (6) & 0.870 (2) \\
    AUTO  & 0.626 (1.5) & 0.698 (3) & 0.752 (6) & 0.718 (5) & 0.716 (4) & 0.782 (7) & 0.626 (1.5) \\
    BANANA & 0.743 (6) & 0.717 (3) & 0.711 (1) & 0.740 (4.5) & 0.740 (4.5) & 0.748 (7) & 0.714 (2) \\
    CAR   & 0.935 (4) & 0.859 (2) & 0.943 (6) & 0.936 (5) & 0.918 (3) & 0.955 (7) & 0.846 (1) \\
    DERMA & 0.932 (2) & 0.939 (4) & 0.973 (6) & 0.935 (3) & 0.942 (5) & 0.979 (7) & 0.881 (1) \\
    ECOLI & 0.791 (4) & 0.788 (3) & 0.793 (5) & 0.776 (2) & 0.775 (1) & 0.798 (6) & 0.805 (7) \\
    GLASS & 0.628 (2) & 0.675 (5) & 0.708 (6) & 0.657 (3) & 0.658 (4) & 0.729 (7) & 0.609 (1) \\
    IRIS  & 0.938 (1.5) & 0.949 (6) & 0.940 (3) & 0.942 (4.5) & 0.942 (4.5) & 0.938 (1.5) & 0.964 (7) \\
    NEWTHYROID & 0.871 (1) & 0.891 (4) & 0.920 (7) & 0.892 (5) & 0.889 (3) & 0.914 (6) & 0.886 (2) \\
    PAGEBLOCKS & 0.950 (5) & 0.942 (1.5) & 0.953 (6) & 0.944 (3) & 0.942 (1.5) & 0.958 (7) & 0.947 (4) \\
    TEXTURE & 0.781 (4) & 0.549 (1) & 0.791 (5) & 0.772 (3) & 0.619 (2) & 0.922 (7) & 0.900 (6) \\
    THYROID & 0.949 (6) & 0.944 (3.5) & 0.947 (5) & 0.944 (3.5) & 0.942 (2) & 0.952 (7) & 0.940 (1) \\
    TITATNIC & 0.786 (4.5) & 0.784 (1.5) & 0.785 (3) & 0.788 (6.5) & 0.788 (6.5) & 0.786 (4.5) & 0.784 (1.5) \\
    VERTEBRAL & 0.845 (7) & 0.815 (3) & 0.832 (6) & 0.809 (2) & 0.817 (4) & 0.829 (5) & 0.717 (1) \\
    VOWEL & 0.499 (2.5) & 0.480 (1) & 0.749 (6) & 0.724 (5) & 0.499 (2.5) & 0.840 (7) & 0.628 (4) \\
    WINE  & 0.877 (1) & 0.925 (2.5) & 0.966 (6) & 0.942 (5) & 0.940 (4) & 0.979 (7) & 0.925 (2.5) \\
    WIRELESS & 0.933 (2) & 0.923 (1) & 0.955 (5) & 0.940 (3) & 0.941 (4) & 0.956 (6) & 0.961 (7) \\
    \midrule
    \textbf{Total Rank} & \textbf{57} & \textbf{46} & \textbf{89} & \textbf{67} & \textbf{60.5} & \textbf{105} & \textbf{51.5} \\
    \bottomrule
    \end{tabular}%
    }
\label{tab:acc_multi_all}%
\vspace{-3ex}
\end{table*}%

\begin{table*}[h!]
\centering
\caption{Average accuracy metrics and their ranking per dataset on multiclass classification (covered instances)}
\resizebox{0.9\textwidth}{!}{%
    \begin{tabular}{llllllllll}
   \toprule
          & RPART & STEL  & Pre-Forest-ORE & Forest-ORE & Forest-ORE+STEL & RF    & RIPPER \\
    \midrule
    ANNEAL & 0.885 (3) & 0.865 (2) & 0.911 (5) & 0.913 (6) & 0.916 (7) & 0.907 (4) & 0.734 (1) \\
    AUTO  & 0.626 (1) & 0.729 (3) & 0.752 (4) & 0.785 (7) & 0.78 (5) & 0.782 (6) & 0.641 (2) \\
    BANANA & 0.743 (5) & 0.788 (7) & 0.712 (2) & 0.741 (3) & 0.742 (4) & 0.748 (6) & 0.696 (1) \\
    CAR   & 0.935 (3) & 0.818 (2) & 0.943 (4) & 0.969 (6) & 0.975 (7) & 0.955 (5) & 0.629 (1) \\
    DERMA & 0.932 (2) & 0.957 (3.5) & 0.973 (6) & 0.957 (3.5) & 0.969 (5) & 0.979 (7) & 0.882 (1) \\
    ECOLI & 0.791 (2) & 0.808 (7) & 0.793 (3) & 0.803 (5.5) & 0.803 (5.5) & 0.798 (4) & 0.726 (1) \\
    GLASS & 0.628 (1) & 0.699 (3.5) & 0.708 (5) & 0.699 (3.5) & 0.711 (6) & 0.729 (7) & 0.639 (2) \\
    IRIS  & 0.938 (1.5) & 0.944 (4) & 0.94 (3) & 0.946 (5) & 0.952 (6) & 0.938 (1.5) & 0.986 (7) \\
    NEWTHYROID & 0.871 (1) & 0.901 (3) & 0.92 (7) & 0.909 (5) & 0.907 (4) & 0.914 (6) & 0.875 (2) \\
    PAGEBLOCKS & 0.95 (2.5) & 0.95 (2.5) & 0.955 (4) & 0.962 (6) & 0.964 (7) & 0.958 (5) & 0.77 (1) \\
    TEXTURE & 0.781 (1) & 0.801 (2) & 0.916 (4) & 0.929 (6) & 0.984 (7) & 0.922 (5) & 0.911 (3) \\
    THYROID & 0.949 (2) & 0.964 (4) & 0.979 (5) & 0.987 (6) & 0.995 (7) & 0.952 (3) & 0.742 (1) \\
    TITATNIC & 0.786 (2.5) & 0.81 (6) & 0.785 (1) & 0.787 (4) & 0.79 (5) & 0.786 (2.5) & 0.89 (7) \\
    VERTEBRAL & 0.845 (7) & 0.83 (4) & 0.832 (5) & 0.828 (2) & 0.838 (6) & 0.829 (3) & 0.585 (1) \\
    VOWEL & 0.499 (1) & 0.543 (2) & 0.755 (4) & 0.799 (6) & 0.79 (5) & 0.84 (7) & 0.683 (3) \\
    WINE  & 0.877 (1) & 0.947 (3) & 0.966 (6) & 0.959 (4.5) & 0.959 (4.5) & 0.979 (7) & 0.939 (2) \\
    WIRELESS & 0.933 (1) & 0.94 (2) & 0.955 (5) & 0.952 (3) & 0.954 (4) & 0.956 (6) & 0.959 (7) \\
    \midrule
    \textbf{Total Rank} & \textbf{37.5} & \textbf{60.5} & \textbf{73} & \textbf{82} & \textbf{95} & \textbf{85} & \textbf{43} \\
    \bottomrule
   \end{tabular}%
    }
\label{tab:acc_multi_cov}%
\end{table*}%

\begin{table*}[h!]
\centering
\caption{Average coverage metrics and their ranking per dataset on multiclass classification}
\resizebox{0.9\textwidth}{!}{%
    \begin{tabular}{llllllllll}
    \toprule
          & RPART & STEL  & Pre-Forest-ORE & Forest-ORE & Forest-ORE+STEL & RF    & RIPPER \\
    \midrule
    ANNEAL & 1.000 (6.5) & 0.948 (2) & 0.993 (5) & 0.960 (4) & 0.952 (3) & 1.000 (6.5) & 0.235 (1) \\
    AUTO  & 1.000 (6) & 0.921 (4) & 1.000 (6) & 0.892 (3) & 0.887 (2) & 1.000 (6) & 0.665 (1) \\
    BANANA & 1.000 (6.5) & 0.695 (2) & 0.999 (5) & 0.982 (4) & 0.971 (3) & 1.000 (6.5) & 0.416 (1) \\
    CAR   & 1.000 (6) & 0.765 (2) & 1.000 (6) & 0.952 (4) & 0.899 (3) & 1.000 (6) & 0.360 (1) \\
    DERMA & 1.000 (6.5) & 0.905 (2) & 0.999 (5) & 0.947 (4) & 0.941 (3) & 1.000 (6.5) & 0.672 (1) \\
    ECOLI & 1.000 (6) & 0.949 (2) & 1.000 (6) & 0.952 (4) & 0.950 (3) & 1.000 (6) & 0.565 (1) \\
    GLASS & 1.000 (6) & 0.849 (2) & 1.000 (6) & 0.917 (4) & 0.898 (3) & 1.000 (6) & 0.580 (1) \\
    IRIS  & 1.000 (6) & 0.809 (2) & 1.000 (6) & 0.982 (4) & 0.900 (3) & 1.000 (6) & 0.640 (1) \\
    NEWTHYROID & 1.000 (6) & 0.926 (2) & 1.000 (6) & 0.977 (4) & 0.937 (3) & 1.000 (6) & 0.262 (1) \\
    PAGEBLOCKS & 1.000 (6.5) & 0.981 (4) & 0.992 (5) & 0.967 (3) & 0.961 (2) & 1.000 (6.5) & 0.088 (1) \\
    TEXTURE & 1.000 (6.5) & 0.584 (2) & 0.793 (4) & 0.759 (3) & 0.537 (1) & 1.000 (6.5) & 0.896 (5) \\
    THYROID & 1.000 (6.5) & 0.897 (4) & 0.901 (5) & 0.877 (3) & 0.857 (2) & 1.000 (6.5) & 0.027 (1) \\
    TITATNIC & 1.000 (6) & 0.842 (2) & 1.000 (6) & 0.982 (4) & 0.959 (3) & 1.000 (6) & 0.145 (1) \\
    VERTEBRAL & 1.000 (6) & 0.860 (2) & 1.000 (6) & 0.957 (4) & 0.955 (3) & 1.000 (6) & 0.468 (1) \\
    VOWEL & 1.000 (6.5) & 0.816 (2) & 0.983 (5) & 0.885 (4) & 0.566 (1) & 1.000 (6.5) & 0.836 (3) \\
    WINE  & 1.000 (6) & 0.834 (2) & 1.000 (6) & 0.966 (4) & 0.958 (3) & 1.000 (6) & 0.591 (1) \\
    WIRELESS & 1.000 (6) & 0.828 (2) & 1.000 (6) & 0.970 (3.5) & 0.970 (3.5) & 1.000 (6) & 0.751 (1) \\
    \midrule
    \textbf{Total Rank} & \textbf{105.5} & \textbf{40} & \textbf{94} & \textbf{63.5} & \textbf{44.5} & \textbf{105.5} & \textbf{23} \\
    \bottomrule
   \end{tabular}%
    }
\label{tab:coverage_multi}%
\end{table*}%

\begin{table*}[h!]
\centering
\caption{Friedman test on the average results per dataset}
\resizebox{0.65\textwidth}{!}{%
    \begin{tabular}{lllr}
    \toprule
    \multicolumn{1}{l}{\textbf{Classification issue}} & \textbf{coverage} & \textbf{metric} & \multicolumn{1}{l}{\textbf{Friedman test pvalue}} \\
     \midrule
    \multicolumn{1}{r}{\multirow{3}[1]{*}{Binary classification}} & all   & accuracy & 9.26E-04 \\
          & covered & accuracy & 4.94E-05 \\
          & covered & coverage & 3.69E-24 \\
    \multicolumn{1}{r}{\multirow{3}[1]{*}{Multiclass classification}} & all   & accuracy & 2.86E-06 \\
          & covered & accuracy & 1.20E-05 \\
          & covered & coverage & 1.56E-17 \\

    \bottomrule
    \end{tabular}%
    }
\label{tab:Friedman_test}%

\end{table*}%

\section{Computational time required by Forest-ORE}
\label{appendix:exetime}
Table \ref{tab:exetime} reports the execution time spent by the Forest-ORE on the benchmarking datasets. 
\begin{table*}[h!]
\centering
\caption{Forest-ORE execution time (with Intel Core i7, in a Windows environment) }
\resizebox{0.9\textwidth}{!}{%
 \begin{tabular}{lrrrrrrrrrr}
    \toprule
          & \multicolumn{2}{c}{Extract rules} & \multicolumn{2}{c}{Preselect rules} & \multicolumn{2}{c}{Prepare opt. Inputs} & \multicolumn{2}{c}{Build opt. model} & \multicolumn{2}{c}{Run opt. Model} \\
          & \multicolumn{1}{c}{Mean} & \multicolumn{1}{c}{SE} & \multicolumn{1}{c}{Mean} & \multicolumn{1}{c}{SE} & \multicolumn{1}{c}{Mean} & \multicolumn{1}{c}{SE} & \multicolumn{1}{c}{Mean} & \multicolumn{1}{c}{SE} & \multicolumn{1}{c}{Mean} & \multicolumn{1}{c}{SE} \\
    \midrule
    ANNEAL & 4.28  & 0.04  & 13.40 & 0.14  & 26.49 & 0.44  & 39.54 & 0.65  & 5.53  & 1.88 \\
    APPENDICITIS & 0.88  & 0.02  & 1.92  & 0.05  & 1.76  & 0.03  & 4.13  & 0.08  & 0.76  & 0.07 \\
    AUTO  & 4.27  & 0.04  & 11.81 & 0.16  & 14.88 & 0.20  & 18.49 & 0.29  & 4.12  & 0.85 \\
    BANANA & 0.66  & 0.01  & 17.67 & 0.64  & 18.14 & 0.32  & 37.25 & 0.76  & 293.17 & 42.86 \\
    BANKNOTE & 1.53  & 0.03  & 9.37  & 0.23  & 11.39 & 0.12  & 29.48 & 0.34  & 0.85  & 0.11 \\
    BRCANCER & 0.99  & 0.02  & 5.66  & 0.09  & 13.52 & 0.30  & 34.86 & 0.77  & 25.81 & 4.93 \\
    CAR   & 7.50  & 0.03  & 37.92 & 0.51  & 93.46 & 0.59  & 157.01 & 1.06  & 41.31 & 7.45 \\
    CRYOTHERAPY & 1.00  & 0.03  & 1.88  & 0.05  & 1.99  & 0.05  & 4.71  & 0.12  & 0.66  & 0.11 \\
    DERMA & 2.26  & 0.05  & 7.07  & 0.12  & 15.95 & 0.46  & 20.55 & 0.58  & 1.49  & 0.47 \\
    ECOLI & 1.89  & 0.03  & 8.04  & 0.10  & 19.36 & 0.33  & 19.65 & 0.32  & 1.31  & 0.21 \\
    GLASS & 3.67  & 0.05  & 7.65  & 0.11  & 10.99 & 0.27  & 13.66 & 0.32  & 1.23  & 0.25 \\
    HABERMAN & 0.88  & 0.02  & 2.12  & 0.04  & 3.88  & 0.05  & 9.55  & 0.12  & 0.34  & 0.06 \\
    HEART & 2.02  & 0.04  & 6.82  & 0.12  & 11.14 & 0.12  & 26.92 & 0.36  & 102.56 & 22.20 \\
    HYPOTHYROID & 4.71  & 0.14  & 49.38 & 1.02  & 53.50 & 0.76  & 129.07 & 1.76  & 89.24 & 27.66 \\
    INDIAN & 6.20  & 0.08  & 10.02 & 0.25  & 11.83 & 0.36  & 27.67 & 0.58  & 85.96 & 23.40 \\
    ION   & 2.42  & 0.05  & 7.08  & 0.08  & 10.48 & 0.25  & 23.32 & 0.21  & 181.01 & 45.86 \\
    IRIS  & 0.50  & 0.02  & 0.78  & 0.03  & 1.23  & 0.05  & 2.31  & 0.08  & 0.10  & 0.04 \\
    MAMMOGRAPHIC & 3.16  & 0.05  & 17.01 & 0.17  & 22.71 & 0.47  & 60.68 & 1.20  & 1.00  & 0.21 \\
    MUSHROOM & 1.02  & 0.02  & 340.93 & 10.37 & 134.50 & 2.32  & 366.40 & 6.68  & 54.53 & 1.77 \\
    MUTAGENESIS & 1.29  & 0.03  & 13.31 & 0.21  & 16.24 & 0.29  & 49.20 & 0.99  & 4.49  & 1.05 \\
    NEWTHYROID & 1.09  & 0.03  & 2.63  & 0.05  & 3.89  & 0.09  & 7.42  & 0.18  & 0.49  & 0.08 \\
    PAGEBLOCKS & 7.65  & 0.11  & 540.69 & 16.98 & 185.05 & 1.85  & 331.24 & 3.34  & 27.89 & 5.56 \\
    PHONEME & 5.97  & 0.03  & 485.75 & 27.44 & 105.49 & 2.43  & 329.81 & 6.15  & 663.57 & 162.03 \\
    TEXTURE & 35.05 & 0.30  & 601.95 & 13.91 & 446.32 & 6.58  & 401.26 & 5.31  & 560.01 & 220.89 \\
    THYROID & 12.61 & 0.18  & 424.87 & 11.25 & 134.45 & 2.54  & 282.47 & 4.72  & 46.18 & 17.90 \\
    TICTACTOE & 3.11  & 0.02  & 15.02 & 0.13  & 28.58 & 0.19  & 78.36 & 0.66  & 3.84  & 0.79 \\
    TITATNIC & 0.32  & 0.03  & 1.69  & 0.02  & 5.15  & 0.09  & 9.89  & 0.19  & 1.60  & 0.09 \\
    VERTEBRAL & 2.47  & 0.06  & 7.50  & 0.12  & 12.17 & 0.18  & 24.91 & 0.30  & 1.54  & 0.32 \\
    VOTE  & 0.98  & 0.01  & 5.58  & 0.05  & 10.96 & 0.13  & 27.96 & 0.35  & 0.64  & 0.10 \\
    VOWEL & 19.26 & 0.12  & 50.24 & 0.61  & 202.25 & 1.94  & 183.55 & 1.89  & 25.34 & 3.81 \\
    WDBC  & 1.53  & 0.02  & 7.15  & 0.11  & 13.16 & 0.39  & 30.97 & 0.71  & 98.64 & 73.55 \\
    WILT  & 3.73  & 0.06  & 122.78 & 5.83  & 75.43 & 0.87  & 185.81 & 2.09  & 31.37 & 4.22 \\
    WINE  & 1.06  & 0.02  & 3.16  & 0.08  & 5.38  & 0.17  & 10.41 & 0.32  & 1.16  & 0.15 \\
    WIRELESS & 4.13  & 0.03  & 58.46 & 1.53  & 171.40 & 1.73  & 288.91 & 2.60  & 23.71 & 7.62 \\
    WISCONSIN & 1.11  & 0.05  & 6.03  & 0.12  & 14.16 & 0.44  & 35.47 & 0.72  & 8.44  & 2.41 \\
    XOR   & 0.34  & 0.00  & 1.28  & 0.02  & 2.05  & 0.04  & 4.19  & 0.04  & 0.05  & 0.00 \\
    \bottomrule
    \end{tabular}%
  
}
\label{tab:exetime}%
\end{table*}%

\end{document}